\documentclass{article}

\PassOptionsToPackage{numbers, sort&compress}{natbib}

\usepackage{tikz}
\usetikzlibrary{positioning}
\usepackage[preprint]{neurips_2026}

\usepackage[utf8]{inputenc} %
\usepackage[T1]{fontenc}    %
\usepackage{hyperref}       %
\usepackage{url}            %
\usepackage{booktabs}       %
\usepackage{longtable}      %
\usepackage{amsfonts}       %
\usepackage{nicefrac}       %
\usepackage{microtype}      %
\usepackage{amsmath}        %
\usepackage{mathrsfs}       %
\usepackage{xcolor}         %
\usepackage{graphicx}       %
\usepackage{subcaption}     %
\usepackage{tcolorbox}      %
\tcbuselibrary{breakable, skins}   %
\usepackage{fvextra}        %

\title{Mechanistically Interpretable Neural Encoding Reveals Fine-Grained Functional Selectivity in Human Visual Cortex}

\author{
    Idan Daniel Grosbard\textsuperscript{1} \qquad
    Mor Geva\textsuperscript{2,}\thanks{Equal senior authorship.} \qquad
    Galit Yovel\textsuperscript{1,3,}\footnotemark[\value{footnote}] \vspace{5pt} \\
    \textsuperscript{1}Sagol School of Neuroscience, \\
    \textsuperscript{2}Blavatnik School of Computer Science and AI, \\
    \textsuperscript{3}School of Psychological Sciences, \\
    Tel Aviv University \vspace{5pt} \\
    \texttt{\{idangrosbard@mail, morgeva@tauex, gality@tauex\}.tau.ac.il}
}

\begin{document}

\maketitle

\begin{abstract}
A central goal in understanding human vision is to uncover the visual features that drive neuronal activity. A growing body of work has used artificial neural networks as encoding models to predict cortical responses to natural images, revealing the visual content that activates category-selective regions. However, existing approaches are largely correlational and treat the encoder as a black box, leaving open which image features drive each voxel's response. We introduce Mechanistically Interpretable Neural Encoding (MINE), a framework that opens this black box by applying mechanistic-interpretability tools to localize the features within natural images that drive millimeter-scale (i.e., voxel-level) activity. MINE predicts each voxel's response using language-aligned image representations, and produces semantically interpretable descriptions of the features critical for the voxel's activation. We further generalize these per-image features into per-voxel functional profiles. To validate the per-image descriptions, we show they are sufficient to generate images that elicit voxel responses matching the responses to the original images, more accurately than images generated from random or low-attribution controls. Moreover, counterfactually inserting or removing the predicted features from images shifts activation in the expected direction, providing causal evidence. Counterfactual editing guided by the per-voxel activation profiles produces even stronger activation shifts, indicating that the profiles faithfully capture each voxel's selectivity. Finally, we apply MINE to well-studied category-selective brain regions, showing it recovers their known categorical preferences while revealing fine-grained unique voxel structure within each region. Overall, our results establish mechanistic interpretability as a path to discover and causally validate fine-grained hypotheses about neural function.
\end{abstract}

\section{Introduction}\label{sec:intro}
A key step towards understanding how the brain processes visual information is to explain the functional role of its underlying computational units \citep{marrUnderstandingComputationUnderstanding1976,marrVisualInformationProcessing1980}. Early works applied controlled comparisons of different visual stimuli to identify functionally selective cortical regions, i.e., regions whose responses are reliably stronger for one stimulus category (e.g., faces, places, bodies) than for others \cite{kanwisherFusiformFaceArea1997,epsteinCorticalRepresentationLocal1998,downingCorticalAreaSelective2001a}. While this approach has led to significant findings, it is limited by its hypothesis-driven nature and the need for large datasets of hand-picked images. Recent works have studied the functional selectivity profiles using the alignment between Artificial Neural Networks (ANNs) and the neural responses of the visual cortex to these images \citep{matsuyamaLaVCaLLMassistedVisual2025,hwangSilicoMappingVisual2025,wassermanBrainExploreLargeScaleDiscovery2025,luoBrainscubaFinegrainedNatural2023,yangCLIPMSMMultiSemanticMapping2025,luoBrainDiffusionVisual2023,gaoBrainLMMLabelFreeFramework2026,luoBrainMappingDense2025}. This path enabled the prediction of neural responses based on large datasets of naturalistic images. For example, \citet{wassermanBrainExploreLargeScaleDiscovery2025, hwangSilicoMappingVisual2025} used this approach to identify novel functional regions encoding concepts such as specific locations and tool use. \citet{luoBrainscubaFinegrainedNatural2023,matsuyamaLaVCaLLMassistedVisual2025} have characterized the functional selectivity of individual voxels ($\sim$1 mm³ volumes sampled by functional magnetic resonance imaging, fMRI), while the works of \citet{yangCLIPMSMMultiSemanticMapping2025,gaoBrainLMMLabelFreeFramework2026} identified poly-semantic functional selectivity profiles.

Although these works provide valuable insights, their correlational nature contains several limitations. First, correlational evidence — however carefully constructed — cannot fully separate genuine selectivity from spurious associations introduced by dataset structure \citep{matsuyamaLaVCaLLMassistedVisual2025,hwangSilicoMappingVisual2025,wassermanBrainExploreLargeScaleDiscovery2025}. Moreover, many studies identified preferred images (i.e., images that elicit the strongest activations), rather than the critical features within images that drive the activation \citep{luoBrainscubaFinegrainedNatural2023,matsuyamaLaVCaLLMassistedVisual2025,hwangSilicoMappingVisual2025,wassermanBrainExploreLargeScaleDiscovery2025}. Other studies that did localize critical features within images had limited validation for their contribution to the voxel's activation \citep{luoBrainMappingDense2025,adeliTransformerBrainEncoders2025,fengInterpretableVisualDecoding2025}. A different line of works, while generating higher-resolution functional profile hypotheses, their validation typically confirms broad alignment with the categorical selectivity of the surrounding region, leaving the fine-grained voxel-level profiles directly untested \citep{yangCLIPMSMMultiSemanticMapping2025,luoBrainDiffusionVisual2023,luoBrainMappingDense2025,luoBrainscubaFinegrainedNatural2023,gaoBrainLMMLabelFreeFramework2026}.  Similar limitations have been identified in the field of ANN interpretability research \citep{bolukbasiInterpretabilityIllusionBERT2021,ahmadCausalAnalysisRobust2024}, and have led to the adoption of \textit{Mechanistic Interpretability} (MI) based research methods \citep{sharkeyOpenProblemsMechanistic2025}, a school focused on understanding the internal mechanisms by which models implement their behavior, often through causal interventions \citep{daviesCognitiveRevolutionInterpretability2024}.

\begin{figure}[t]
  \centering
  \begin{subfigure}[t]{0.22\textwidth}
    \centering
    \includegraphics[height=2.7cm]{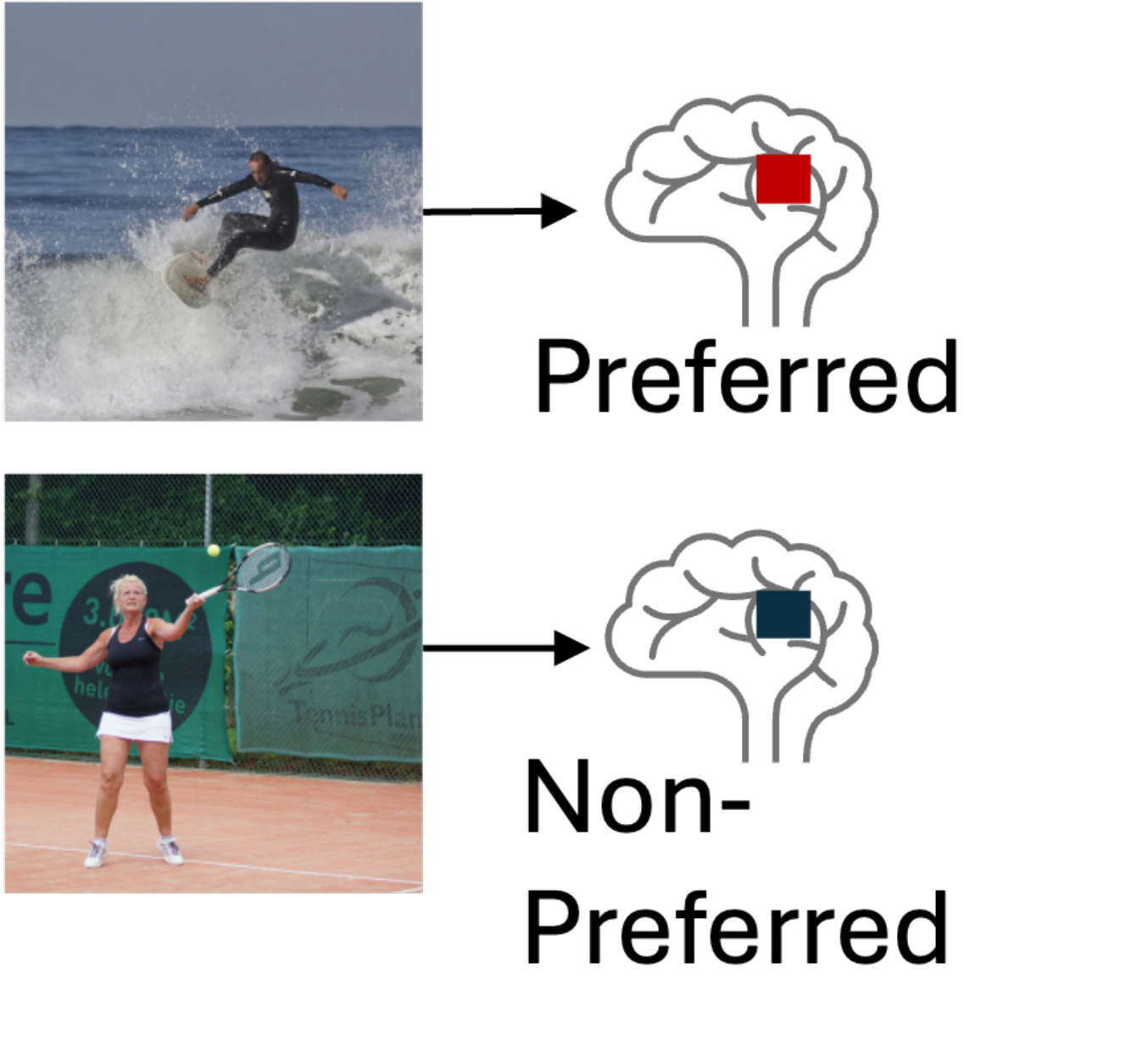}
    \caption{Encoder training}
    \label{fig:pipeline:stage1}
  \end{subfigure}
  \hfill
  \begin{subfigure}[t]{0.36\textwidth}
    \centering
    \includegraphics[height=2.7cm]{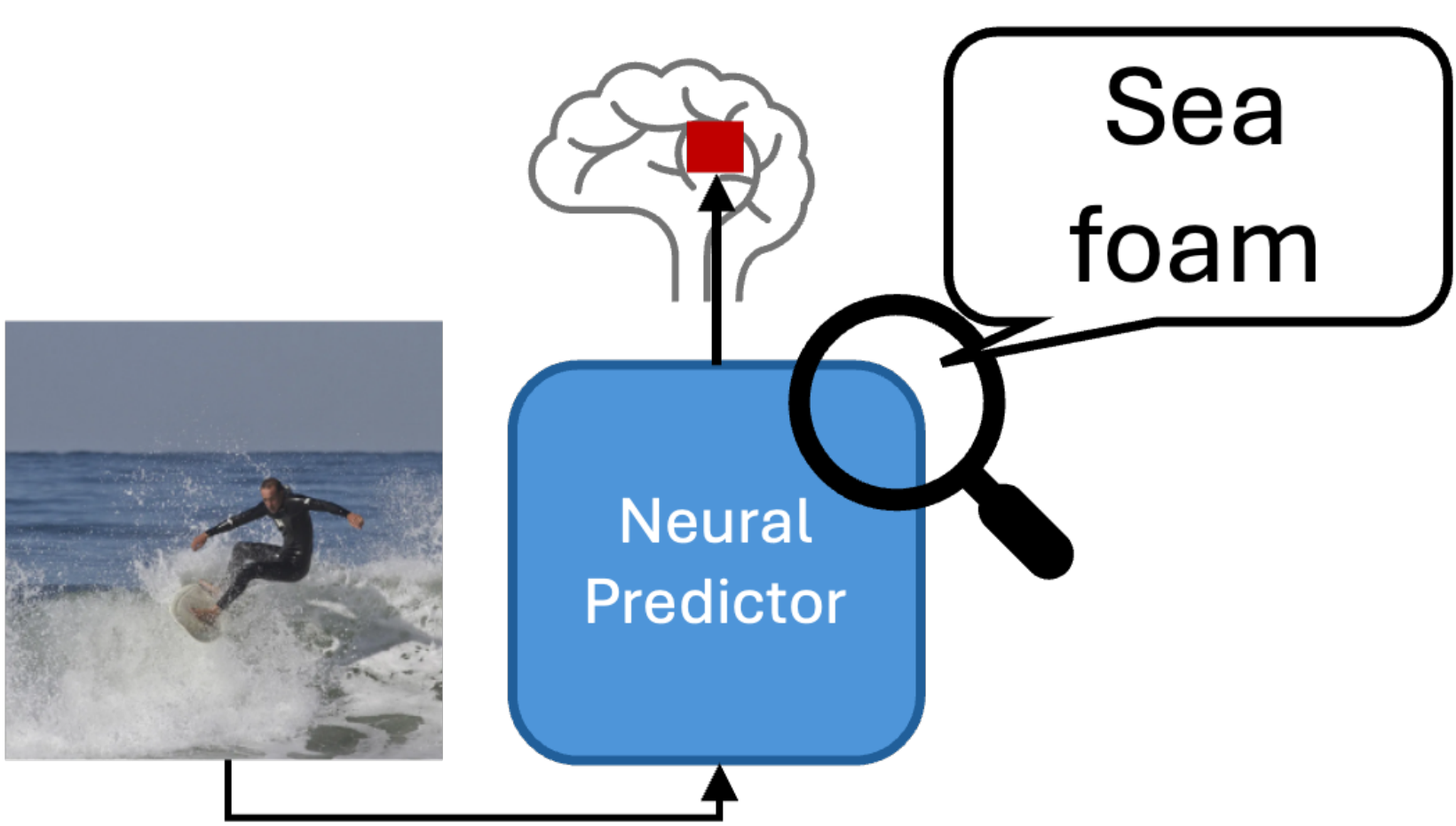}
    \caption{Hypothesis generation}
    \label{fig:pipeline:stage2}
  \end{subfigure}
  \hfill
  \begin{subfigure}[t]{0.37\textwidth}
    \centering
    \includegraphics[height=2.7cm]{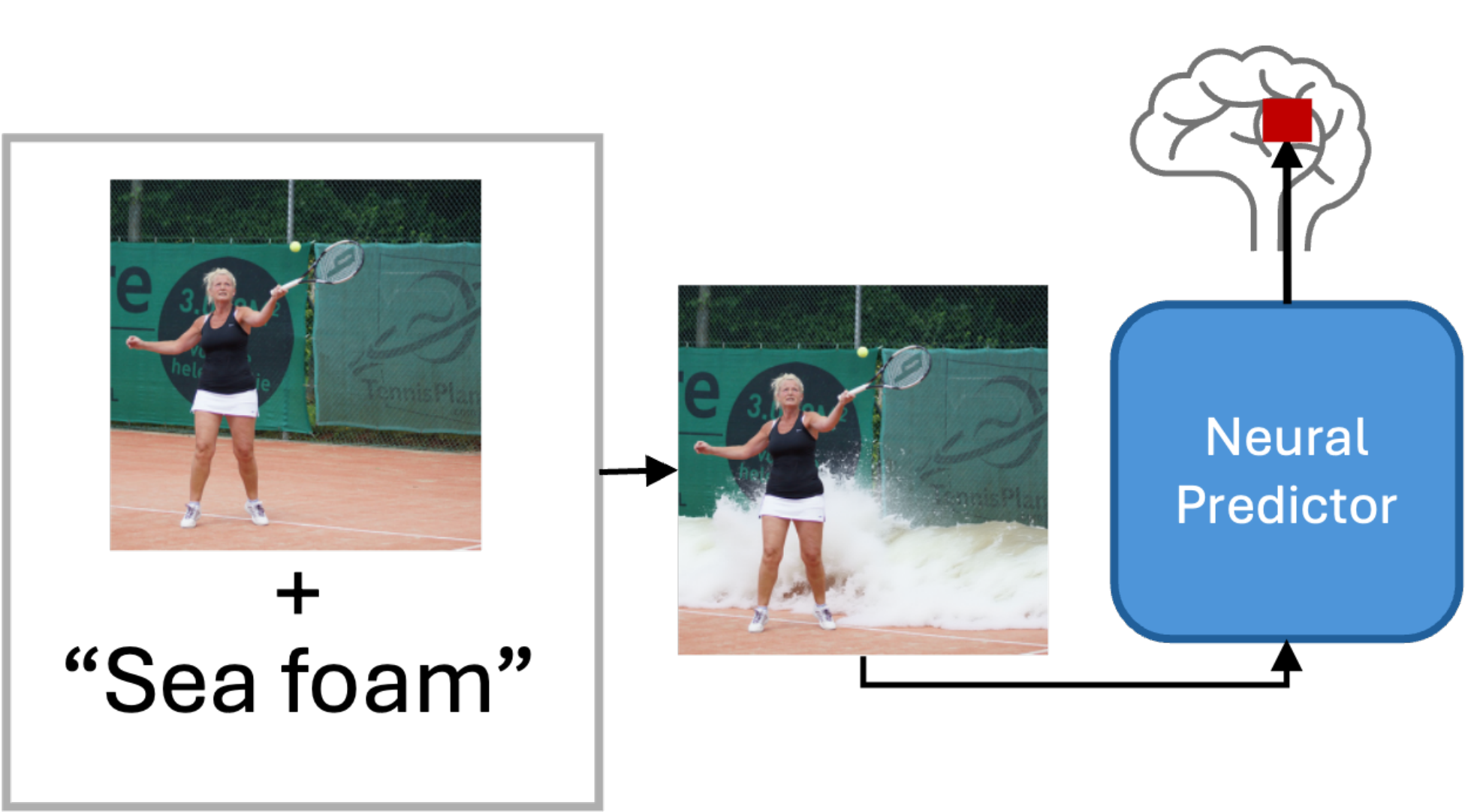}
    \caption{Counterfactual validation}
    \label{fig:pipeline:stage3}
  \end{subfigure}
  \caption{Overview of the MINE framework. (a) A neural encoder is trained on textually aligned image representations to predict voxel activations to natural images. This reveals preferred images - images that generate strong activations, and non-preferred images - images that generate weak activations. (b) Interpretability tools are applied to the trained encoder to extract per-image critical features from preferred images and aggregate them into a voxel profile. (c) The hypothesized critical features are validated using counterfactual image editing, which manipulates images according to the identified features and measures the predicted change in voxel activation.}
  \label{fig:pipeline}
\end{figure}

Drawing on this paradigm shift in ANN interpretability, we propose a new framework for interpreting neural encoding models, termed \textbf{Mechanistically Interpretable Neural Encoding} (MINE)\footnote{Code is available at \url{https://github.com/idangrosbard/MINE-Framework.git}.}, which relies on MI tools. In this framework, we study the mechanism of how an encoder uses the input information to predict neural activity. By understanding the model mechanism, we can identify the per-image visual features that drive the neural response prediction. Specifically, to ensure the model uses semantically interpretable features we train an encoder to predict neural recordings based on highly-detailed textually-aligned image representations. We then localize the critical information used to predict the neural responses. Using standard interpretability tools for decoding representations \citep{nostalgebraistInterpretingGPTLogit2020}, we can generate textual hypotheses about the visual features that a voxel responds to. Finally, we validate these hypotheses using counterfactual image editing, surgically manipulating images to change the model prediction by adding, or removing, candidate critical features. An overview of the MINE framework is illustrated in Figure~\ref{fig:pipeline}. To summarize, our contribution is threefold:
\begin{enumerate}
    \item We propose a mechanistic framework for modeling neural activity with built-in interpretability, to generate fine-grained, testable hypotheses about the features driving individual voxels' responses.
    \item We validate the framework with causal counterfactual analyses, showing that our method faithfully reveals the critical features driving each voxel's response.
    \item Applied to category-selective regions, our method recovers their known categorical preferences and uncovers novel fine-grained unique voxel structure, providing both validation and novel insights into voxel-level organization.
\end{enumerate}

\section{Related work}\label{sec:related}

\paragraph{Hypothesis generation based on neural modelling}\label{sec:related:encoding}
Several works have used encoding models, i.e., neural networks trained to predict brain activity from a stimulus, to profile voxels' functional selectivity.
One line of work \citep{yangCLIPMSMMultiSemanticMapping2025,luoBrainscubaFinegrainedNatural2023,yuMetaLearningInContextTransformer2025a} has utilized the high alignment between CLIP \citep{radfordLearningTransferableVisual2021} and neural responses to images in high-level visual cortex \citep{wangBetterModelsHuman2023} to effectively map voxels to embeddings in CLIP's representational space, allowing the use of decoding methods that rely on CLIP's language alignment \citep{oikarinenClipdissectAutomaticDescription2022,mokadyClipCapCLIPPrefix2021}. While providing a powerful framework for succinct concepts, these methods share a common limitation introduced by CLIP's short captions \cite{urbanekPictureWorthMore2024}, preventing the model from generating fine-grained hypotheses (see Appendix~\ref{app:captions} for an empirical comparison between caption-based and image-based encoders). We overcome this limitation by mapping voxel representations to LLaVA image-tokens \citep{liuVisualInstructionTuning2023,liuImprovedBaselinesVisual2024}, which provide detailed and grounded descriptions of the image content \citep{neoInterpretingVisualInformation2024}.
A second line of work similarly fit in-silico models of the visual cortex, which were used to evaluate the predicted neural response to large image datasets \citep{gaoBrainLMMLabelFreeFramework2026,matsuyamaLaVCaLLMassistedVisual2025,hwangSilicoMappingVisual2025,wassermanBrainExploreLargeScaleDiscovery2025}. These works used Vision Language Models (VLMs) to caption images preferred by different neural regions. While revealing novel functional selectivity patterns, these works suffer from limitations inherent in correlational analyses --- an annotator bias introduced by the image-captioning LLM, and imbalanced evaluation datasets \citep{bolukbasiInterpretabilityIllusionBERT2021}. To mitigate annotator bias, we analyzed our model mechanistically and apply causal validation to every step in the hypothesis generation pipeline. Furthermore, we assessed the model on images from the NSD dataset \citep{allenMassive7TFMRI2022}, limiting dataset to images that were presented in the fMRI recording.

\paragraph{Hypothesis generation based on image-generation}\label{sec:related:decoding}
Complementary to encoding-based approaches, several studies have adopted a decoding-based approach to characterize voxel profiles. The works of \citep{gazivSelfsupervisedNaturalImage2022,gazivMoreMeetsEye2021} used fMRI-to-image decoders to evaluate the information available in different neural regions by applying counterfactual manipulation of neural response and observing the changes in the generated images. However, this approach is only applicable for identifying the shared contribution of voxel groups, due to the high degree of mutual information between voxels. To overcome this, we use an encoding approach that models each voxel independently.
A different approach attempted to identify voxels' selectivity by identifying repeating patterns in attention maps \citep{luoBrainMappingDense2025,fengInterpretableVisualDecoding2025,adeliTransformerBrainEncoders2025} or gradient-based image generation \citep{luoBrainDiffusionVisual2023}; however, they lack causal evaluation of these regions' contribution to the model \citep{ahmadCausalAnalysisRobust2024,jainAttentionNotExplanation2019}. We overcome this by causally validating the critical tokens' contribution \citep{mengLocatingEditingFactual2022}, and decode these tokens to reveal semantic critical features.

\paragraph{Inspecting representations of large language and vision models}\label{sec:related:interpretability}
Our framework relies on understanding the neural encoder model based on its use of VLM representations. Understanding these internal representations has been a central target of interpretability research, highlighted as a potential path to increase the safety and usability of these models \citep{sharkeyOpenProblemsMechanistic2025}. A common approach for inspecting representations is \texttt{logit-lens} \citep{nostalgebraistInterpretingGPTLogit2020}, which converts them into a distribution over vocabulary tokens via linear projection onto the model's embedding space.
\citet{gevaTransformerFeedForwardLayers2022,darAnalyzingTransformersEmbedding2023} showed that linearly decoding LLM internal representations using the model's vocabulary often reveals distributions of semantic or syntactic concepts. 
\citet{gevaDissectingRecallFactual2023} used this property to decode intermediate representations of named entities to describe their related attributes. 
Due to the reliance on pretrained LLMs in VLMs, works such as \citet{neoInterpretingVisualInformation2024} have further applied this approach to image tokens, revealing a distribution of visually grounded tokens. 
Our work harnesses these findings using LLaVA representations \citep{liuVisualInstructionTuning2023,liuImprovedBaselinesVisual2024} to predict fMRI recordings, and to generate hypotheses about voxels' preferred features in natural images by decoding critical tokens.

\section{The MINE framework}\label{sec:methods}
In this section, we introduce the MINE framework. Section~\ref{sec:methods:setting} formalizes the task of identifying voxels' functional selectivity. Section~\ref{sec:methods:model} presents the MINE model, explaining how the proposed architecture can approximate voxel-level signal-detection behavior. Finally, Section~\ref{sec:methods:hypothesis} shows how our model can be used to extract textual explanations of the learned hypotheses for each voxel.

\subsection{Modeling voxel selectivity}\label{sec:methods:setting}
Voxels in high-level visual cortex tend to respond strongly when their preferred features are present in the stimulus, and remain near or below baseline otherwise. Following this observation, we model each voxel as a signal detector: its activation reflects the probability that the stimulus is drawn from a ``signal'' distribution $H_1$ (containing the voxel's preferred features) rather than a ``noise'' distribution $H_0$ (see Appendix~\ref{app:neuroimaging} for a brief introduction to neuroimaging, and justification for this approach). Formally, given a general set of visual features $\mathcal{F}$, we consider a stimulus $x$ as a subset of visual features $x\subset\mathcal{F}$. For example, an image of a surfer could be described as a set containing the surfer, surfboard, ocean, waves, skies, etc. Given a subset of features $f\subset\mathcal{F}$, we define $H_1$ as follows:
\[\text{Pr}[H_1 | x]=\mathbb{I}[f \cap x \neq \emptyset]\]
Where $\mathbb{I}$ is the indicator function. That is, images sampled from the signal distribution $H_1$ contain a voxel's critical feature. Denoting all voxels of interest as $\mathcal{V}$, we define the underlying function of a voxel $v\in\mathcal{V}$ that detects a set of features $f_v\subset \mathcal{F}$ as:
\[h:\mathcal{V}\times \mathcal{P}(\mathcal{F})\rightarrow\mathbb{R}, \quad h(v, x)\approx \text{Pr}[H_1 | x]
\]
Where $\mathcal{P}(\mathcal{F})$ is the power set of $\mathcal{F}$. Under this setting, we attempt to train a model $\hat{h}:\mathcal{V}\times \mathcal{P}(\mathcal{F})\rightarrow\mathbb{R}$ that approximates $h$, and understand its mechanism in order to identify for each voxel $v\in\mathcal{V}$ its critical set of features $f_v\subset \mathcal{F}$.

\subsection{Neural encoder model}\label{sec:methods:model}
Let $(v,x,y)\sim\mathcal{D}$ denote a voxel $v\in\mathcal{V}$, stimulus $x\subset\mathcal{F}$, and recorded fMRI activation $y\in\mathbb{R}$, with voxel function $h(v,x)=y$. We train a neural encoder $\hat{h}$ to approximate $h$ by minimizing the MSE loss $\|\hat{h}(v,x) - y\|_2^2$. We implement this as a query operation over stimulus-related information. Specifically, given a stimulus representation $x\in\mathbb{R}^{s\times d_s}$, where $s$ is a sequence-length of image-tokens representing the stimulus, and $d_s$ is the image-token dimension, our model learns a $d_v$-dimensional voxel embedding $e_v\in\mathbb{R}^{d_v}$ and applies cross-attention over the stimulus-tokens \citep{vaswaniAttentionAllYou2017}:
\[\operatorname{Attention}(e_v, x) = \operatorname{softmax}\left(\frac{e_v W^Q (x W^K)^T}{\sqrt{d_v}}\right) x W^V\]
Where $W^Q\in\mathbb{R}^{d_v\times d_v}$ and $W^K,W^V\in\mathbb{R}^{d_s\times d_v}$ are the query, key, and value matrices of the attention layer, respectively. The attended information is then projected through a feedforward layer, and a voxel-specific linear projection to output the estimated response $\hat{y}$. Further details on the model and its training are provided in Section~\ref{sec:results:setting}. As shown in previous works, transformer layers can implement trigger-conditional behavior both theoretically and empirically \citep{ran-miloAttentionSinksAre2026,barberoWhyLLMsAttend2025,guoActiveDormantAttentionHeads2024}, making this a viable model to approximate signal-detection.

\subsection{Model analysis and hypothesis generation}\label{sec:methods:hypothesis}
By choosing an input representation that conforms to the linear representation hypothesis \citep[LRH; ][]{parkLinearRepresentationHypothesis2024} with respect to a feature set $\mathcal{F}$ --- i.e., features in $\mathcal{F}$ correspond to linear directions in the representation space --- we can re-express the encoder's attention in terms of feature coordinates and identify which features it relies on. Concretely, the LRH assumes the existence of a matrix $E_\mathcal{F} \in \mathbb{R}^{\mathcal{F} \times d_s}$, where each row is a specific feature's direction in the representation space, and a decoding matrix $E_\mathcal{F}^{-1} \in \mathbb{R}^{d_s \times \mathcal{F}}$ such that $x\cdot E_\mathcal{F}^{-1}$ expresses the how well each feature in $\mathcal{F}$ describes the stimulus $x$. Denoting $g(v,x):=\operatorname{softmax}\left(\frac{e_v W^Q (x W^K)^T}{\sqrt{d}}\right)$ where $g:\mathcal{V}\times\mathbb{R}^{s \times d_s}\rightarrow[0,1]^{s}$ calculates importance weights for each image-token, we can write the attention operation as:
\[\operatorname{Attention}(e_v, x) = \Big(g(v,x)\cdot x\cdot E_\mathcal{F}^{-1}\Big) \cdot \Big(E_{\mathcal{F}} \cdot W^V\Big)\]
Where $g(v,x)\cdot x\cdot E_\mathcal{F}^{-1}\in\mathbb{R}^\mathcal{F}$ is a weight vector over the feature set, and $E_{\mathcal{F}} \cdot W^V \in \mathbb{R}^{\mathcal{F} \times d_v}$ are learned feature embeddings for use in fMRI prediction. Ergo, information querying from the image-token representations can be regarded as a weight calculation over the feature set that describes the image content.
To support open-ended hypothesis generation, we choose an input representation that can be decoded into natural language, setting $\mathcal{F}:=\mathcal{W}$, where $\mathcal{W}$ is a textual vocabulary. Previous results have shown that LLM token representations and VLM image-token representations can be linearly decoded to the model's vocabulary \citep{nostalgebraistInterpretingGPTLogit2020, darAnalyzingTransformersEmbedding2023,neoInterpretingVisualInformation2024} using a linear unembedding operator $E_\mathcal{W}^{-1} \in \mathbb{R}^{d_s\times |\mathcal{W}|}$, such that $x\cdot E_\mathcal{W}^{-1} \in \mathbb{R}^{s \times|\mathcal{W}|}$ is a weight vector with higher values to vocabulary-tokens describing the input-image-token. Furthermore, recent work has shown that LLMs can infer the meaning of the input token based on this vocabulary-token weight vector \citep{gur-ariehEnhancingAutomatedInterpretability2025}. Plugging in this unembedding matrix to the previous equation, we get that the model queries image features to predict the neural response. We use this approach to interpret the critical feature for each (voxel, image) pair.

\section{Framework evaluation}\label{sec:results}

\subsection{Experimental setting}\label{sec:results:setting}
We trained our neural encoder model on the Natural Scenes Dataset (\texttt{NSD}) \citep{allenMassive7TFMRI2022}, a neuro-imaging dataset containing fMRI scans of 8 subjects observing a total of $73{,}000$ images from the \texttt{MSCOCO} dataset \citep{linMicrosoftCocoCommon2014}. In this work we focused on fMRI recordings from 4 subjects, who viewed a shared set of $1{,}000$ images (held out for model analysis). Each subject viewed a total of $10{,}000$ images, with 3 repetitions per image. We trained our model to predict the average fMRI response to an image across repetitions and focused on voxels in the middle visual cortex, as defined by \citet{wangProbabilisticMapsVisual2015}, with a noise ceiling of at least 0.4 (an upper bound on each voxel's predictability set by trial-to-trial measurement noise), yielding a total of $83{,}071$ voxels. We trained the model on batches of 64 images, each containing 256 voxels, using a single cross-attention transformer layer with 128 attention heads. For image representations, we used the 30th layer of \texttt{LLaVA-1.5-7B} \citep{liuImprovedBaselinesVisual2024,liuVisualInstructionTuning2023}. For further details on the dataset, image representations, and model training, we refer the reader to Appendices \ref{app:model:data}, \ref{app:model:rep}, and \ref{app:model:train}. Before turning to hypothesis generation, we verified that the model achieved competitive predictive performance: it reached an average explained variance of $\bar{R^2}=0.3$ (95\% CI across voxels = [$0.2992$, $0.3011$]; per-voxel $R^2$ range = [$-0.0573$, $0.7756$]), comparable to other state-of-the-art models \citep{beliyWisdomCrowdBrains2025,adeliTransformerBrainEncoders2025}. For further details on model evaluation, please refer to Appendix~\ref{app:model:eval}. For further details about implementation and compute, please refer to \ref{app:compute}.

\subsection{Identifying per-image critical features}\label{sec:results:hypotheses}
Next, we identified the critical features used by the model to predict the neural response for each individual image. For each (voxel, image) pair, we located the critical image-tokens needed to predict the voxel response. Specifically, to rank the image-tokens we used the Integrated Gradients (IG) method \citep{sundararajanAxiomaticAttributionDeep2017} a gradient based attribution method to quantify the contribution of each input-token to the model's performance. We then retained the top 50 for analysis (see Appendix~\ref{app:token-patching} for an analysis of how the top-K IG tokens contribute to voxel-response prediction). We decoded these tokens using the transposed vocabulary embedding $E_\mathcal{W}^T$, as suggested by \citet{darAnalyzingTransformersEmbedding2023}. For each of the top 50 image-tokens (by IG score), we extracted the top 10 vocabulary-tokens (by \texttt{logit-lens} score), yielding 500 (non-unique) words per (voxel, image) pair. Given this set of words, we used \texttt{Claude-Haiku-4.5} \citep{IntroducingClaudeHaiku} to automatically describe the visual content that drives the voxel's response to each image, returning a short sentence describing the critical feature \mbox{\citep{gur-ariehEnhancingAutomatedInterpretability2025,singhExplainingBlackBox2023}}. To assess if this method can identify selectivity patterns, we applied it to artificial neurons with predetermined selectivity profiles and confirmed that it recovers the planted features; further details are available in Appendix~\ref{app:artificial-neurons}. In Section~\ref{sec:results:profile}, we aggregate these per-image hypotheses to produce per-voxel selectivity profiles.

\subsection{Decoded-features based image-reconstruction}\label{sec:results:reconstruction}
To validate the hypothesis generation method, we first performed a reconstruction-based evaluation. Recalling the setting defined in Section~\ref{sec:methods:setting}, we note the following: Let $\mathcal{D}_f$ be the distribution of images that contain a non-empty subset of $f$, and $\mathcal{D}_{\setminus f}$ be the distribution that does not contain such a subset it follows that: 
\[x,\hat{x}\sim\mathcal{D}_f. |h(v,x) - h(v,\hat{x})|\approx 0, \quad x,\hat{x}\sim\mathcal{D}_{\setminus f}. |h(v,x) - h(v,\hat{x})|\approx 0\]
Meaning that a decoding function $d(v,x):\mathcal{V}\times \mathcal{P}(\mathcal{F}) \rightarrow \mathcal{P}(\mathcal{F})$ is accurate if the images that contain the same critical feature as the source image, will have a similar response to the recorded response to the image $\mathbb{E}_{\hat{x}\sim\mathcal{D}_{d(v,x)}}[h(v,x) - h(v,\hat{x})]\approx0$. To test this, we compared our decoding method with several control: decoding critical features based on 50 randomly selected image-tokens, and on the 50 lowest IG scores image-tokens. We generated candidate images from the decoded descriptions using \texttt{FLUX.2-Klein-9B} \citep{BlackforestlabsFlux22026}, a text-conditioned image generation model. We measured the error between the recorded fMRI response and the predicted response for the generated images. To test differences between conditions, we used a linear mixed model (LMM) with decoding method as a fixed effect and accounted for random effects at the voxel- and image-levels. As shown in Figure~\ref{fig:generative:token-selection-differences}, images generated from top-IG image-tokens had a lower average prediction error than those generated from either randomly chosen, or the lowest-IG image-tokens.

\subsection{Decoded-features discriminability}\label{sec:results:discriminability}
\begin{figure}[t]
  \centering
  \begin{subfigure}[t]{0.28\textwidth}
    \centering
    \includegraphics[width=\textwidth, height=6.5cm, keepaspectratio=true]{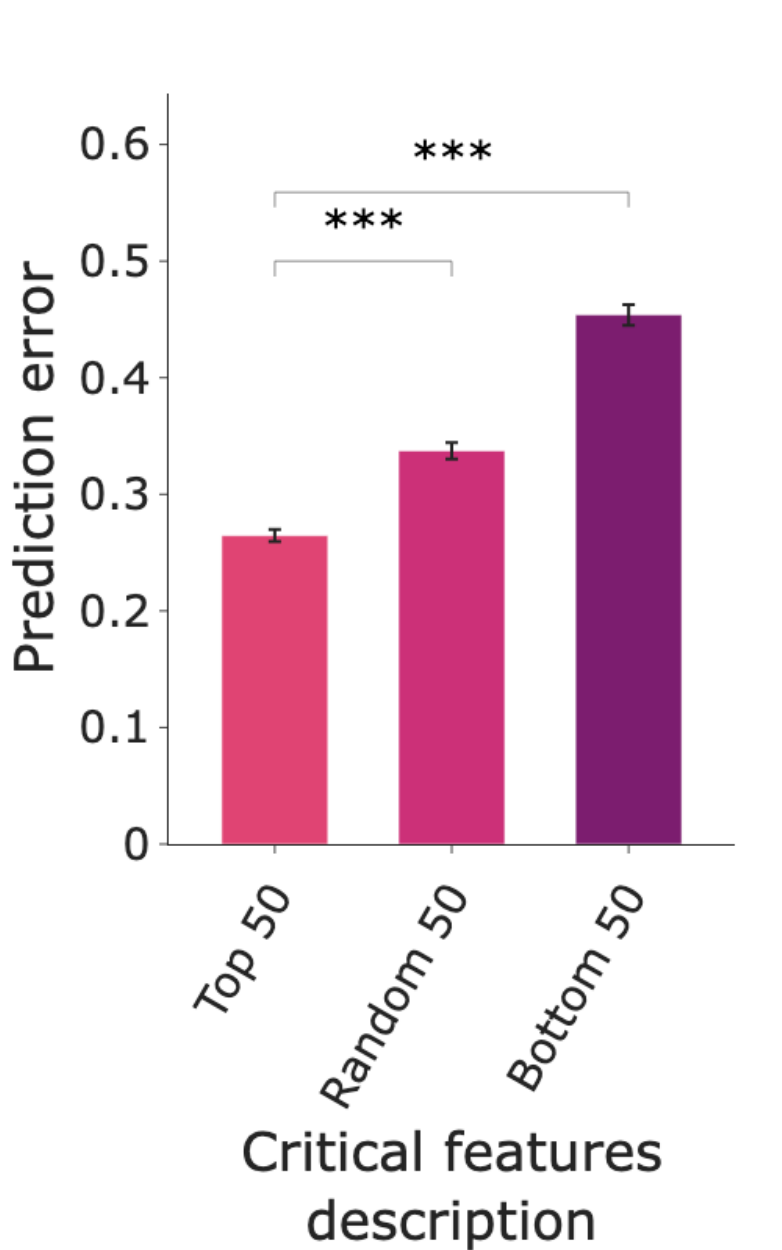}
    \caption{Reconstruction error per description source}
    \label{fig:generative:token-selection-differences}
  \end{subfigure}
  \hfill
  \begin{subfigure}[t]{0.36\textwidth}
    \centering
    \includegraphics[width=\textwidth, height=6.5cm, keepaspectratio=true]{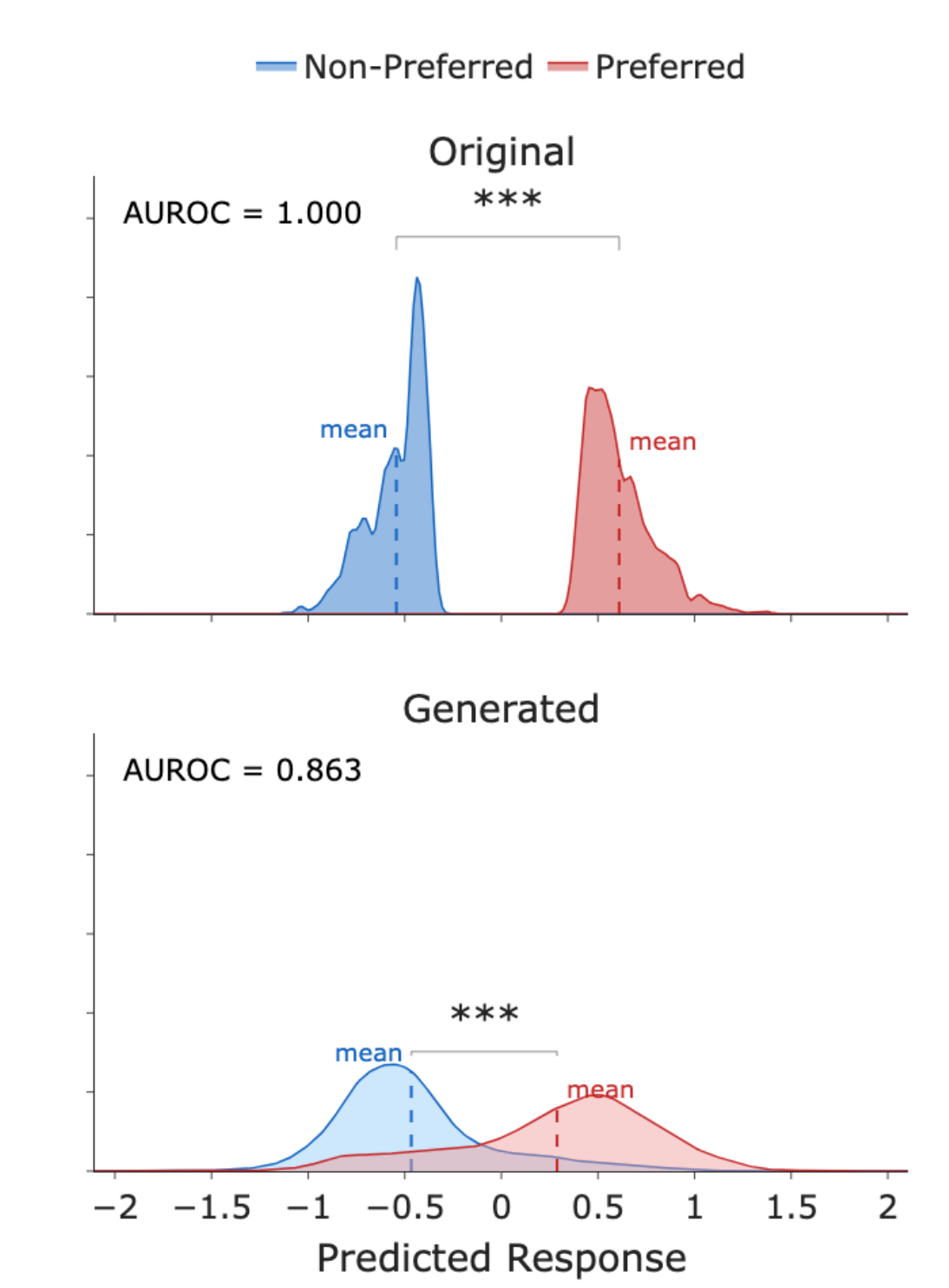}
    \caption{Activation distributions for original and generated images}
    \label{fig:generative:generated-distributions}
  \end{subfigure}
  \hfill
  \begin{subfigure}[t]{0.32\textwidth}
    \centering
    \includegraphics[width=\textwidth, height=6.5cm, keepaspectratio=true]{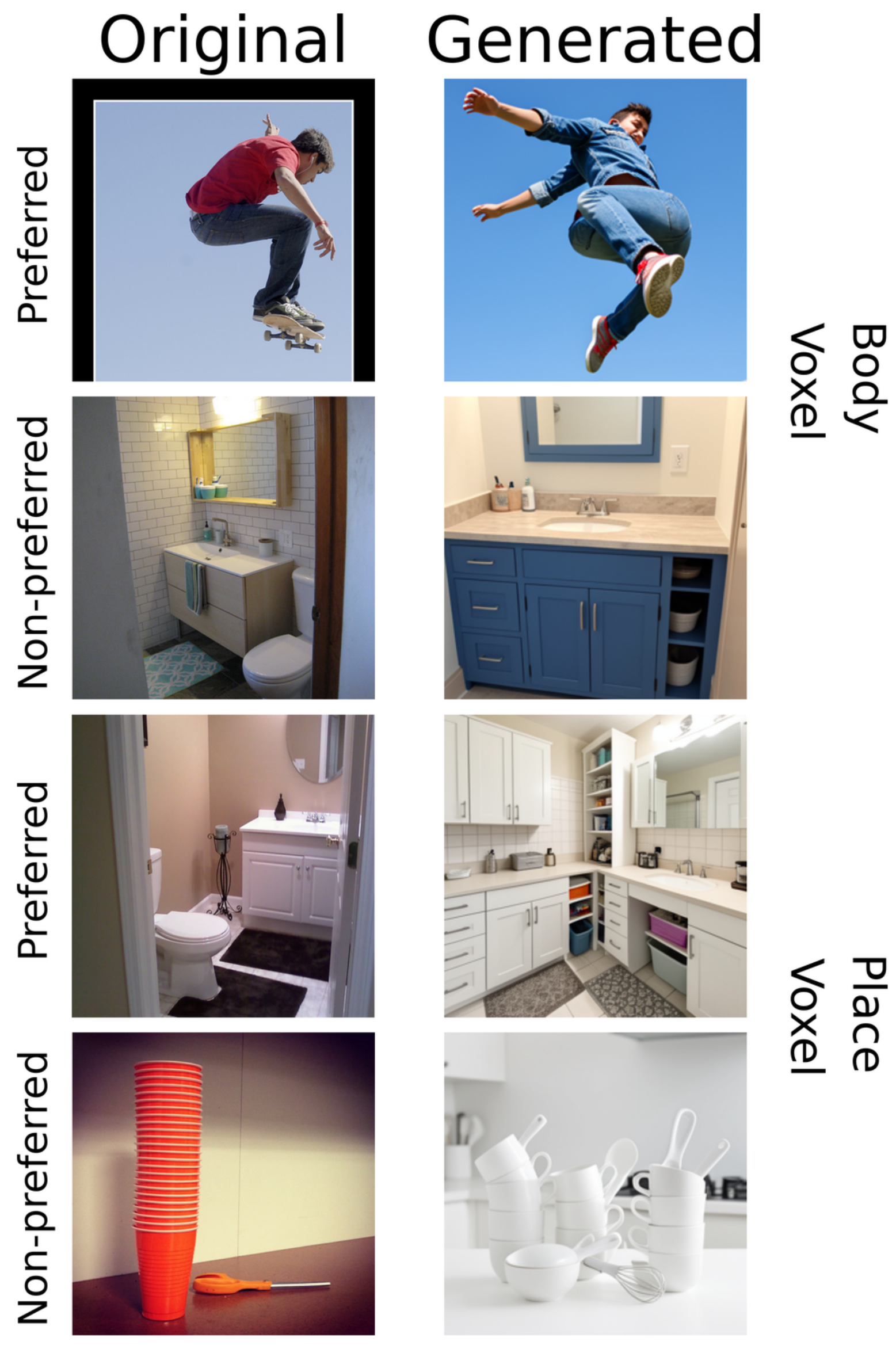}
    \caption{Original and generated samples}
    \label{fig:generative:generated-samples}
  \end{subfigure}
  \caption{Predicted activations for stimuli generated according to critical features. (a) Error distribution for images generated according to features from image-tokens with highest IG scores, random image-tokens, and lowest IG scores. (b) Distribution of predicted activation for preferred (red) and non-preferred (blue) images, for original stimuli (top), and generated stimuli (bottom). (c) Examples of original (top) preferred, and non-preferred images, and images generated according to their critical features (bottom). Significance markers *** indicate a p-value $\leq 0.001$.}
\end{figure}

For further validation, we note a second observation: Let $f\subset\mathcal{F}$ be a set of critical features for a voxel $v\in\mathcal{V}$. The definition in Section~\ref{sec:methods:setting}, $h(v,x)\approx\mathbb{I}[f\cap x\neq\emptyset]$, suggests discriminability between the two distributions:
\[\mathbb{E}_{x\sim\mathcal{D}_f, x'\sim\mathcal{D}_{\setminus f}}[h(v,x) - h(v,x')]\approx 1\]
To evaluate this, we used the method in Section~\ref{sec:methods:hypothesis} to identify the critical features and to generate images from the two distributions using \texttt{FLUX.2-Klein-9B} \citep{BlackforestlabsFlux22026}. An example of original and generated images is shown in Figure~\ref{fig:generative:generated-samples}. For each voxel, we identified preferred (high-activation) and non-preferred (low-activation) test images, applied our hypothesis-generation pipeline to each set, and generated new images from the resulting critical features, generating samples from $\mathcal{D}_{d(v,x)}$ and $\mathcal{D}_{\setminus d(v,x)}$ respectively. We then tested whether we could distinguish between the activation distributions of the generated images by fitting an LMM with fixed effects for the source image distribution (preferred vs. non-preferred) and random effects for voxels. As shown in Figure~\ref{fig:generative:generated-distributions}, the model clearly separates between activating-, and non-activating-generated images, demonstrating that our approach successfully decodes the critical features used by our model per-image.

\subsection{Counterfactual analysis}\label{sec:results:counterfactual}
\begin{figure}[t]
  \centering
  \begin{subfigure}[t]{0.25\textwidth}
    \centering
    \includegraphics[width=\textwidth, height=5.5cm, keepaspectratio=true]{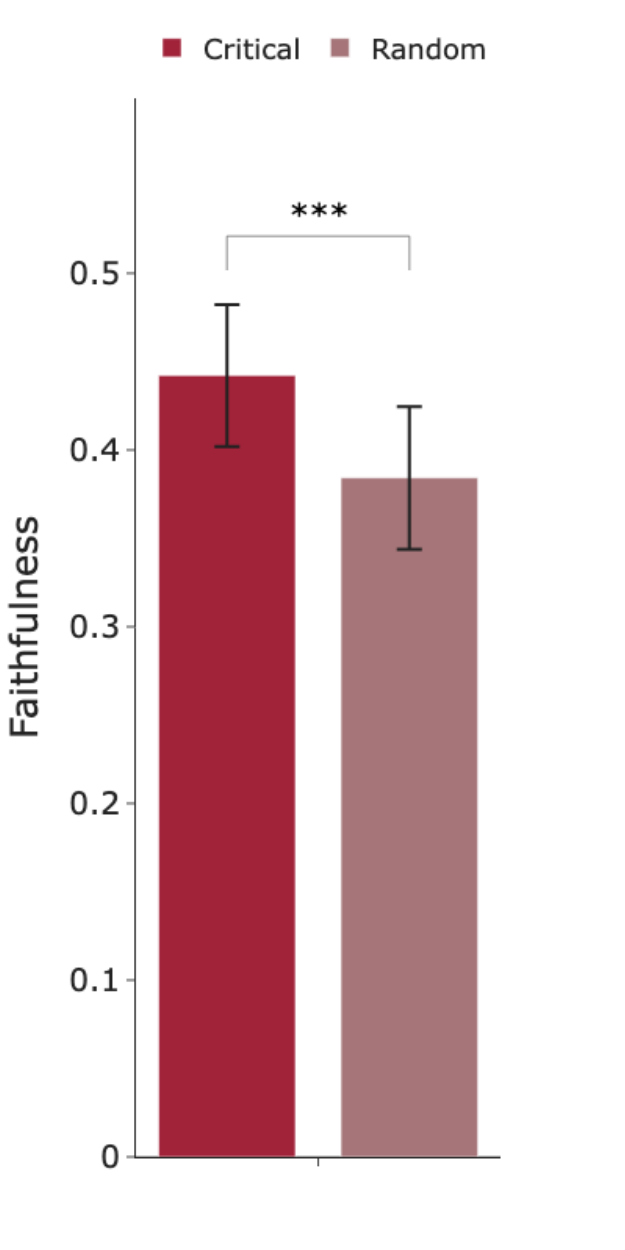}
    \caption{}
    \label{fig:counterfactual:token-selection-differences}
  \end{subfigure}
  \hfill
  \begin{subfigure}[t]{0.25\textwidth}
    \centering
    \includegraphics[width=\textwidth, height=5.5cm, keepaspectratio=true]{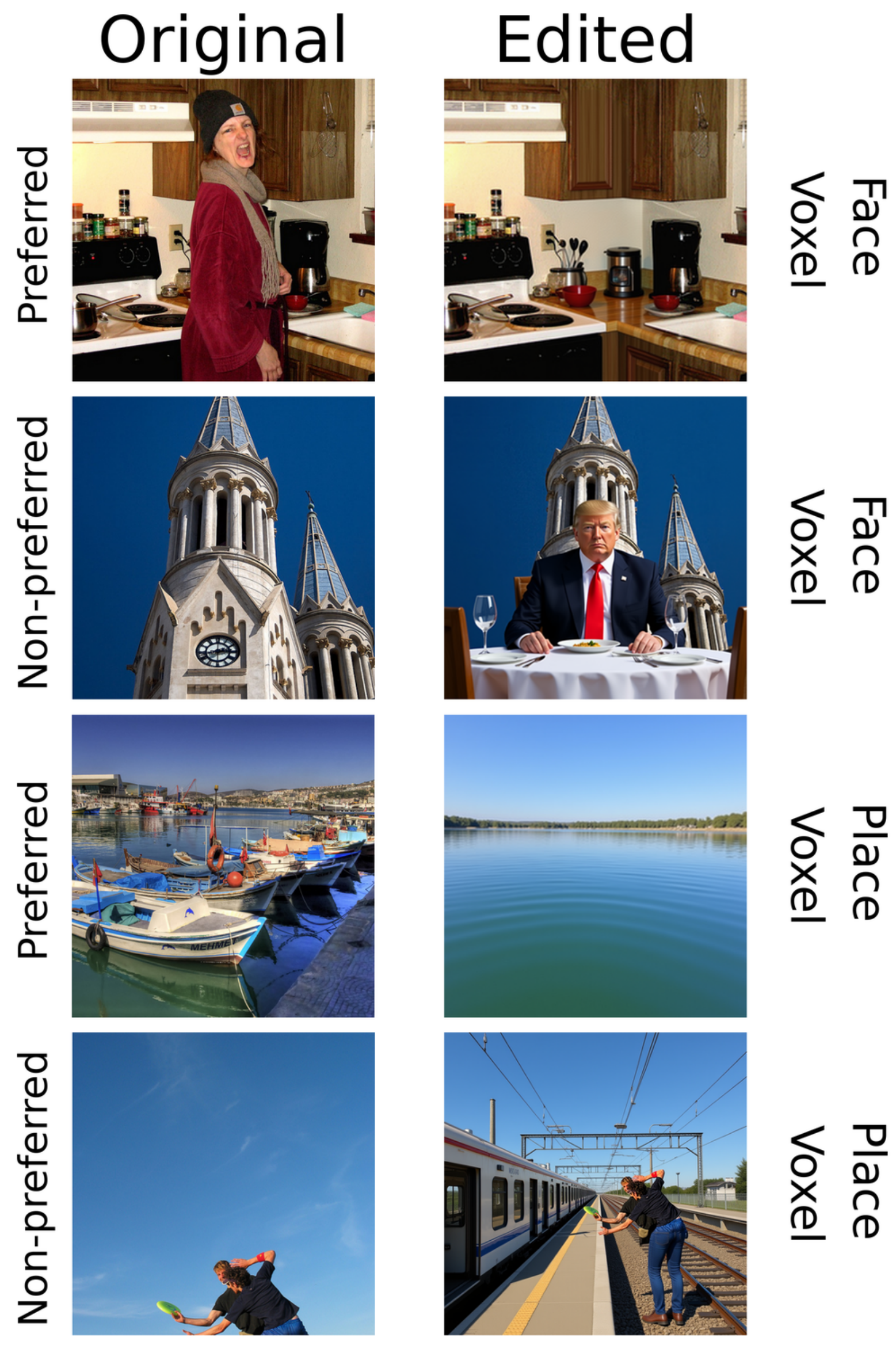}
    \caption{}
    \label{fig:counterfactual:generated-samples}
  \end{subfigure}
  \hfill
  \begin{subfigure}[t]{0.48\textwidth}
    \centering
    \includegraphics[width=\textwidth, height=5.5cm, keepaspectratio=true]{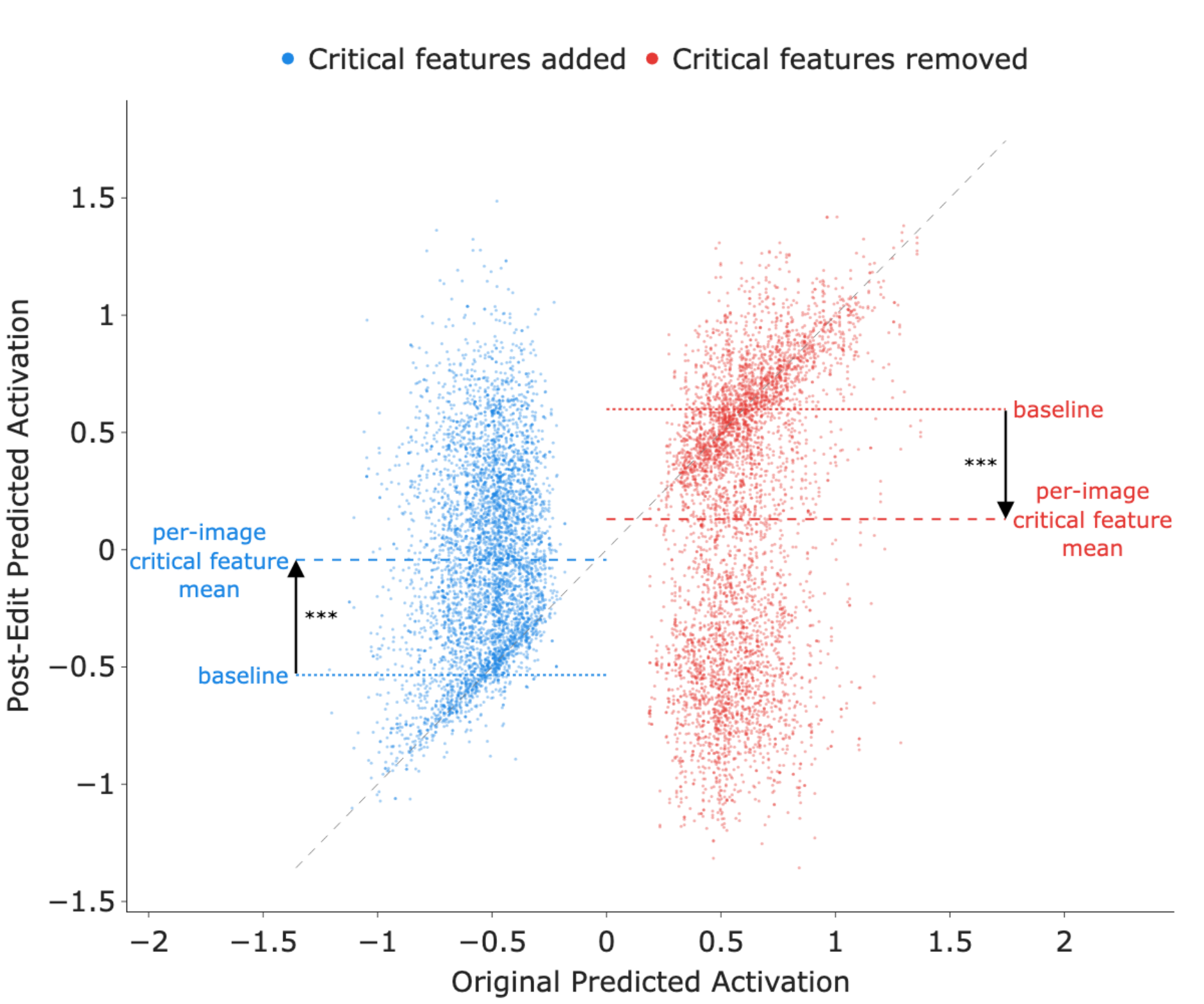}
    \caption{}
    \label{fig:counterfactual:activation-distributions}
  \end{subfigure}
  \caption{Predicted activations for images counterfactually edited according to critical features. (a) Faithfulness of counterfactually edited images according to features extracted from critical image-tokens (red), or random image-tokens (pink), after adding features from preferred images. (b) Examples of original (top) preferred (right), and non-preferred (left) images, and images counterfactually edited according to critical features (bottom). (c) Distribution of predicted activation for preferred (red) and non-preferred (blue) images. The X-axis coordinate indicates the image's recorded response. The y-axis coordinate indicates the model-predicted activation after counterfactual editing. Significance markers ** indicate a p-value $\leq 0.01$, *** indicate a p-value $\leq 0.001$.}
\end{figure}

Thus far, our analyses showed that images generated from critical features yield low reconstruction error (Section~\ref{sec:results:reconstruction}) and reproduce voxel discriminability between preferred and non-preferred stimuli (Section~\ref{sec:results:discriminability}). However, our approach enables a causal test of whether the discovered features drive the model's predictions. First, should a feature $f\subset x$ drives the predicted activation, then removing said feature from the image should produce a weaker response than removing other image-features.
\[x' = x \setminus f \quad \Rightarrow \quad h(v,\, x') < h(v,\, x)\]
Similarly, adding critical features to non-preferred images should produce a stronger voxel response.
\[x' \in \mathcal{F} \setminus f, \quad \hat{x} = x' \cup f \quad \Rightarrow \quad h(v,\, \hat{x}) > h(v,\, x')\]
To evaluate these metrics we applied counterfactual editing and measured its effect on the voxels' activation . Specifically, given a set of preferred images, we identified the critical features in each image using the method in Section~\ref{sec:results:hypotheses}, and applied counterfactual image editing by removing the critical feature from the image, or adding them to non-preferred images. We then generated an editing prompt for \texttt{FLUX.1-Kontext-dev} \citep{labsFLUX1KontextFlow2025}, a diffusion model trained for localized image editing, to apply the relevant edit operation. Examples of such edits are shown in Figure~\ref{fig:counterfactual:generated-samples}. To test the change in activation, we fitted an LMM with a fixed effect for type of editing (add, remove), and random effects for voxels, edited images, and reference images. Critical-feature-based edits produce significant shifts in predicted activation in the expected direction (decrease for removal, increase for addition), see Figure~\ref{fig:counterfactual:activation-distributions}.
Second, should the critical feature $f$ faithfully explain the predicted activation, we can measure the ratio of change in predicted response when we add $f$ to a non-preferred image. Given two stimuli $x\in\mathcal{F}, x'\in\mathcal{F}\setminus f$ with $f \subset x \setminus x'$, we define the faithfulness metric \citep{hannaHaveFaithFaithfulness2024} as:
\[\operatorname{faithfulness}(x, x', f)=\frac{h(v,x'\cup f) - h(v,x')}{h(v,x) - h(v,x')}\]
Specifically, given a critical feature $f$, we expect the following to hold:
\[\mathbb{E}_{f' \sim x}[\operatorname{faithfulness}(x, x', f) - \operatorname{faithfulness}(x, x', f')] > 0\]
Where $f' \sim x$ means a random sample of features from $x$. To measure faithfulness, we applied counterfactual editing as described previously, editing non-preferred images by adding either critical features or features derived from random image-tokens taken from preferred images. To test whether critical-features-based editing is significantly more faithful than random features-based editing, we fitted an LMM with a fixed effect for critical-feature decoding method, and random effects for voxels, non-preferred images, and preferred reference images. As shown in Figure~\ref{fig:counterfactual:token-selection-differences}, performing counterfactual editing based on critical image-tokens is significantly more faithful than when using randomly chosen image image-tokens. 

\subsection{Voxel profile evaluation}\label{sec:results:profile}

\begin{figure}[t]
  \centering
  \begin{minipage}[c]{0.40\textwidth}
    \centering
    \begin{subfigure}[t]{0.49\textwidth}
      \centering
      \includegraphics[width=\textwidth, height=3.4cm, keepaspectratio=true]{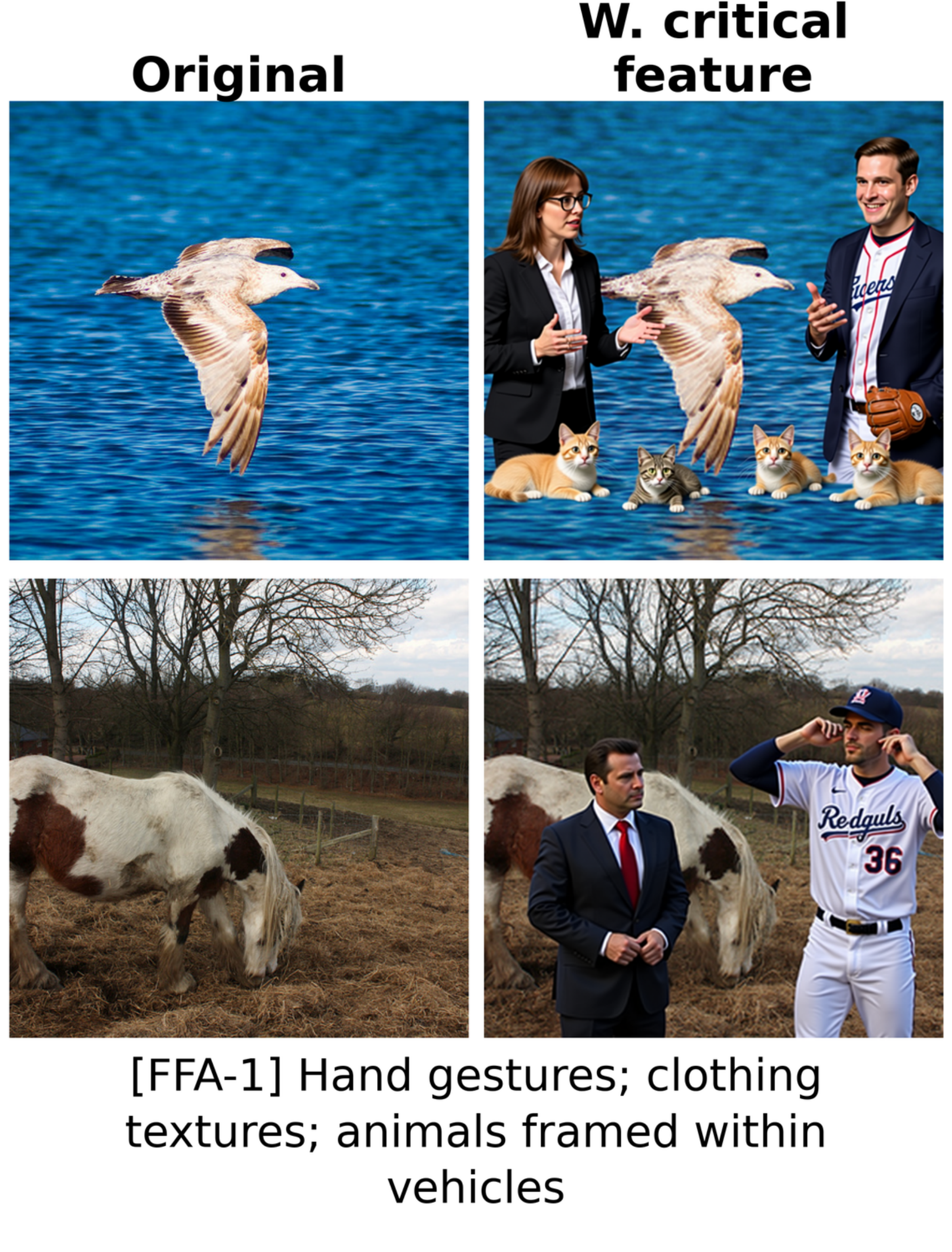}
      \caption{}
      \label{fig:profiles:samples-v6036}
    \end{subfigure}
    \hfill
    \begin{subfigure}[t]{0.49\textwidth}
      \centering
      \includegraphics[width=\textwidth, height=3.4cm, keepaspectratio=true]{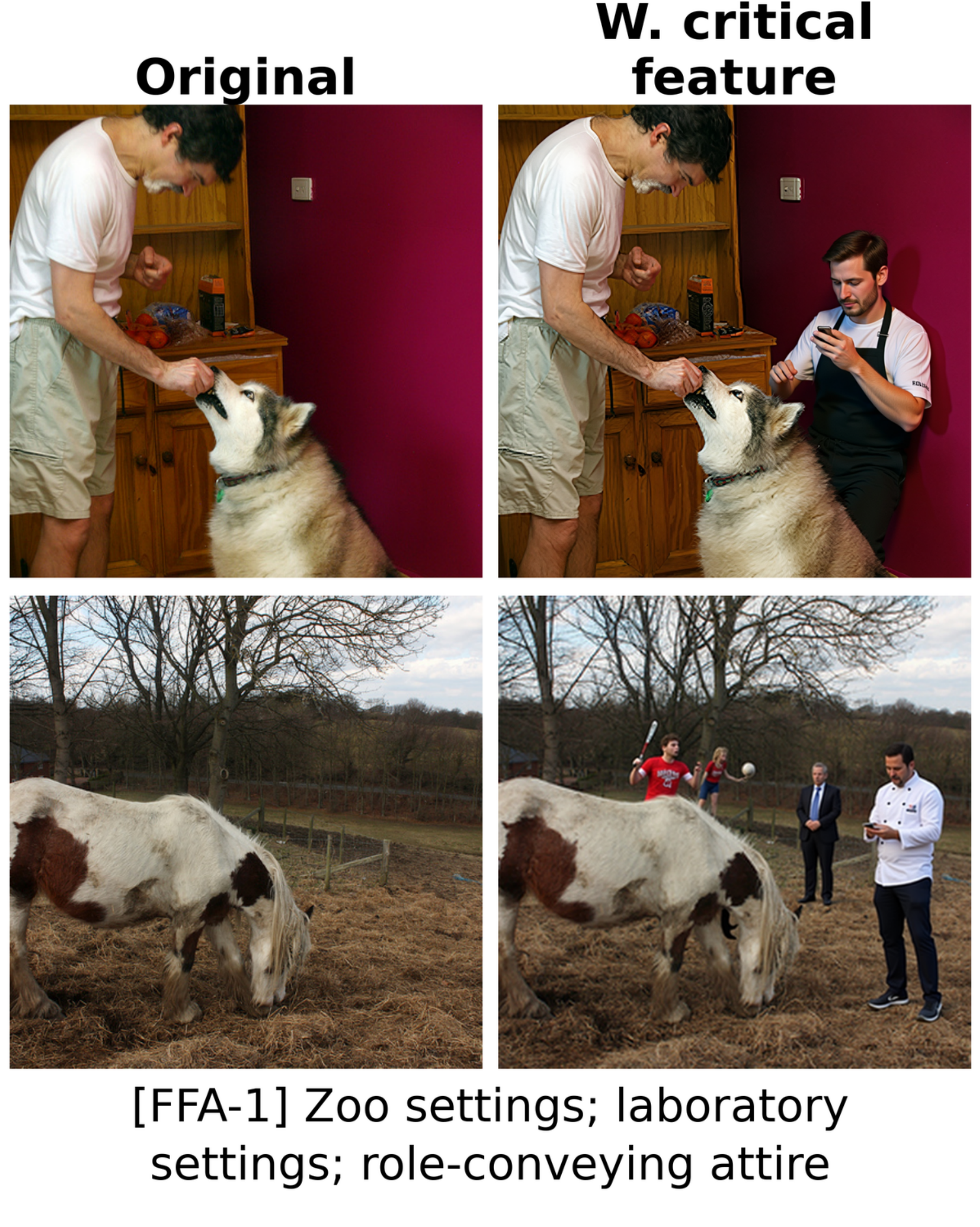}
      \caption{}
      \label{fig:profiles:samples-v6287}
    \end{subfigure}
    \\[0.5em]
    \begin{subfigure}[t]{0.49\textwidth}
      \centering
      \includegraphics[width=\textwidth, height=3.4cm, keepaspectratio=true]{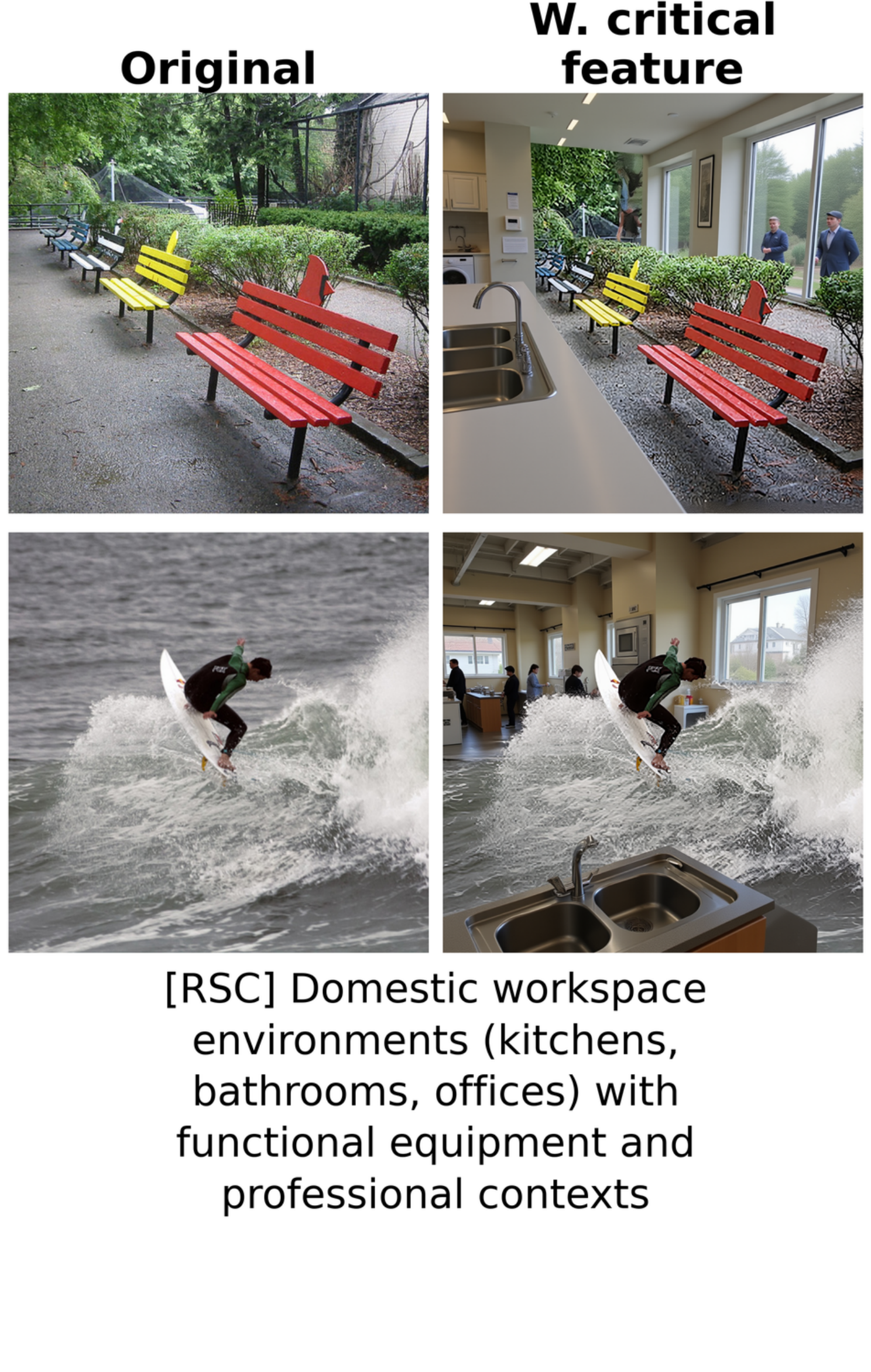}
      \caption{}
      \label{fig:profiles:samples-v7835}
    \end{subfigure}
    \hfill
    \begin{subfigure}[t]{0.49\textwidth}
      \centering
      \includegraphics[width=\textwidth, height=3.4cm, keepaspectratio=true]{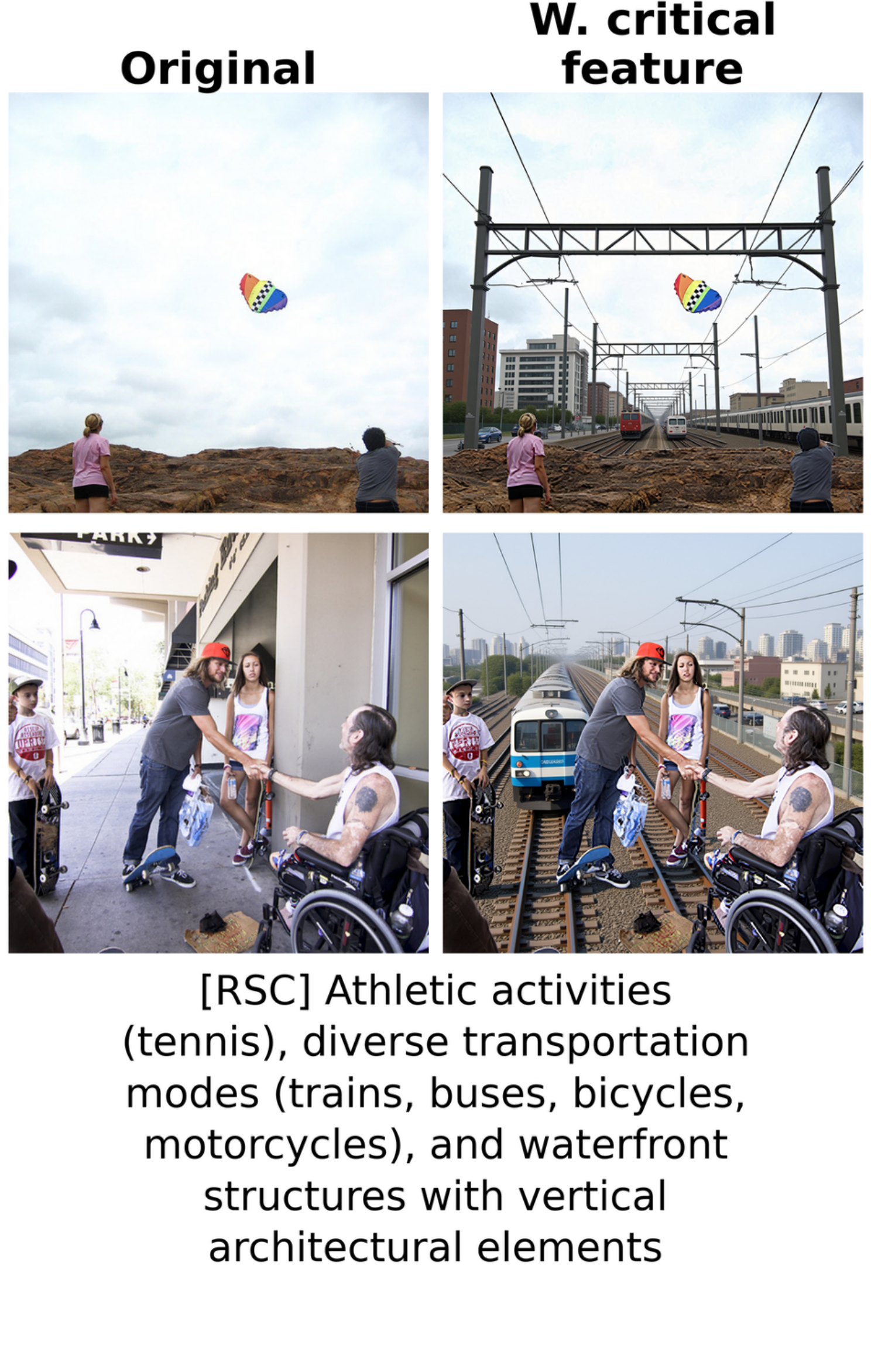}
      \caption{}
      \label{fig:profiles:samples-v18720}
    \end{subfigure}
  \end{minipage}
  \hfill
  \begin{subfigure}[c]{0.52\textwidth}
    \centering
    \includegraphics[width=\textwidth, keepaspectratio=true]{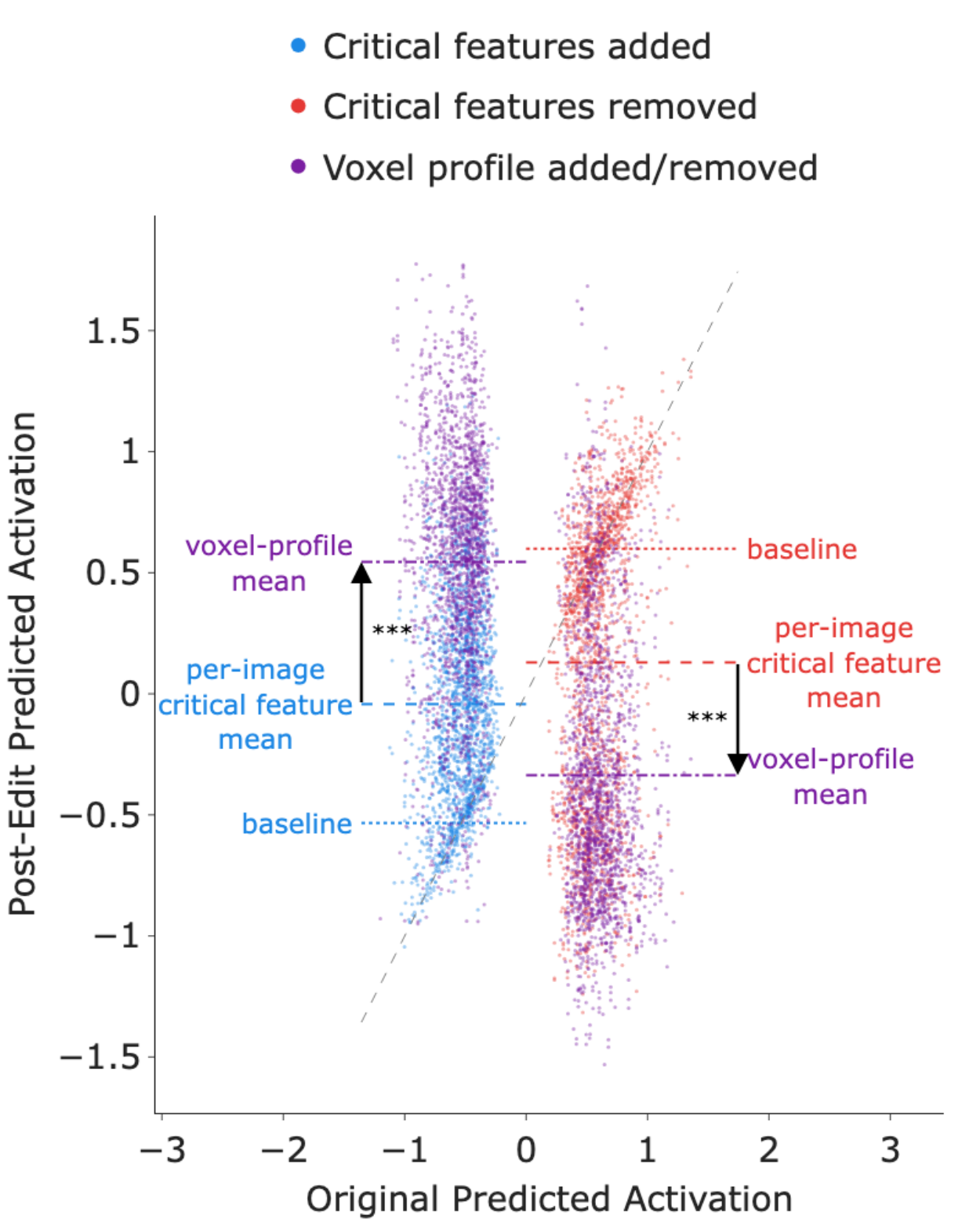}
    \caption{Counterfactual evaluation}
    \label{fig:profiles:counterfactual-performance}
  \end{subfigure}
  \caption{Evaluation of voxel profiles via counterfactual editing. (a--d) Examples of original non-preferred images and counterfactually edited images for four voxels, based on profiles decoded from critical-tokens (left) or from captions (right). (e) Counterfactual evaluation: predicted change in activations by editing according to per-image critical features, compared to editing based on voxel profile. As in Figure~\ref{fig:counterfactual:activation-distributions}, the X-axis indicates the image's initial recorded response, and the Y-axis indicates the predicted response after counterfactual editing. Blue dots show per-image counterfactual editing by adding critical features to non-preferred images. Red dots show per-image counterfactual editing by removing critical features from preferred images. Purple dots show new results from counterfactual editing based on the per-voxel distribution of critical features.}
\end{figure}

Having established our ability to identify critical features in individual images, we turn to aggregating the image-specific features into a general voxel profiles. For each voxel, we aggregated the critical features identified across its highest-faithfulness counterfactual editing trials (see Section~\ref{sec:results:counterfactual}), and passed them to \texttt{Claude-Haiku-4.5} \citep{IntroducingClaudeHaiku} to generate a natural-language description of the voxel's preferred content. Full details of trial selection, vocabulary-token aggregation, weighting, and prompting are provided in Appendix~\ref{app:profile-generation}. To assess whether our discovered critical features yield a faithful voxel-profile hypothesis, we applied counterfactual editing using the voxel profile to the same non-preferred images used in Section~\ref{sec:results:counterfactual}, as exemplified in Figures~\ref{fig:profiles:samples-v6036}--\ref{fig:profiles:samples-v18720}, and measured the predicted change in voxel activation. To assess significance, we fitted a linear mixed model on the change in activation, with a fixed effect for editing type (voxel-profile editing vs. per-image critical-feature editing) and random effects for voxels and non-preferred images. As shown in Figure~\ref{fig:profiles:counterfactual-performance}, voxel-profile editing produced significantly stronger predicted activations, indicating that the voxel profile correctly captures voxel-activating features.

\section{Functional selectivity discovery}
Having established the validity of the MINE framework, we then applied it to identify unique voxel patterns within previously identified functional ROIs. Using the voxel profiles from the evaluation, we attempted to identify shared patterns within functional ROIs and the unique characteristics of individual voxels. First, as shown in Table~\ref{tab:roi-shared-sample}, the shared features in the profiles from pre-defined functional ROIs replicate previous findings \mbox{\citep{kanwisherFusiformFaceArea1997,epsteinCorticalRepresentationLocal1998,priebeSpecificBodyImage2001}}. However, observing the unique features per-voxel, we can identify finer voxel-level patterns that are consistent with the general ROI selectivity. For example, as shown in \ref{tab:roi-unique-sample}, unique voxel features in the Fusiform Face Area (FFA) \citep{kanwisherFusiformFaceArea1997} are all face selective, while each attends to a different property, such as clothes, facial-features, or animal faces. Unique features in the Parahippocampal Place Area (PPA) \citep{epsteinCorticalRepresentationLocal1998} show a different focus for the outdoors and the indoors, corroborated by \citep{wassermanBrainExploreLargeScaleDiscovery2025}, while also providing higher resolution features such as snow and ocean. Unique features in the Extrastriate Body Area (EBA) \cite{downingCorticalAreaSelective2001a} exhibit slightly less structured patterns, yet still fully describe body and activity-related features. Results of sampled voxels across all 14 ROIs are provided in Tables~\ref{tab:roi-shared} and~\ref{tab:roi-unique} in Appendix~\ref{app:discovery}.

\begin{table}[h]
\caption{Representative shared functional selectivity profiles. Full results across all 14 ROIs are reported in Table~\ref{tab:roi-shared} (Appendix~\ref{app:discovery}).}
\label{tab:roi-shared-sample}
\centering
\footnotesize
\begin{tabular}{p{0.13\textwidth} p{0.78\textwidth}}
\toprule
\textbf{ROI} & \textbf{Shared profile} \\
\midrule
Face-selective\newline(FFA-1) & Human faces and upper bodies in professional, formal, or distinctive contexts (business attire, uniforms, sports wear); animals with visible anatomical and facial details; dynamic physical activities and action scenes; sensitivity to clothing, facial features, and compositional structure. \\
\addlinespace
Place-selective\newline(PPA) & Preference for structured, organized environments with clear spatial hierarchies and functional purposes, including both indoor domestic/functional spaces (kitchens, bathrooms, offices) and outdoor scenes with defined architectural or infrastructural elements. \\
\addlinespace
Body-selective\newline(EBA) & Dynamic human figures engaged in athletic activities and sports (particularly water sports like surfing and winter sports like skiing), wearing athletic or specialized gear, often in motion or action contexts. Secondary consistent preference for animals in natural or outdoor settings. \\
\bottomrule
\end{tabular}
\end{table}

\begin{table}[h]
\caption{Representative unique voxel profiles (3 voxels per ROI). Full results across all voxels are reported in Table~\ref{tab:roi-unique} (Appendix~\ref{app:discovery}).}
\label{tab:roi-unique-sample}
\centering
\footnotesize
\begin{tabular}{p{0.10\textwidth} p{0.82\textwidth}}
\toprule
\textbf{ROI} & \textbf{Unique profiles} \\
\midrule
Face-selective\newline(FFA-1) &
\textbullet\ \textit{voxel\_6016}: uniforms (chef, sports, transportation) and people through glass/windows. \newline
\textbullet\ \textit{voxel\_6035}: facial features, hair styling, accessories (glasses, hats); dynamic water activities. \newline
\textbullet\ \textit{voxel\_6063}: close-up animal faces (horses, dogs, cats) and group animal scenes (elephants). \\
\midrule
Place-selective\newline(PPA) &
\textbullet\ \textit{voxel\_8629}: winter sports activities and snowy landscapes alongside kitchen/bathroom fixtures. \newline
\textbullet\ \textit{voxel\_8734}: symmetric, organized arrangements of objects with high-weight feature emphasis across diverse interior environments. \newline
\textbullet\ \textit{voxel\_8862}: water/maritime environments and organized spatial arrangements with utilitarian purposes. \\
\midrule
Body-selective\newline(EBA) &
\textbullet\ \textit{voxel\_2271}: winter sports emphasis with visible athletic gear (jackets, helmets, gloves); upper body visibility and formal authority figures. \newline
\textbullet\ \textit{voxel\_2529}: occupational roles like chefs; group scenes with winter clothing; sensitivity to both body movement and outdoor scenarios. \newline
\textbullet\ \textit{voxel\_3095}: everyday activities and interactions in professional, recreational, and domestic settings; emphasis on human-animal interactions. \\
\bottomrule
\end{tabular}
\end{table}

\section{Conclusions and discussion}
\paragraph{Conclusions} We introduce MINE, an MI-based framework for neural encoders that yields causally validated, highly-detailed functional-selectivity hypotheses at single-voxel resolution. Rather than treating the encoder as a black box, MINE identifies the visual features driving each voxel's predicted response and verifies them through counterfactual image editing. This shifts visual-cortex characterization beyond coarse category labels (e.g., scenes) toward exemplar-level selectivity (e.g., ocean), revealing the heterogeneous, fine-grained preferences individual voxels within a category-selective region.

\paragraph{Limitations and future work}\label{sec:limitations}
First, our analysis is causal at the model level but correlational with respect to the brain; the resulting voxel-profile hypotheses still require validation through neuroimaging experiments, a limitation shared by all model-based brain analyses \citep{matsuyamaLaVCaLLMassistedVisual2025,hwangSilicoMappingVisual2025,wassermanBrainExploreLargeScaleDiscovery2025,luoBrainscubaFinegrainedNatural2023,yangCLIPMSMMultiSemanticMapping2025,luoBrainDiffusionVisual2023,gaoBrainLMMLabelFreeFramework2026,luoBrainMappingDense2025}. Second, image representations are biased toward features that appear most often during training. For our current implementation, the features would be biased towards descriptions of visual-question-answering training (e.g., `person' over `torso' or `neck'). Third, the reliance on generative models makes evaluation stochastic, with errors from decoding only a subset of the vocabulary, from subjectivity in identifying the critical feature, and from hallucinations of the diffusion-model prior. Fourth, we apply a fixed top-50 cutoff on IG scores (Section~\ref{sec:results:hypotheses}), although the optimal number of critical image-tokens likely varies by voxel and image. Future works should study flexible approaches for choosing the number of critical image-tokens. We provide further consideration of positive and negative societal implications of this work in Appendix~\ref{app:social}.

\begin{ack}
This work was supported by a research grant from the Center for Artificial Intelligence and Data Science (TAD) at Tel-Aviv University.
\end{ack}

\newpage

\bibliographystyle{unsrtnat}
\bibliography{references}

\appendix

\section{Social impacts}\label{app:social}
\paragraph{Positive societal implications} Our framework offers concrete positive contributions. Methodologically, it brings mechanistic interpretability tools from machine learning into systems neuroscience, replacing purely correlational characterizations with causally validated voxel-level hypotheses. This raises the standard of evidence in the field and reduces the risk of spurious functional claims. Practically, the ability to generate fine-grained, textually phrased hypotheses about what each voxel responds to could accelerate the basic-science study of high-level visual cortex --- a process that has historically required years of hand-designed stimulus experiments per region. The pipeline is also modality-agnostic: it requires only a stimulus-conditioned neural recording dataset and a pretrained, tokenized encoder of those stimuli, so the same critical-feature extraction and counterfactual-validation procedure naturally extends to other sensory and cognitive cortices --- e.g. audio encoders applied to auditory cortex, or language models applied to cortical language regions. In a clinical direction, more accurate and interpretable encoding models of category-selective regions could in time inform surgical planning, the study of disorders such as prosopagnosia and visual agnosia, and the development of assistive brain--computer interfaces that aim to restore visual function for people with sensory impairments.

\paragraph{Possible negative societal implications} This work is foundational neuroscience research and its potential negative societal impacts are indirect. Improvements in fMRI encoding contribute to a longer-term research trajectory --- including brain--computer interfaces and stimulus decoding --- that raises mental-privacy concerns, though our framework models responses to externally presented images rather than subjective thought. Two interpretive caveats are more proximate: the textual voxel profiles are summarized via an LLM and may inherit its biases, and our analyses cover only four NSD subjects viewing MS-COCO images, so claims phrased about ``the visual cortex'' should be read as hypotheses about this specific sample. Finally, the counterfactual-editing component is in principle dual-use in attention-shaping applications such as advertising, although subject-specific encoders are needed for transfer to real brains. We mitigate these risks by framing all voxel profiles as hypotheses requiring biological validation (Section~\ref{sec:limitations}) and by restricting analyses to the publicly released, consented NSD dataset.

\section{Model training and evaluation}\label{app:model}

\subsection{fMRI dataset and preprocessing}\label{app:model:data}
We train our neural encoder to predict the recorded fMRI responses to the Natural Scenes Dataset (NSD) \citep{allenMassive7TFMRI2022}, a large-scale fMRI dataset of 8 subjects viewing tens of thousands of natural images from the \texttt{MSCOCO} dataset \citep{linMicrosoftCocoCommon2014} across 40 scanning sessions. Following the protocol of \citet{conwellLargescaleExaminationInductive2024}, we use the preprocessed data provided by the NSD team and apply further per-session normalization and train the model to predict the mean response across repetitions of the same image. We focused on a subset of voxels belonging to subjects $1,2,5,7$ with a noise ceiling of at least $0.4$. Due to our goal of identifying textual descriptions of the individual voxels' selectivity, we focus on voxels defined as intermediate visual cortex using the \texttt{streams} categorization, based on the probabilistic atlas of \citet{wangProbabilisticMapsVisual2015}, and on voxels identified as category selective using functional localizers \citep{stiglianiTemporalProcessingCapacity2015}, resulting in an overall $83071$ voxels. For further details on the dataset and preprocessing, see \citet{allenMassive7TFMRI2022}.

\subsection{Image representations}\label{app:model:rep}
We train our model to predict the neural response to images based on representations extracted from LLaVA-1.5-7B \citep{liuImprovedBaselinesVisual2024,liuVisualInstructionTuning2023} VLM, consisting of 576 image-token representations, each of dimension $4096$ (sans the first, \texttt{BOS} token). Specifically, we used the representation from the residual stream \citep{heDeepResidualLearning2016} after the 30th layer, following previous findings by \citet{neoInterpretingVisualInformation2024}, which show that projecting image-token representations from late layers of LLaVA to the vocabulary yields a bag-of-words representation describing the image-token's visual content.
We trained the model on $7200$ unique images per subject, evaluated performance during training on a set of $900$ unique images, and tested the final model on a set of $900$ unique images. All analyses are based on a shared set of $1000$ images shown to all $4$ subjects, which were not used during training.

\subsection{Model architecture and training procedure}\label{app:model:train}
Similarly to the work of \citet{adeliTransformerBrainEncoders2025}, we train a single cross-attention transformer layer \citep{vaswaniAttentionAllYou2017} using mean-squared error (MSE). The model's internal dimension is set to $512$, with $128$ attention heads and prenorm RMS normalization \citep{zhangRootMeanSquare2019,xiongLayerNormalizationTransformer2020}. We used AdamW \citep{loshchilovDecoupledWeightDecay2017} optimization with OneCycle learning rate scheduling \citep{smithSuperconvergenceVeryFast2019} with an initial learning rate of 1e-6 and a maximum learning rate of 1.5e-5. All other optimization parameters are set to their default values as specified in the PyTorch documentation \citep{paszkePyTorchImperativeStyle2019}. The model is trained for 100 epochs, with 1000 steps per epoch. Each step is trained on a batch of $64$ images, optimizing $256$ voxels per image.

\subsection{Model performance}\label{app:model:eval}
\begin{figure}[t]
  \centering
  \begin{subfigure}[b]{0.32\textwidth}
    \centering
    \includegraphics[width=\textwidth]{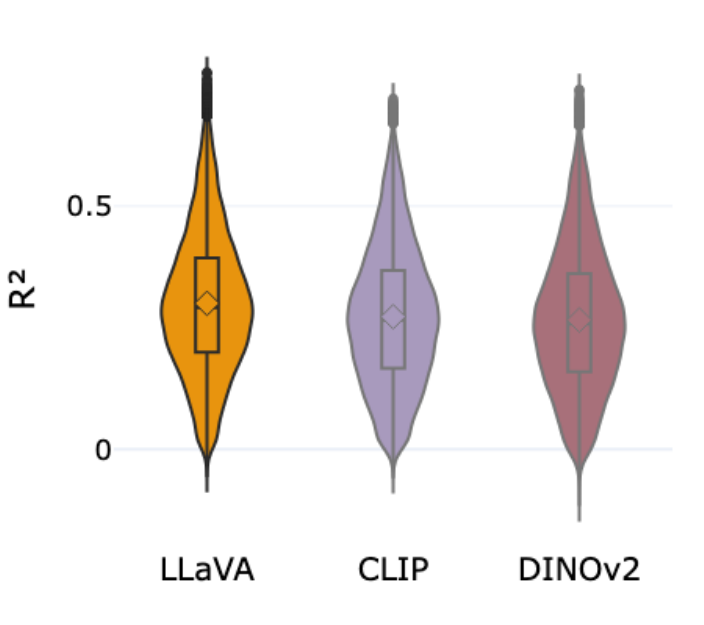}
    \caption{Performance distribution}
    \label{fig:model-perf-violin}
  \end{subfigure}
  \begin{subfigure}[b]{0.32\textwidth}
    \centering
    \includegraphics[width=\textwidth]{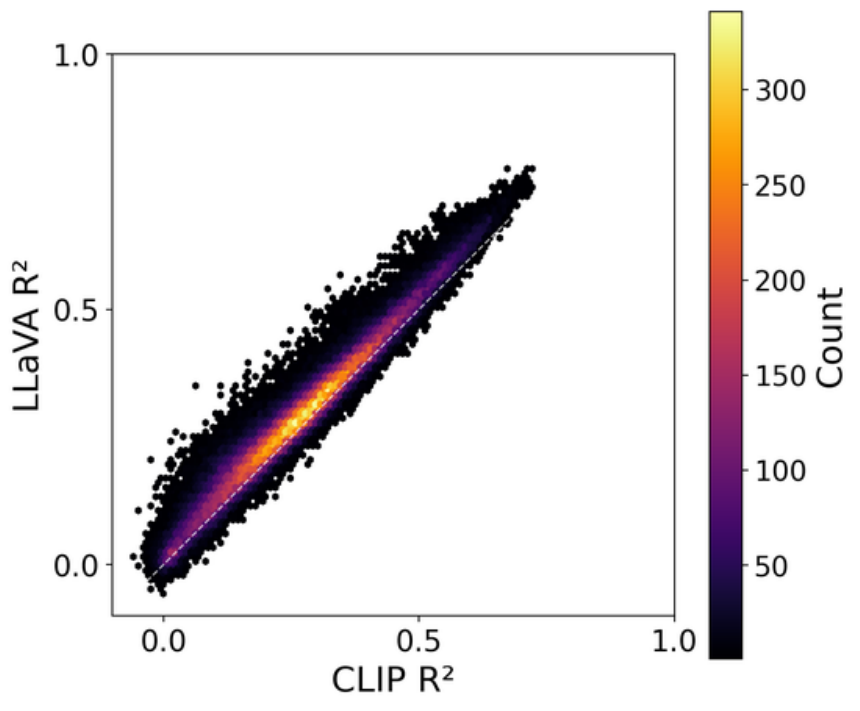}
    \caption{CLIP vs.\ LLaVA}
    \label{fig:model-perf-clip}
  \end{subfigure}
  \hfill
  \begin{subfigure}[b]{0.32\textwidth}
    \centering
    \includegraphics[width=\textwidth]{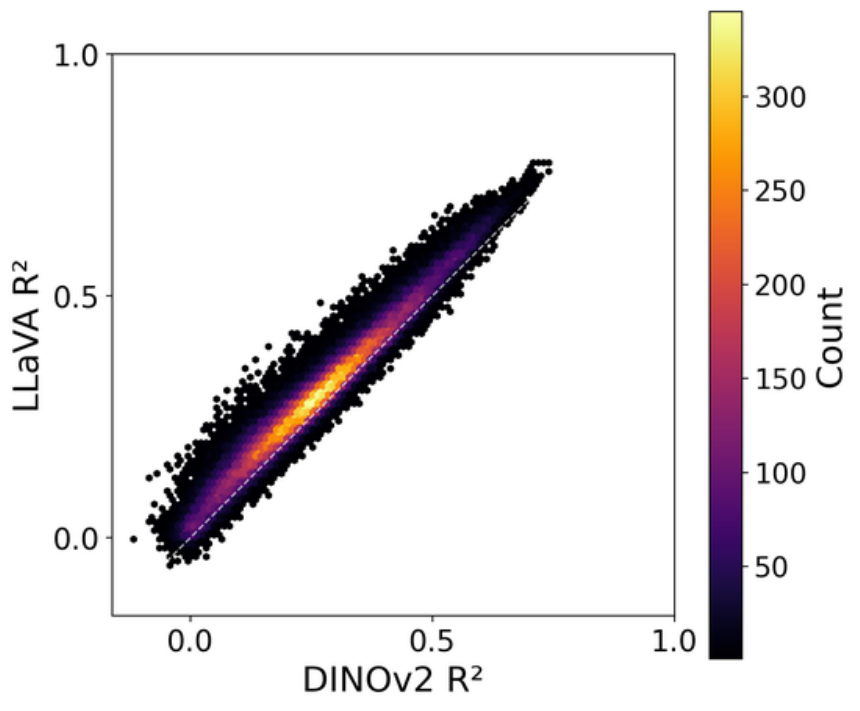}
    \caption{DINOv2 vs.\ LLaVA}
    \label{fig:model-perf-dino}
  \end{subfigure}
  \hfill
  
  \caption{Per-voxel model performance comparison. (a)~Distribution of per-voxel $R^2$ across all three models. The y-axis indicates the voxel's $R^2$. The gold violin shows the distribution of individual voxel $R^2$ values for the LLaVA-based variant. The purple violin shows the distribution for the CLIP-based variant. The red violin shows the distribution for the DINOv2-based variant. (b)~CLIP-based vs.\ LLaVA-based encoder $R^2$. The X-axis indicates CLIP-based $R^2$. Y-axis indicates LLaVA-based $R^2$. Colorbar indicates the density of voxels. Points along the dashed diagonal indicate voxels in which both encoders perform similarly. (c)~DINOv2-based vs.\ LLaVA-based encoder $R^2$. X-axis indicates DINOv2-based $R^2$. Y-axis indicates LLaVA-based $R^2$. Colorbar indicates the density of voxels. Points along the dashed diagonal indicate voxels in which both encoders perform similarly.}
  \label{fig:model-performance}
\end{figure}

For the discovered hypotheses to be valid, the model must accurately predict the neural response. We therefore compare the performance of our models, previously validated as SOTA for predicting neural responses on the same dataset \citep{adeliTransformerBrainEncoders2025,beliyWisdomCrowdBrains2025}. We train two further models using the same architecture and training procedure, but using as input the image-token features extracted from the final layers of CLIP \citep{radfordLearningTransferableVisual2021} and DINOv2 \citep{oquabDINOv2LearningRobust2023}. As shown in Figure~\ref{fig:model-perf-violin}, we find that our model performs on par with a transformer based on CLIP and DINOv2 representation. Furthermore, as shown in Figure~\ref{fig:model-perf-clip} and Figure~\ref{fig:model-perf-dino}, when comparing the performance per individual voxel, it is evident that the LLaVA-based variant has a slight advantage over the CLIP and DINOv2 variants, consistent with previous results \citep{bavarescoVisionLanguageModelsAlign2026}. To compare with previous results, we also report the voxel prediction accuracy, defined as the per-voxel $R^2$ normalized by the noise-ceiling \citep{adeliTransformerBrainEncoders2025}. The Per-subject and overall prediction accuracy values for the three encoders are reported in Table~\ref{tab:model-perf}. Comparing with previous SOTA results, we can that our neural encoder performs superiorly when predicting the responses of subjects S5 and S7, while previous results achieve better results on subjects S1 and S2. While we are unable to methodically compare our results due to missing details for voxel exclusion, we argue that our model's performance is high enough comparable to SOTA, making the approximate hypotheses learned by the model viable.

\begin{table}[t]
\centering
\caption{Per-subject and overall predictive performance for the three encoder variants. The first row shows the number of analyzed voxels per subject; the Total column reports the voxel-weighted mean across subjects.}
\label{tab:model-perf}
\begin{tabular}{lrrrrr}
\toprule
 & S1 & S2 & S5 & S7 & Total \\
\midrule
Number of voxels    & 21{,}334 & 25{,}565 & 26{,}125 & 10{,}047 & 83{,}071 \\
\midrule
DINOv2 & 0.433 & 0.460 & 0.501 & 0.519 & 0.473 \\
CLIP   & 0.445 & 0.474 & 0.515 & 0.534 & 0.487 \\
LLaVA  & 0.500 & 0.519 & 0.571 & 0.577 & 0.537 \\
\bottomrule
\end{tabular}
\end{table}

\subsection{Inherent limitation of captions in the prediction of the visual cortex}\label{app:captions}
\begin{figure}[t]
  \centering
  \includegraphics[width=0.5\textwidth, keepaspectratio=true]{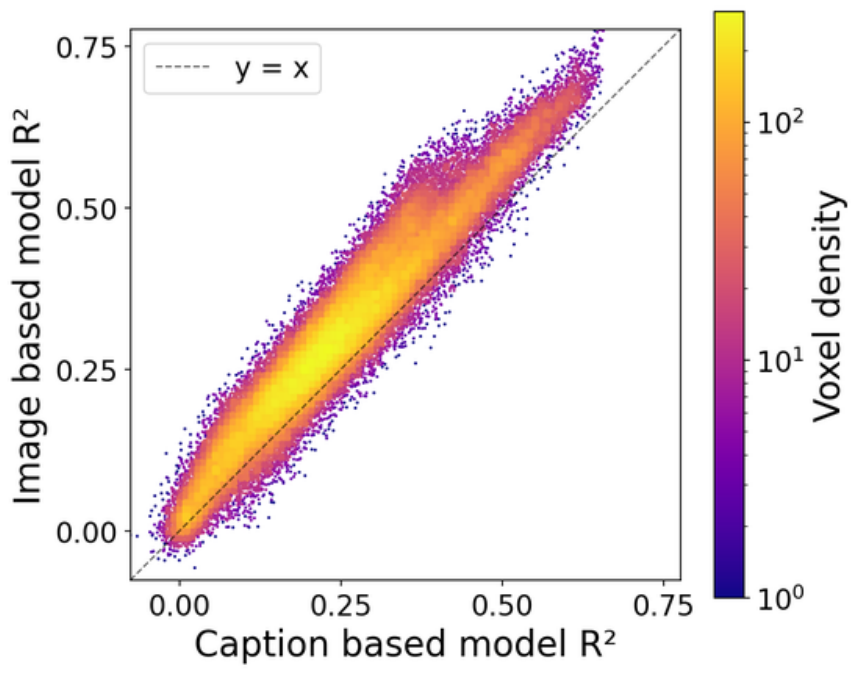}
  \caption{Per-voxel comparison of explained variance ($R^2$) between the LLaVA-based image encoder and the OLMo-based caption encoder. The X-axis shows the voxels' $R^2$ when predicted using the caption embeddings. The y-axis shows the voxels' $R^2$ when predicted using the image embeddings. The color scale indicates voxel density based on these $R^2$ values. The dashed line indicates voxels with the same $R^2$ when predicted from captions as from images.}
  \label{fig:captions:r2-comparison}
\end{figure}

A core component in previous works is the reliance on general-purpose image captions, either by directly captioning activating images \citep{matsuyamaLaVCaLLMassistedVisual2025,hwangSilicoMappingVisual2025,wassermanBrainExploreLargeScaleDiscovery2025,gaoBrainLMMLabelFreeFramework2026}, or due to reliance on CLIP's language alignment \citep{luoBrainscubaFinegrainedNatural2023,yuMetaLearningInContextTransformer2025a,yangCLIPMSMMultiSemanticMapping2025}, which was shown to lead to encodings with shallow visual information \citep{urbanekPictureWorthMore2024}. In this section, we show the inherent limitation of using general-purpose image captions to identify the underlying features that predict the visual cortex. Specifically, \citet{shohamHighlevelVisualCortex2025} showed that the predictive power of LLM embeddings for visual-cortex activity hinges on the degree of visual detail they carry. Since general-purpose captions tend to be visually shallow, we hypothesized they would underperform image-based encoders. To test this, we trained a second neural encoder to predict the fMRI response to each image from its caption embedding, using the same hyperparameters as in Section~\ref{sec:results:setting}. Caption embeddings were extracted from an intermediate layer (layer 16 out of 32) of the pretrained LLM OLMo-2-7B \citep{olmo2OLMo22025}, applied to MSCOCO captions \citep{linMicrosoftCocoCommon2014}. We then compared per-voxel $R^2$ between the two encoders. Figure~\ref{fig:captions:r2-comparison} compares the models' $R^2$. The LLaVA-based encoder significantly outperforms the OLMo-based encoder ($t(83{,}070)=400.89$, $p<0.001$), with an average advantage of $5.5\%$ in explained variance (95\% CI = [$5.49\%$, $5.55\%$]; $d_z=1.39$). This shows that while captions carry some predictive signal, they fall meaningfully short of image-based features, motivating the use of vision-language model representations as a richer source of visual information.

\section{Compute resources and implementation details}\label{app:compute}
All experiments were conducted using a single A-100 or H-100 GPUs, with at most 256GB of RAM, and 8 CPU cores. Each model training took between 48-72 hours. Model training, evaluation, and analysis was implemented using PyTorch \citep{paszkePyTorchImperativeStyle2019}. To extract image representations from pretrained models, we use the \texttt{HuggingFace Transformers} library \citep{wolfHuggingfacesTransformersStateoftheart2019}. For image generation and editing we used the \texttt{HuggingFace Diffusers} library \citep{von-platen-etal-2022-diffusers}. To query \texttt{Claude-Haiku-4.5} \citep{IntroducingClaudeHaiku}, we used the \texttt{DSPy} library \citep{khattab2022demonstrate,khattab2024dspy}. Figures were generated using the \texttt{MatPlotLib} library \citep{Hunter:2007} and \texttt{Plotly} \citep{incCollaborativeDataScience2015}

\section{Framework evaluation detailes}
\subsection{Evaluating tokens importance}\label{app:token-patching}
\begin{figure}[t]
  \centering
  \begin{subfigure}[b]{0.49\textwidth}
    \centering
    \includegraphics[width=\textwidth]{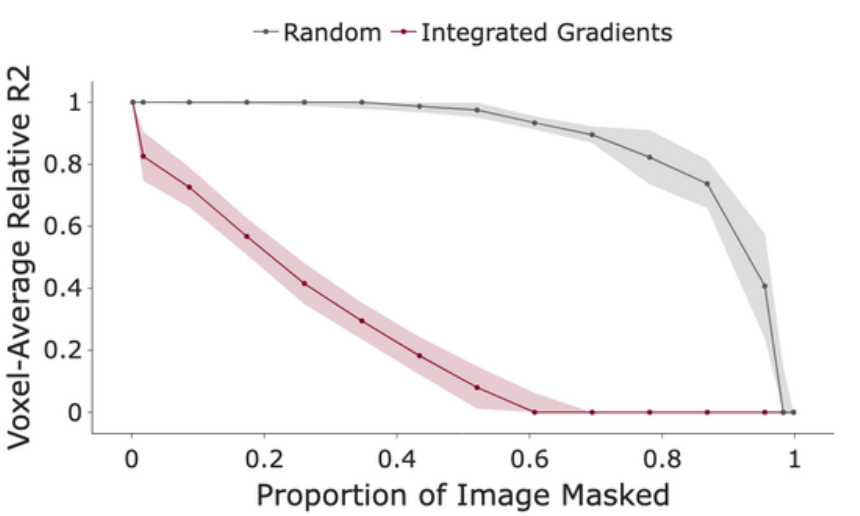}
    \caption{Necessary patching results}
    \label{fig:necessary-results}
  \end{subfigure}
  \hfill
  \begin{subfigure}[b]{0.49\textwidth}
    \centering
    \includegraphics[width=\textwidth]{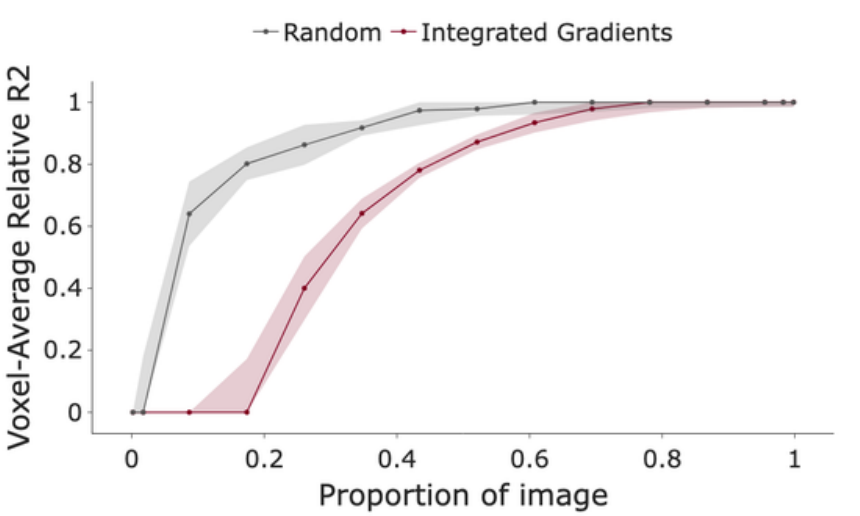}
    \caption{Sufficient patching results}
    \label{fig:sufficient-results}
  \end{subfigure}
  \caption{Results for the mean-patching analysis, when targeting $K$ random image-tokens (grey) or $K$ image-tokens with the highest IG score (red). (a) Relative model performance after patching the most critical tokens. The y-axis indicates the relative performance of the model compared to the full model. The X-axis indicates the ratio of image tokens being replaced. Lower values indicate greater dependence on the selected image tokens. (b) Relative model performance after keeping only the most critical tokens. The y-axis indicates the relative performance of the model compared to the full model. The X-axis indicates the ratio of image tokens being replaced. Higher values indicate the remaining image tokens are sufficient to predict the fMRI recording.}
\end{figure}

To identify critical image-tokens according to importance in the prediction of neural response, we used the IG method \citep{sundararajanAxiomaticAttributionDeep2017} to rank the contribution using a gradient-based method. Similar to \citet{neoInterpretingVisualInformation2024}, we use a mean image-token from the \texttt{ImageNet} validation set \citep{russakovskyImageNetLargeScale2014}, and approximate the integrated gradients between the mean image-token and each image representation along 10 steps. We apply this analysis to each voxel and each image from a set of $1{,}000$ held-out images. To validate the importance of said images, we apply representation-patching \citep{mengLocatingEditingFactual2022} over the top K most important image-tokens, replacing image-tokens with the same ImageNet average image-token representation \citep{zhangBestPracticesActivation2024} and compare them with patching randomly chosen image-tokens to control for the effect of random removal of image-tokens. First, we evaluate if the discovered patches are necessary by removing the top K image-tokens with highest IG score. Second, we assess the sufficiency of the image-tokens by using only the top K IG-score image-tokens to predict voxel activations. Here we report the proportion of explained variance by the model after applying patching, relative the full model results $\bar{R}_k^2 / \bar{R}^2$.
As shown in Figure~\ref{fig:necessary-results}, the IG method discovers image-tokens that are critically necessary for the correct voxel prediction, with $\bar{R}^2$ dropping by 27.4\% in model's performance after removing 50 image-tokens, indicating that these image-tokens are good candidates to generate voxel selectivity hypotheses. Somewhat conflicting results are observed in Figure~\ref{fig:sufficient-results}, showing that predicting voxel activation from randomly selected image-tokens yields more accurate predictions than from image-tokens selected by their IG score. We hypothesize that this could be attributed to random selection of image-tokens, which provides a low-resolution approximation of the image that is more faithful to the complete image than related image-tokens that convey a single feature. Although we do not directly test this hypothesis, further results from Section~\ref{sec:results:reconstruction} and Section~\ref{sec:results:counterfactual} show that the content encoded in these image-tokens is indeed sufficient to change the voxel prediction of the image, supporting the findings in \ref{fig:sufficient-results}.

\subsection{Artificial neurons validation}\label{app:artificial-neurons}
A fundamental building block of our framework is the generation of a textual description from the critical image representations. However, our work focuses on the discovery of the functional selectivity of previously unknown computational units; thus, the correctness of this step relies on further validation steps with inherent error rates (e.g., errors in image generation). To validate our method for hypothesis generation, we apply our method to decode the functional selectivity patterns of predetermined artificial neurons. Specifically, we trained our model with the same hyperparameters as described in Appendix~\ref{app:model:train}, to predict class labels for images from the \texttt{CIFAR-10} dataset \citep{Krizhevsky09learningmultiple}. To adapt the model to binary target labels, we trained it to minimize the Binary Cross-Entropy loss between the predicted and correct class probabilities. We trained the model using the full 50,000-image training split and evaluated it using the 10,000-image validation split. The trained model reached a perfect validation accuracy of 100\% on the validation split.
Due to the relatively simple nature of \texttt{CIFAR-10} images, where each image is dominated by a single foreground object, we perform the decoding analysis on a subset of images from the MSCOCO dataset \citep{linMicrosoftCocoCommon2014}, where at least a single \texttt{CIFAR-10} object is present. As two \texttt{CIFAR-10} classes (deer, frog) are not present in the MSCOCO dataset, we performed this analysis on the remaining 8 classes. For hypothesis generation, we filtered the MSCOCO images to those that the model correctly identified as containing the specific object. We then analyzed 10 images per neuron using the IG algorithm \citep{sundararajanAxiomaticAttributionDeep2017} to identify the top 50 most important image tokens and project them into the vocabulary space, retrieving the top 30 image tokens with the highest logits. Finally, we passed the list of most important image tokens to Claude-Sonnet-4.5 \citep{anthropic_claude_sonnet_4.5} to encode each image independently, and prompted it with the instructions in Figure~\ref{fig:image-prompt} to decode the image content. All 10 image descriptions were then passed to Claude-Sonnet-4.5 using the prompt in Figure~\ref{fig:neuron-prompt} to identify the shared information for the neuron. As shown in Figure~\ref{fig:text-boxes}, our approach correctly identified all 8 classes and also revealed correlated patterns.

\tcbset{
  cifarbox/.style={
    colback=gray!8,
    colframe=black!60,
    colbacktitle=gray!20,
    coltitle=black,
    boxrule=0.4pt,
    arc=4pt,
    left=8pt, right=8pt, top=4pt, bottom=6pt,
    toptitle=3pt, bottomtitle=3pt,
    fonttitle=\small\bfseries
  }
}

\captionsetup[figure]{hypcap=false}

\begin{figure}[t]
\centering
\begin{tcolorbox}[cifarbox, notitle]
\small
You are analyzing tokens extracted from the top-K most important image patches for predicting a neuron activation in a neural network.\\[4pt]
These tokens represent what visual features were most critical for the classification. Based on the following token list, provide a 2-3 sentence description of what visual elements likely drove the prediction.\\[4pt]
Focus on:\\
\textbullet\ Objects, textures, or patterns that appear frequently\\
\textbullet\ Spatial relationships or scene context\\
\textbullet\ Colors or visual attributes\\[4pt]
Do not mention that you are analyzing tokens - just describe what visual content the image likely contains based on these clues.
\end{tcolorbox}
\caption{Prompt for image level analysis.}
\label{fig:image-prompt}
\end{figure}

\vspace{1em}

\begin{figure}[t]
\centering
\begin{tcolorbox}[cifarbox, notitle]
\small
You are analyzing the role of a classification neuron in a vision model.\\[4pt]
Below are 10 descriptions of critical visual features from different correctly classified images:\\[4pt]
\dots\\[4pt]
Based on these descriptions, identify the shared underlying visual information that this neuron appears to encode. What specific features, patterns, or concepts does this neuron detect?\\[4pt]
Provide a 1-sentence analysis that identifies the common visual elements across all descriptions
\end{tcolorbox}
\caption{Prompt for neuron level analysis.}
\label{fig:neuron-prompt}
\end{figure}

\vspace{1em}

\begin{figure}[t]
\centering
\begin{tcolorbox}[cifarbox, title=Airplane]
\small
This neuron detects \textbf{commercial aircraft in airport ground environments}, consistently activating for images containing planes (typically white/commercial airliners) positioned on tarmacs, runways, or taxiways, often with associated ground support infrastructure like cargo equipment, service vehicles, and airport markings.
\end{tcolorbox}

\vspace{4pt}
\begin{tcolorbox}[cifarbox, title=Automobile]
\small
This neuron appears to detect \textbf{vehicles and automotive-related objects}, consistently activating for images containing cars, trucks, SUVs, motorcycles, taxis, and their associated components (mirrors, windows, hoods) as well as vehicle-adjacent infrastructure (parking meters, street signs, poles).
\end{tcolorbox}

\vspace{4pt}
\begin{tcolorbox}[cifarbox, title=Bird]
\small
This neuron appears to detect \textbf{sky-related visual features or upward-viewing perspectives}, as evidenced by the consistent presence of elements photographed from below or against the sky (birds in flight, clock towers, boat masts, railings) and outdoor scenes with aerial or elevated components.
\end{tcolorbox}

\vspace{4pt}
\begin{tcolorbox}[cifarbox, title=Cat]
\small
This neuron detects \textbf{cats} in indoor/enclosed domestic settings, particularly when positioned near, on, or interacting with household objects, barriers, or structures (such as windows, beds, cars, doors, and furniture).
\end{tcolorbox}

\vspace{4pt}
\begin{tcolorbox}[cifarbox, title=Dog]
\small
This neuron detects the presence of \textbf{dogs} in various settings and contexts, consistently activating regardless of the dog's breed, color, activity, or surrounding environment.
\end{tcolorbox}

\vspace{4pt}
\begin{tcolorbox}[cifarbox, title=Horse]
\small
This neuron detects the presence of \textbf{horses} in various contexts, consistently activating when equine animals appear in images regardless of whether they are working draft animals, show horses, police mounts, or in pastoral settings.
\end{tcolorbox}

\vspace{4pt}
\begin{tcolorbox}[cifarbox, title=Ship]
\small
This neuron detects \textbf{boats and watercraft in maritime settings}, consistently activating for scenes containing various types of vessels (sailboats, motorboats, rowboats, ships) moored at or near docks, harbors, and waterfront locations.
\end{tcolorbox}

\vspace{4pt}
\begin{tcolorbox}[cifarbox, title=Truck]
\small
This neuron appears to detect \textbf{vehicles and transportation infrastructure in urban street settings}, consistently activating for images containing cars, buses, trucks, trains, and their associated infrastructure elements like traffic signals, power lines, road markings, and street furniture.
\end{tcolorbox}

\caption{Decoded descriptions for eight CIFAR-10 classes.}
\label{fig:text-boxes}
\end{figure}

\subsection{Critical feature decoding}\label{app:feature-decoding}
In this section, we describe the process of decoding the critical feature for each individual image. First, we selected 50 image-tokens for decoding, either based on IG scores or randomly. Afterward, we decoded each image token using the \texttt{logit-lens} approach and retained the top 10 vocabulary tokens with the highest logit values. We then aggregated all tokens and passed them to an LLM for decoding. The prompt passed to the LLM is shown in Figure~\ref{app:fig:feature-decoding-prompt}. To generate a description based on \texttt{MSCOCO} captions \citep{linMicrosoftCocoCommon2014}, the captions are passed using the system prompt shown in Figure~\ref{app:fig:caption-decoding-prompt}

\begin{figure}[t]
\centering
\begin{tcolorbox}[cifarbox, notitle, breakable, enhanced jigsaw]
\begin{Verbatim}[fontsize=\footnotesize]
Your input fields are:
1. `token_table` (str): Markdown table of decoded tokens with columns: Rank,
   Word, Mean Logit, Count. Higher Mean Logit indicates stronger association
   with voxel activation.
Your output fields are:
1. `description` (ImageContentDescription): Structured description of the
   inferred visual content. Include primary content, scene context, visual
   attributes, and semantic categories.
All interactions will be structured in the following way, with the
appropriate values filled in.
[[ ## token_table ## ]]
{token_table}
[[ ## description ## ]]
{description}        # note: the value you produce must adhere to the JSON schema:
{"type": "object",
 "properties": {
   "primary_content": {"type": "string",
     "description": "Main visual element or subject in the image
       (1-2 sentences)"},
   "scene_context": {"type": "string",
     "description": "Scene type and spatial context (indoor/outdoor,
       setting)"},
   "visual_attributes": {"type": "array", "items": {"type": "string"},
     "description": "List of visual attributes (colors, textures,
       patterns)"},
   "semantic_categories": {"type": "array", "items": {"type": "string"},
     "description": "High-level categories (faces, places, objects,
       text, animals)"},
   "confidence_note": {"type": "string",
     "description": "Any uncertainty or ambiguity in the
       interpretation"}
 },
 "required": ["primary_content", "scene_context"]}
[[ ## completed ## ]]
In adhering to this structure, your objective is:
        Describe visual content from decoded vocabulary tokens.
        Given a table of decoded words with their frequencies from
        top-K most important image patches for predicting an fMRI voxel's
        response, describe what visual content likely activated this brain region.
        The token table contains:
        - Word: The decoded vocabulary token
        - Mean Logit: Average logit value (higher = more associated with activation)
        - Count: Number of occurrences
        Focus on inferring visual content (objects, scenes, textures, colors)
        rather than discussing the tokens themselves.
\end{Verbatim}
\end{tcolorbox}
\caption{Prompt for feature-decoding.}
\label{app:fig:feature-decoding-prompt}
\end{figure}

\begin{figure}[t]
\centering
\begin{tcolorbox}[cifarbox, notitle, breakable, enhanced jigsaw]
\begin{Verbatim}[fontsize=\footnotesize]
Your input fields are:
1. `captions` (str): Human-annotated MSCOCO captions for an image, one per
   line. Each caption describes the same image from a different annotator's
   perspective.
Your output fields are:
1. `description` (ImageContentDescription): Structured description synthesized
   from the captions. Include primary content, scene context, visual
   attributes, and semantic categories.
All interactions will be structured in the following way, with the
appropriate values filled in.
[[ ## captions ## ]]
{captions}
[[ ## description ## ]]
{description}        # note: the value you produce must adhere to the same
                     # JSON schema as Section A.
[[ ## completed ## ]]
In adhering to this structure, your objective is:
        Describe visual content from MSCOCO human-annotated image captions.
        Given multiple human captions describing the same image, produce a
        structured description of the visual content. The captions are
        ground-truth annotations from MSCOCO and describe what is actually
        visible in the image.
        Synthesize information across all captions to produce a comprehensive,
        structured description.
\end{Verbatim}
\end{tcolorbox}
\caption{Prompt for caption-based decoding (control baseline).}
\label{app:fig:caption-decoding-prompt}
\end{figure}

\subsection{Voxel profile generation}\label{app:profile-generation}
For each voxel, we collected all descriptions from the counterfactual image-editing pipeline (see Section~\ref{sec:results:counterfactual}) whose faithfulness score lay in the upper quartile of that voxel's faithfulness distribution and required a minimum of three surviving trials per voxel. This yielded 225 voxel-subject profiles, each supported by a mean of 175 trials (range 79-650, median 159), drawn from a mean of 40 distinct images (range 25-54, median 43). For each surviving trial we extracted a token list from one of two sources, evaluated as separate variants. In the critical-features variant, tokens were the LLM-identified critical descriptors stored on each row of the upstream counterfactual editing pipeline. In the control, caption-based, variant, tokens were the whitespace-tokenized concatenation of the five human captions associated with the trial's source image. Token lists from all surviving trials of a voxel were collapsed into a frequency-ordered bag and weighted by the mean faithfulness of the trials in which each token appeared. This bag, the corresponding weights, and up to five free-text descriptions of contributing images were passed to \texttt{Claude Haiku 4.5} \citep{IntroducingClaudeHaiku} via a structured prompt requesting (i) a one- to three-sentence natural-language description of the voxel's preferred content and (ii) a deduplicated list of 3-10 normalized concept-level features. The system prompt passed to the LLM is shown in Figure~\ref{app:fig:profile-prompt}.

\begin{figure}[t]
\centering
\begin{tcolorbox}[cifarbox, notitle, breakable, enhanced jigsaw]
\begin{Verbatim}[fontsize=\footnotesize]
Your input fields are:
1. `critical_features` (str): Comma-separated raw critical features observed
   across successful counterfactual edits, ordered by frequency (most common first).
2. `feature_weights` (str): Comma-separated mean-faithfulness weights aligned
   with ``critical_features`` (higher = more reliable).
Your output fields are:
1. `result` (VoxelProfileOutput): Structured voxel profile containing a
   natural-language description and a canonical-features list.
All interactions will be structured in the following way, with the
appropriate values filled in.
[[ ## critical_features ## ]]
{critical_features}
[[ ## feature_weights ## ]]
{feature_weights}
[[ ## result ## ]]
{result}        # note: the value you produce must adhere to the JSON schema:
{"type": "object",
 "properties": {
   "profile_description": {"type": "string",
     "description": "A concise 1-3 sentence description of what visual or
       semantic concepts this voxel appears to prefer, written in natural
       language."},
   "canonical_features": {"type": "array", "items": {"type": "string"},
     "description": "A deduplicated, concept-level list of 3-10 normalized
       feature phrases representing the voxel's preference. These will be used
       as target features for counterfactual editing downstream."}
 },
 "required": ["profile_description", "canonical_features"]}
[[ ## completed ## ]]
In adhering to this structure, your objective is:
        Aggregate noisy per-edit critical features into a coherent voxel profile.
        You are given a list of *critical features* that were observed across
        many successful counterfactual edits on a single fMRI voxel's predicted
        response.  Each feature comes with a weight (mean faithfulness of the
        edits it appeared in -- higher = more reliable signal).  You also get a
        small sample of free-text source captions/descriptions for disambiguation.
        The raw features are NOISY.  They may be:
        - Partial words from BPE tokenizers (e.g. "orsche" -> "Porsche")
        - Concatenated token fragments that should be split
        - Synonyms or near-duplicates that should be merged
        - Occasionally pure noise that should be discarded
        Produce:
        1. A concise natural-language ``profile_description`` summarizing what
           this voxel seems to care about, grounded in the features and contexts.
        2. A clean ``canonical_features`` list of 3-10 normalized concepts.
\end{Verbatim}
\end{tcolorbox}
\caption{Prompt for voxel profile generation.}
\label{app:fig:profile-prompt}
\end{figure}

\section{Voxel profile discovery}\label{app:discovery}

The 14 functional regions of interest (ROIs) used in Tables~\ref{tab:roi-shared} and~\ref{tab:roi-unique} are: \textbf{FFA-1, FFA-2} (Fusiform Face Area, two sub-regions), \textbf{OFA} (Occipital Face Area), \textbf{aTL-faces} (anterior Temporal Lobe, face-selective); \textbf{OPA} (Occipital Place Area), \textbf{PPA} (Parahippocampal Place Area), \textbf{RSC} (Retrosplenial Cortex); \textbf{EBA} (Extrastriate Body Area), \textbf{FBA-1, FBA-2} (Fusiform Body Area, two sub-regions); \textbf{VWFA-1, VWFA-2} (Visual Word Form Area, two sub-regions), \textbf{mTL-words} (medial Temporal Lobe, word-selective), and \textbf{mfs-words} (mid-fusiform sulcus, word-selective).

\begin{figure}[t]
\centering
\begin{tcolorbox}[cifarbox, notitle, breakable, enhanced jigsaw]
\begin{Verbatim}[fontsize=\footnotesize]
Your input fields are:
1. `voxel_descriptions` (list[str]): Profile descriptions for voxels in
   one ROI, indexed by position 0..N-1.
Your output fields are:
1. `shared_content` (str): Visual patterns/content present across all
   voxel descriptions.
2. `unique_content` (list[str]): Voxel-specific patterns NOT shared by
   all. MUST be the same length and order as voxel_descriptions;
   entry i corresponds to the i-th voxel.
All interactions will be structured in the following way, with the
appropriate values filled in.

[[ ## voxel_descriptions ## ]]
{voxel_descriptions}

[[ ## shared_content ## ]]
{shared_content}

[[ ## unique_content ## ]]
{unique_content}        # note: the value you produce must adhere to
                        # the JSON schema: {"type": "array",
                        # "items": {"type": "string"}}

[[ ## completed ## ]]
In adhering to this structure, your objective is:
        Identify visual content shared across all voxels vs. unique
        to each voxel within an ROI.

        Given an ordered list of voxel profile descriptions for one
        ROI, return:
          - shared_content: a single description capturing visual
            patterns present in ALL voxels.
          - unique_content: one short description per voxel listing
            patterns that voxel has but are NOT shared by all voxels
            in the group.
\end{Verbatim}
\end{tcolorbox}
\caption{Prompt used to identify shared and unique voxel profile content per ROI.}
\label{app:fig:roi-summary-prompt}
\end{figure}

{\footnotesize
\begin{longtable}{p{0.14\textwidth} p{0.10\textwidth} p{0.67\textwidth}}
\caption{Shared functional selectivity profile common across voxels within each ROI.}
\label{tab:roi-shared} \\
\toprule
\textbf{Selectivity} & \textbf{ROI} & \textbf{Shared profile} \\
\midrule
\endfirsthead

\multicolumn{3}{l}{\textit{Table~\ref{tab:roi-shared} continued from previous page}} \\
\toprule
\textbf{Selectivity} & \textbf{ROI} & \textbf{Shared profile} \\
\midrule
\endhead

\midrule
\multicolumn{3}{r}{\textit{Continued on next page}} \\
\endfoot

\bottomrule
\endlastfoot

Face & FFA-1 & Human faces and upper bodies in professional, formal, or distinctive contexts (business attire, uniforms, sports wear); animals with visible anatomical and facial details; dynamic physical activities and action scenes; sensitivity to clothing, facial features, and compositional structure. \\
\cmidrule{2-3}
 & FFA-2 & Strong responsiveness to human figures and animals in dynamic or characteristic contexts, with sensitivity to distinctive visual features including clothing, body positioning, and facial/anatomical details. Consistent activation for active subjects (athletes, people in motion, wildlife) and outdoor or natural settings. \\
\cmidrule{2-3}
 & OFA & Dynamic human figures engaged in physical activities, sports, and movement across diverse contexts (skiing, baseball, skateboarding, tennis, surfing), combined with sensitivity to athletic/specialized gear and apparel, as well as animals in motion and outdoor/active scenarios. \\
\cmidrule{2-3}
 & aTL-faces & All voxels show strong responsiveness to human subjects with attention to facial features, distinctive clothing details, and contextual settings. Secondary responsiveness to animals (particularly mammals like dogs) with emphasis on facial features and distinctive visual characteristics is present across voxels. Sensitivity to both portrait/close-up compositions and full-body depictions in various contexts (professional, athletic, formal) is shared. \\
\midrule
Place & OPA & Strong preference for structured, functional indoor spaces (particularly kitchens and bathrooms with fixtures), organized domestic interiors with clear architectural elements, and human-made environments with purposeful design and infrastructure. \\
\cmidrule{2-3}
 & PPA & Preference for structured, organized environments with clear spatial hierarchies and functional purposes, including both indoor domestic/functional spaces (kitchens, bathrooms, offices) and outdoor scenes with defined architectural or infrastructural elements. \\
\cmidrule{2-3}
 & RSC & Structured environments with organized spatial elements, human figures in various contexts, and architectural or infrastructural frameworks. All voxels show sensitivity to both indoor functional spaces and outdoor scenes with clear organizational or structural components. \\
\midrule
Body & EBA & Dynamic human figures engaged in athletic activities and sports (particularly water sports like surfing and winter sports like skiing), wearing athletic or specialized gear, often in motion or action contexts. Secondary consistent preference for animals in natural or outdoor settings. \\
\cmidrule{2-3}
 & FBA-1 & Strong preference for human subjects in distinctive contexts (professional attire, sports activities, formal settings) and animals with visible anatomical or facial details. Consistent sensitivity to portraits, close-up facial features, and subjects engaged in identifiable roles or actions. \\
\cmidrule{2-3}
 & FBA-2 & Strong responsiveness to human and animal subjects in dynamic or formal contexts, with consistent sensitivity to facial features, distinctive clothing/attire, and outdoor or athletic settings. Activation across diverse subjects including people engaged in sports (skiing, surfing, tennis), wildlife (elephants, polar bears, cattle), and individuals in formal or winter wear. \\
\midrule
Word & VWFA-1 & Strong preference for human subjects across professional, formal, and athletic contexts. Consistent sensitivity to facial features, clothing details (particularly formal wear like suits and uniforms), and human upper bodies/portraits. Secondary responsiveness to animals with prominent facial features. \\
\cmidrule{2-3}
 & VWFA-2 & Dynamic action and athletic activities with people in motion, wearing sport-specific attire and protective gear, engaged in physical movement and sports contexts. \\
\cmidrule{2-3}
 & mTL-words & All voxels show strong responsiveness to human subjects and animals with emphasis on facial features, distinctive visual characteristics, and contextual details. There is consistent sensitivity to both people and animals across various compositions and attire types. \\
\cmidrule{2-3}
 & mfs-words & Close-up facial imagery and anatomical details of both humans and animals, with sensitivity to portraits, faces with distinct features, and animals depicted with visible body parts and physical characteristics. \\

\end{longtable}
}

{\footnotesize
\begin{longtable}{p{0.14\textwidth} p{0.10\textwidth} p{0.68\textwidth}}
\caption{Distinctive feature preferences of individual voxels within each ROI.}
\label{tab:roi-unique} \\
\toprule
\textbf{Selectivity} & \textbf{ROI} & \textbf{Unique profiles} \\
\midrule
\endfirsthead

\multicolumn{3}{l}{\textit{Table~\ref{tab:roi-unique} continued from previous page}} \\
\toprule
\textbf{Selectivity} & \textbf{ROI} & \textbf{Unique profiles} \\
\midrule
\endhead

\midrule
\multicolumn{3}{r}{\textit{Continued on next page}} \\
\endfoot

\bottomrule
\endlastfoot

Face & FFA-1 &
\textbullet\ \textit{voxel\_6016}: uniforms (chef, sports, transportation) and people through glass/windows. \newline
\textbullet\ \textit{voxel\_6035}: facial features, hair styling, accessories (glasses, hats); dynamic water activities. \newline
\textbullet\ \textit{voxel\_6063}: close-up animal faces (horses, dogs, cats) and group animal scenes (elephants). \newline
\textbullet\ \textit{voxel\_6258}: athletes in action poses; winter outdoor settings; natural animal settings. \newline
\textbullet\ \textit{voxel\_6483}: distinctive headwear; organized composition; primates; skiing and outdoor activities. \newline
\textbullet\ \textit{voxel\_8755}: close-up dog faces and heads; hand-object interactions (food, controllers, remotes). \newline
\textbullet\ \textit{voxel\_8759}: large animals with anatomical detail; food imagery (stir-fried rice); athletic jumping and skateboarding. \newline
\textbullet\ \textit{voxel\_8912}: winter sports (skiing, snowboarding) and water sports (surfing, kiteboarding) with specialized equipment. \newline
\textbullet\ \textit{voxel\_17046}: cattle with prominent facial features; formal attire emphasis. \newline
\textbullet\ \textit{voxel\_16744}: compositional scenes with both human and animal elements; natural or athletic settings. \newline
\textbullet\ \textit{voxel\_16449}: wildlife diversity (elephants, zebras, cattle, polar bears); outdoor or formal settings. \newline
\textbullet\ \textit{voxel\_6023}: headshots; everyday activities (phone use); indifference to specific activities. \newline
\textbullet\ \textit{voxel\_6036}: hand gestures; clothing textures; animals framed within vehicles. \newline
\textbullet\ \textit{voxel\_6064}: clothing texture; grooming details; intimate portrayals; mid-action or deliberate poses. \newline
\textbullet\ \textit{voxel\_6287}: zoo settings; laboratory settings; role-conveying attire. \newline
\textbullet\ \textit{voxel\_6681}: person-activity associations; distinctive dress codes; Philadelphia sports context. \newline
\textbullet\ \textit{voxel\_8646}: large terrestrial animals; body parts (legs, feet, tails); safari and pastoral settings. \newline
\textbullet\ \textit{voxel\_8754}: birds (especially blue-colored songbirds); diverse animal depictions in natural or action settings. \newline
\textbullet\ \textit{voxel\_8758}: domestic/social scenes involving people and animals; pets and wildlife. \newline
\textbullet\ \textit{voxel\_17042}: profile or frontal close-up views; eyes, ears, facial structure; body posture in action. \newline
\textbullet\ \textit{voxel\_16741}: patterned dresses; group compositions; semi-natural animal environments. \\
\cmidrule{2-3}

 & FFA-2 &
\textbullet\ \textit{voxel\_20794}: winter sports and water sports activities; zoo/natural animal settings; yellow garments and winter attire. \newline
\textbullet\ \textit{voxel\_20732}: body part visibility (legs, arms, hands); outdoor/elevated settings with sky or vegetation; brightly colored garments and winter gear. \newline
\textbullet\ \textit{voxel\_20795}: formal settings and portrait contexts; dynamic poses; wildlife in natural environments (elephants, zebras). \newline
\textbullet\ \textit{voxel\_20737}: multiple subjects in close proximity; grouped animals (herds, family units); formal portrait photography emphasis. \newline
\textbullet\ \textit{voxel\_20662}: close-up and portrait-oriented compositions; prominent eye, ear, and facial anatomy visibility; contextual and formal settings. \\
\cmidrule{2-3}

 & OFA &
\textbullet\ \textit{voxel\_2271}: winter sports emphasis with visible athletic gear (jackets, helmets, gloves) and upper body visibility; authority figures in formal attire. \newline
\textbullet\ \textit{voxel\_2529}: occupational roles like chefs; group scenes with winter clothing; explicit sensitivity to human body movement patterns. \newline
\textbullet\ \textit{voxel\_3095}: everyday activities and interactions in professional, recreational, and domestic settings; emphasis on human-animal interactions in natural environments. \newline
\textbullet\ \textit{voxel\_2528}: secondary sensitivity to indoor domestic scenes with media/technology elements; interactions with specific animals (zebras, horses, dogs). \newline
\textbullet\ \textit{voxel\_3094}: professional uniforms and specialized gear (chef, rescue worker); casual human activities like computer work or gaming; particular focus on human-dog interactions. \\
\cmidrule{2-3}

 & aTL-faces &
\textbullet\ \textit{voxel\_9866}: particular sensitivity to upper-body and portrait compositions with emphasis on athletic wear, sleeveless garments, structured clothing with visible design elements (straps, zippers, pockets), and distinctive styling like curled hair; horses as notable animal preference. \newline
\textbullet\ \textit{voxel\_9868}: strong emphasis on animal facial features and textures, particularly dogs and mammals; animals appear as primary rather than secondary response category. \newline
\textbullet\ \textit{voxel\_9865}: specific sensitivity to glasses, formal business wear (suits, jackets with buttons), culinary uniforms, and foxes as distinctive animal preference. \newline
\textbullet\ \textit{voxel\_9867}: specialized preference for close-up and detailed facial views emphasizing eye gaze, natural expressions, and facial structure; wildlife subjects (sheep, elephants, bears) and profile/direct-facing portrait compositions. \newline
\textbullet\ \textit{voxel\_9870}: strong contextual sensitivity to professional/specialized settings with activity-specific clothing (chefs in kitchens, tennis players on courts, lab workers); winter/outdoor scenes with distinctive visual elements. \newline
\textbullet\ \textit{voxel\_21321}: emphasis on dynamic activities and sports, particularly tennis players in action; full-body action poses and outdoor/active contexts as primary focus. \\
\midrule

Place & OPA &
\textbullet\ \textit{voxel\_3867}: computer workstations, desktop setups, recreational/sports environments like tennis courts, and urban street scenes. \newline
\textbullet\ \textit{voxel\_4191}: maritime and waterfront scenes with boats, iconic public structures like British phone boxes and clock towers, human-scale infrastructure. \newline
\textbullet\ \textit{voxel\_3866}: winter sports scenes (skiing/snowboarding), maritime/water-related activities, active structured environments. \newline
\textbullet\ \textit{voxel\_4190}: doors, windows, and architectural elements with cabinetry emphasis. \newline
\textbullet\ \textit{voxel\_4194}: doors and doorways (especially open doors and opening actions), winter sports activities (skiing/snowboarding), urban outdoor scenes with pedestrians, park settings with natural elements. \\
\cmidrule{2-3}

 & PPA &
\textbullet\ \textit{voxel\_8629}: winter sports activities and snowy landscapes alongside kitchen/bathroom fixtures. \newline
\textbullet\ \textit{voxel\_8734}: symmetric, organized arrangements of objects with high-weight feature emphasis across diverse interior environments. \newline
\textbullet\ \textit{voxel\_8862}: water/maritime environments and organized spatial arrangements with utilitarian purposes. \newline
\textbullet\ \textit{voxel\_8851}: urban infrastructure, transportation systems (trains, railways), and elevated architectural viewpoints with technical elements like tracks and overhead wires. \newline
\textbullet\ \textit{voxel\_8869}: human figures engaged in sports, recreation, and domestic activities, particularly in winter sports and snowy environments. \\
\cmidrule{2-3}

 & RSC &
\textbullet\ \textit{voxel\_7650}: formal attire (business suits, winter sports gear) and preference for organized interior fixtures. \newline
\textbullet\ \textit{voxel\_7835}: domestic workspace environments (kitchens, bathrooms, offices) with functional equipment and professional contexts. \newline
\textbullet\ \textit{voxel\_7651}: dynamic human activities, winter sports with protective gear, elevated positioning, and nighttime urban scenes. \newline
\textbullet\ \textit{voxel\_8015}: infrastructure and structural elements with elevated/distant viewpoints, transportation systems, railway tracks, and maritime settings. \newline
\textbullet\ \textit{voxel\_18720}: athletic activities (tennis), diverse transportation modes (trains, buses, bicycles, motorcycles), and waterfront structures with vertical architectural elements. \\
\midrule

Body & EBA &
\textbullet\ \textit{voxel\_2271}: winter sports emphasis with visible athletic gear (jackets, helmets, gloves); upper body visibility and formal authority figures. \newline
\textbullet\ \textit{voxel\_2529}: occupational roles like chefs; group scenes with winter clothing; sensitivity to both body movement and outdoor scenarios. \newline
\textbullet\ \textit{voxel\_3095}: everyday activities and interactions in professional, recreational, and domestic settings; emphasis on human-animal interactions. \newline
\textbullet\ \textit{voxel\_16289}: polar bears and zebras with emphasis on body parts and group compositions; robust water and snow sports preference. \newline
\textbullet\ \textit{voxel\_16557}: specialized athletic equipment and gear; minor sensitivity to distinctive red headwear and formal attire. \newline
\textbullet\ \textit{voxel\_2528}: secondary sensitivity to indoor domestic scenes with media/technology elements alongside outdoor sports. \newline
\textbullet\ \textit{voxel\_3094}: professional uniforms and specialized gear (chef, rescue worker); human-animal interactions particularly with dogs; casual activities like computer work or gaming. \newline
\textbullet\ \textit{voxel\_16563}: broad preference for people in motion, distinctive clothing, or natural environment interactions; formal attire sensitivity. \newline
\textbullet\ \textit{voxel\_16852}: occasional natural subjects like birds; emphasis on dynamic wave interaction and splashing effects in water sports. \newline
\textbullet\ \textit{voxel\_16553}: equestrian activities and sports uniforms; polar bears in naturalistic poses; formal contexts. \\
\cmidrule{2-3}

 & FBA-1 &
\textbullet\ \textit{voxel\_6063}: secondary sensitivity to animal faces and bodies in group scenes. \newline
\textbullet\ \textit{voxel\_6483}: broader responsiveness to subjects with clear visual structure and organized composition. \newline
\textbullet\ \textit{voxel\_8755}: particular preference for hand-object interactions and detailed close-up views of dogs' faces. \newline
\textbullet\ \textit{voxel\_8759}: robust sensitivity to large animals with anatomical detail and secondary sensitivity to food imagery. \newline
\textbullet\ \textit{voxel\_8912}: specialized focus on dynamic action sports with winter sports (skiing, snowboarding) and water sports (surfing, kiteboarding) emphasis. \newline
\textbullet\ \textit{voxel\_9023}: consistent preference for winter sports and racquet sports with emphasis on protective gear and active body positioning. \newline
\textbullet\ \textit{voxel\_6064}: encoding of clothing texture, grooming details, and subjects captured mid-action or in deliberate poses. \newline
\textbullet\ \textit{voxel\_6287}: sensitivity to subjects performing actions or wearing distinctive attire that conveys role or activity. \newline
\textbullet\ \textit{voxel\_6681}: encodes person-activity associations and distinctive dress codes across varied human-centered scenes. \newline
\textbullet\ \textit{voxel\_8646}: strong focus on large terrestrial animals and their anatomical features (legs, feet, tails) in natural outdoor settings. \newline
\textbullet\ \textit{voxel\_8754}: consistent activation across diverse animal depictions in natural or action settings, including birds with color sensitivity. \newline
\textbullet\ \textit{voxel\_8758}: particular sensitivity to human faces in business attire and domestic/social scenes involving people and animals. \newline
\textbullet\ \textit{voxel\_8915}: preference spans both action/movement component and athletic apparel or equipment in diverse sports contexts. \\
\cmidrule{2-3}

 & FBA-2 &
\textbullet\ \textit{voxel\_20794}: winter sports emphasis (paragliding, snowboarding) and zoo animals; yellow garments sensitivity. \newline
\textbullet\ \textit{voxel\_17046}: strong focus on formal professional contexts and headwear; cattle with prominent facial features. \newline
\textbullet\ \textit{voxel\_16744}: compositional sensitivity to mixed human-animal scenes and focal subject positioning in natural settings. \newline
\textbullet\ \textit{voxel\_20732}: emphasis on visible body parts (legs, arms, hands) and elevated outdoor settings with sky/vegetation. \newline
\textbullet\ \textit{voxel\_16449}: broad anatomical detail sensitivity across diverse formal and outdoor activity contexts. \newline
\textbullet\ \textit{voxel\_20795}: consistent activation for tennis players specifically and characteristic animal environments. \newline
\textbullet\ \textit{voxel\_20737}: strong preference for grouped subjects in close proximity and portrait photography compositions. \newline
\textbullet\ \textit{voxel\_17042}: extreme close-up focus on distinctive facial anatomy (eyes, ears, structure) in profile or frontal views. \newline
\textbullet\ \textit{voxel\_16741}: emphasis on structured formal attire (suits, patterned dresses) and group compositions. \newline
\textbullet\ \textit{voxel\_20662}: portrait-oriented close-up compositions with clear eye and ear visibility emphasis. \\
\midrule

Word & VWFA-1 &
\textbullet\ \textit{voxel\_6016}: preference for people photographed through glass/windows; specific interest in horses. \newline
\textbullet\ \textit{voxel\_6035}: emphasis on hair styling and accessories (glasses, hats); sensitivity to dynamic water activities. \newline
\textbullet\ \textit{voxel\_6258}: strong focus on action poses and natural animal settings (zebras, elephants); winter outdoor contexts. \newline
\textbullet\ \textit{voxel\_6023}: indifference to specific activities; emphasis on headshots and everyday casual interactions; structured compositional scenes. \newline
\textbullet\ \textit{voxel\_6036}: attention to hand gestures and clothing textures; animals framed within vehicles or with prominent physical features. \\
\cmidrule{2-3}

 & VWFA-2 &
\textbullet\ \textit{voxel\_8912}: winter sports (skiing, snowboarding) and water sports (surfing, kiteboarding) with specialized equipment in challenging terrain or water conditions. \newline
\textbullet\ \textit{voxel\_9023}: winter sports (skiing, snowboarding) and racquet sports (tennis) with emphasis on winter clothing and protective gear. \newline
\textbullet\ \textit{voxel\_16964}: indoor domestic and functional spaces (dining tables, workstations, kitchens), architectural landmarks (clocks, towers), transportation (trains, aircraft), and outdoor natural settings. \newline
\textbullet\ \textit{voxel\_16974}: trains and railway infrastructure (locomotives, cars, tracks), computer workstations and monitors, food items (grilled meats, sandwiches), and architectural landmarks (clock towers). \newline
\textbullet\ \textit{voxel\_8915}: diverse sports contexts (tennis, baseball, skateboarding, wrestling, horseback riding) and wildlife scenes featuring large animals. \\
\cmidrule{2-3}

 & mTL-words &
\textbullet\ \textit{voxel\_9866}: preference for upper-body and portrait compositions with emphasis on clothing design elements (straps, zippers, pockets) and distinctive styling like curled hair; strong response to athletic wear and sleeveless garments. \newline
\textbullet\ \textit{voxel\_9868}: particular emphasis on animal facial features and fur texture; strong mammal recognition especially for dogs; sensitivity to formal attire contexts. \newline
\textbullet\ \textit{voxel\_9865}: strong focus on professional and formal attire details including glasses, business wear with buttons, and culinary uniforms; emphasis on distinctive markings and clothing in specialized contexts. \newline
\textbullet\ \textit{voxel\_9867}: preference for close-up and detailed facial views in profile or direct-facing compositions; emphasis on eye gaze, natural expressions, and emotional content; includes wildlife subjects like sheep, elephants, and bears. \newline
\textbullet\ \textit{voxel\_9870}: strong response to people in professional/specialized settings engaged in distinct activities and roles (chefs, tennis players, business professionals, lab workers); emphasis on contextually appropriate environments and winter/outdoor scenes. \\
\cmidrule{2-3}

 & mfs-words &
\textbullet\ \textit{voxel\_8755}: hand-object interactions (grasping, manipulating food, controllers, remote devices) and outdoor activity scenes. \newline
\textbullet\ \textit{voxel\_8759}: dynamic physical activities and sports (athletic jumping, skiing, skateboarding), food imagery (stir-fried rice), and professional/formal human subjects (business attire, chefs). \newline
\textbullet\ \textit{voxel\_8646}: large terrestrial animals in safari and pastoral settings (elephants, zebras, horses, cattle, big cats) with focus on body parts like legs, feet, and tails. \newline
\textbullet\ \textit{voxel\_8754}: birds (especially blue-colored songbirds) and consistent activation across diverse animal depictions in natural or action settings. \newline
\textbullet\ \textit{voxel\_8758}: professional/formal settings and domestic/social scenes involving people and animals together. \\

\end{longtable}
}

\begin{figure}[!ht]
\centering
\begin{subfigure}[t]{0.18\textwidth}\centering\includegraphics[width=\textwidth]{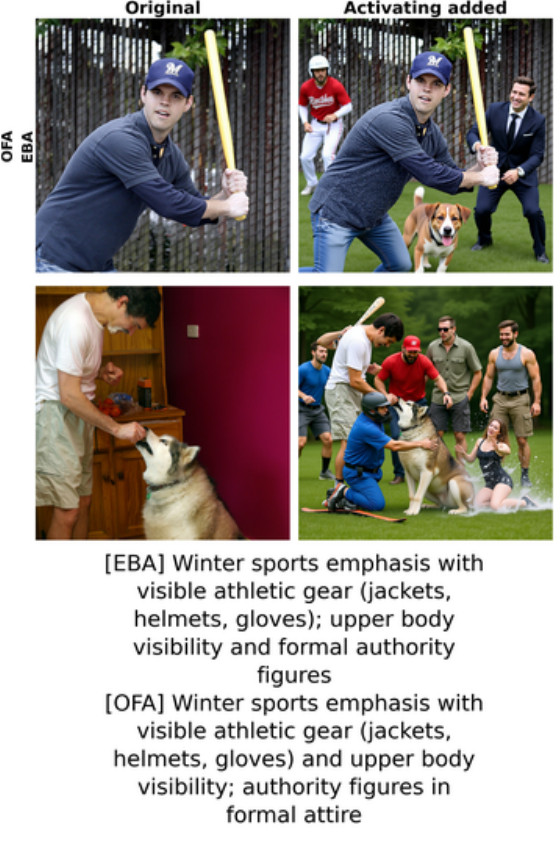}\caption*{\footnotesize voxel\_2271}\end{subfigure}\hfill
\begin{subfigure}[t]{0.18\textwidth}\centering\includegraphics[width=\textwidth]{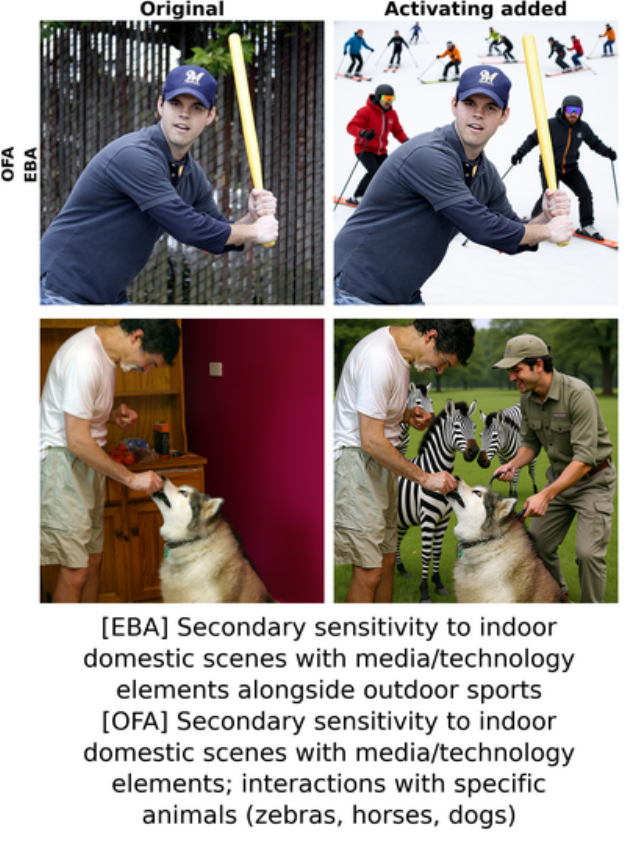}\caption*{\footnotesize voxel\_2528}\end{subfigure}\hfill
\begin{subfigure}[t]{0.18\textwidth}\centering\includegraphics[width=\textwidth]{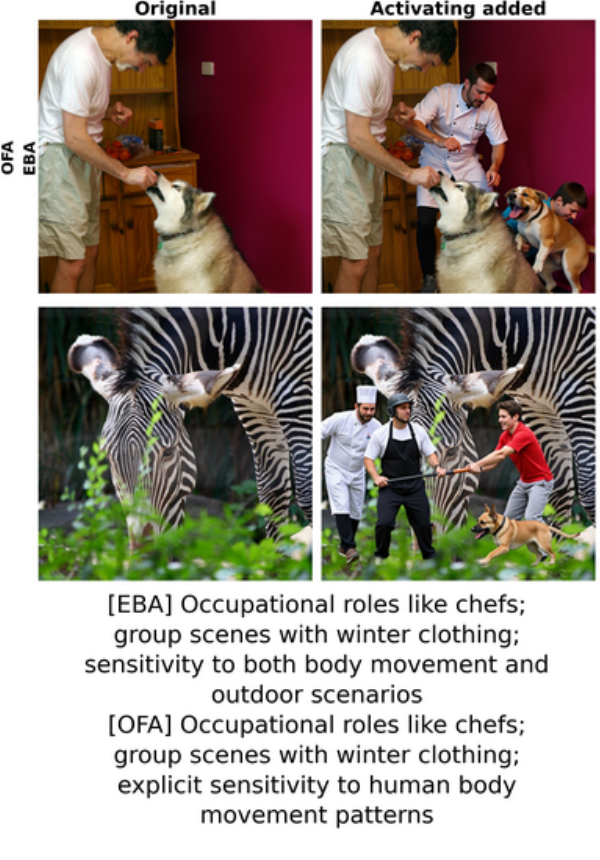}\caption*{\footnotesize voxel\_2529}\end{subfigure}\hfill
\begin{subfigure}[t]{0.18\textwidth}\centering\includegraphics[width=\textwidth]{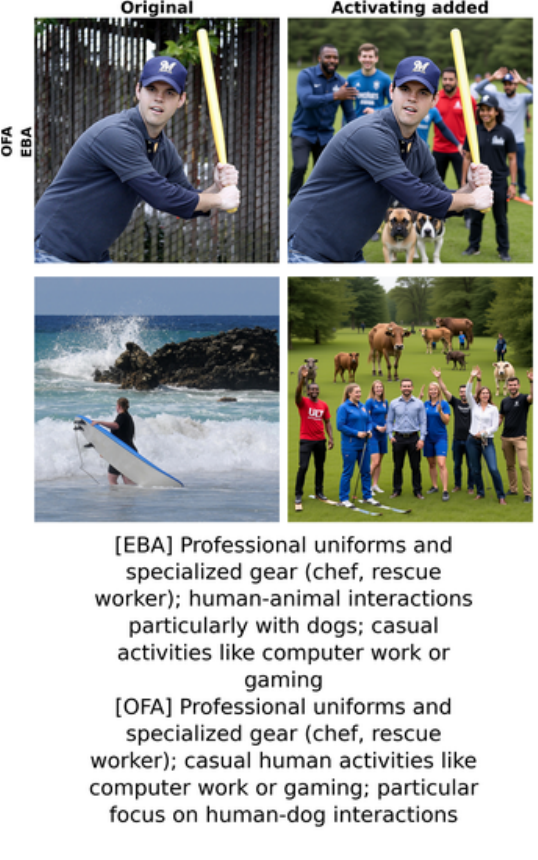}\caption*{\footnotesize voxel\_3094}\end{subfigure}\hfill
\begin{subfigure}[t]{0.18\textwidth}\centering\includegraphics[width=\textwidth]{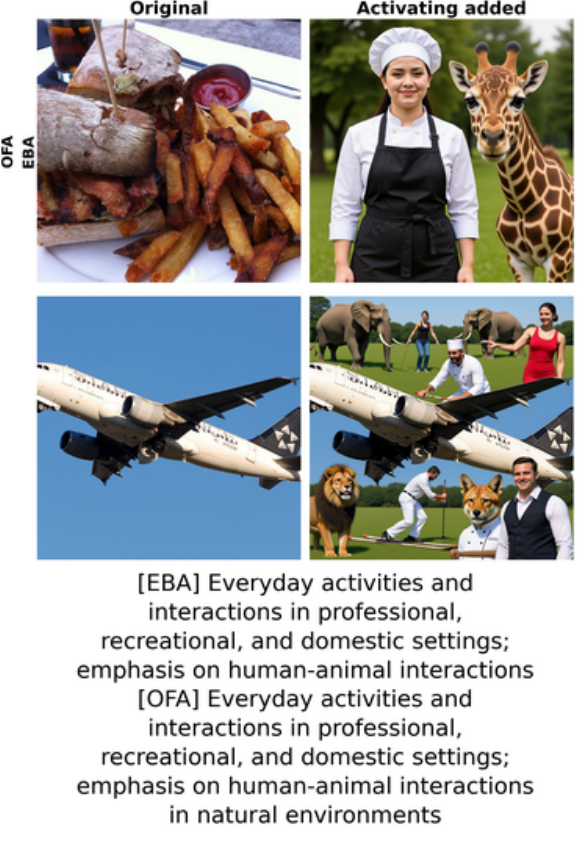}\caption*{\footnotesize voxel\_3095}\end{subfigure}\\[0.8em]
\begin{subfigure}[t]{0.18\textwidth}\centering\includegraphics[width=\textwidth]{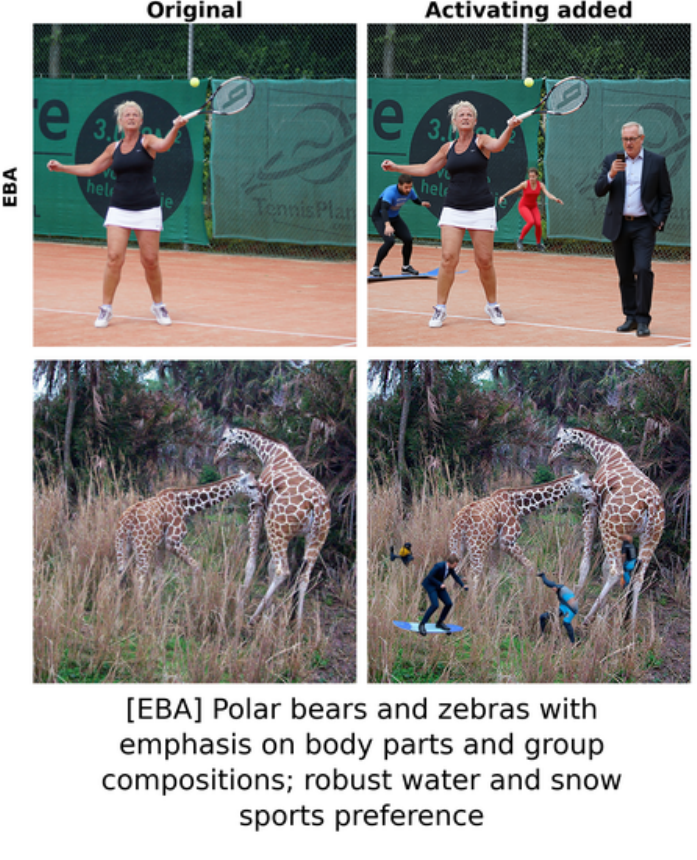}\caption*{\footnotesize voxel\_16289}\end{subfigure}\hfill
\begin{subfigure}[t]{0.18\textwidth}\centering\includegraphics[width=\textwidth]{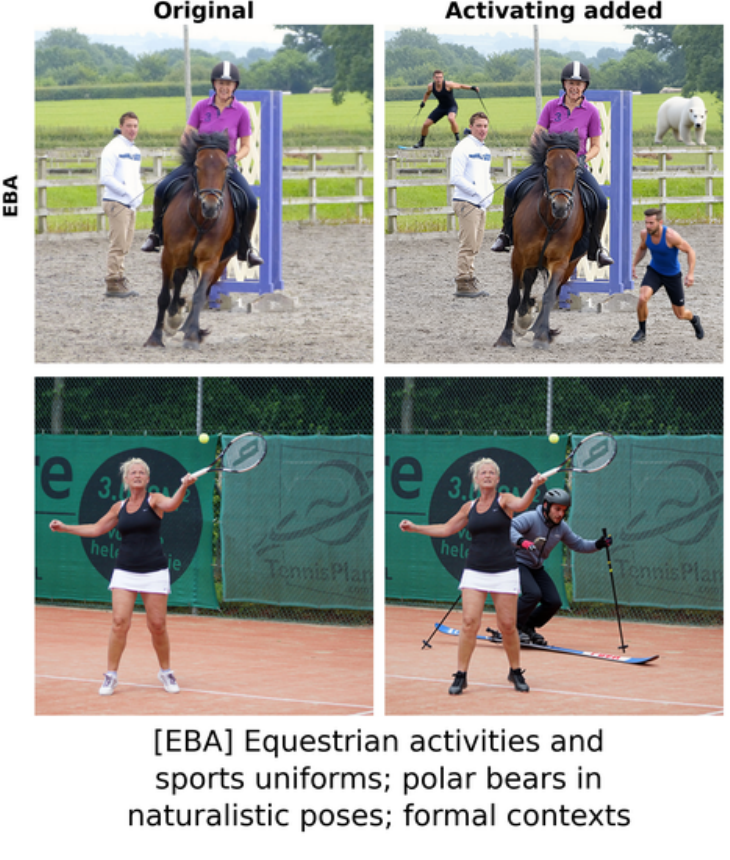}\caption*{\footnotesize voxel\_16553}\end{subfigure}\hfill
\begin{subfigure}[t]{0.18\textwidth}\centering\includegraphics[width=\textwidth]{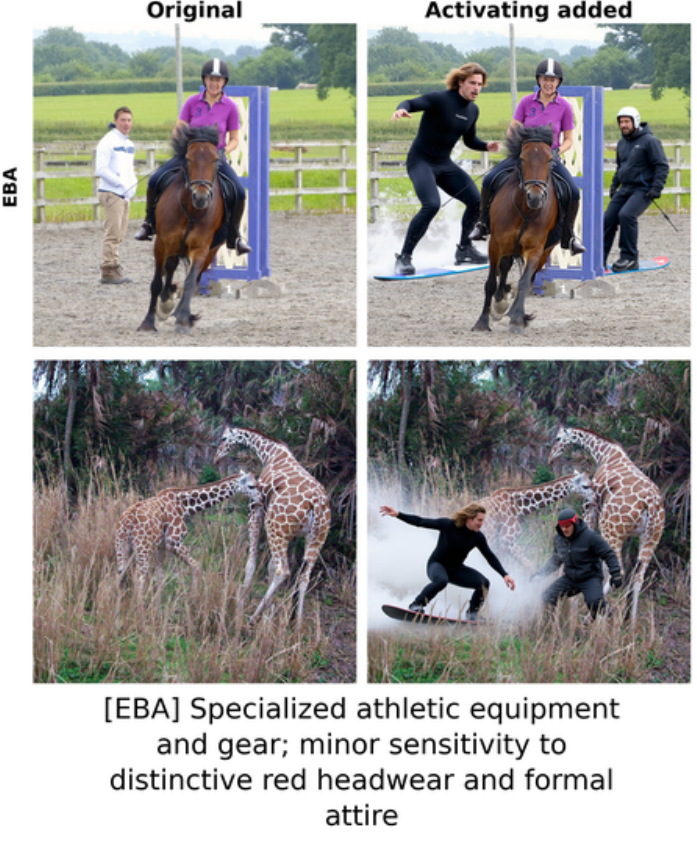}\caption*{\footnotesize voxel\_16557}\end{subfigure}\hfill
\begin{subfigure}[t]{0.18\textwidth}\centering\includegraphics[width=\textwidth]{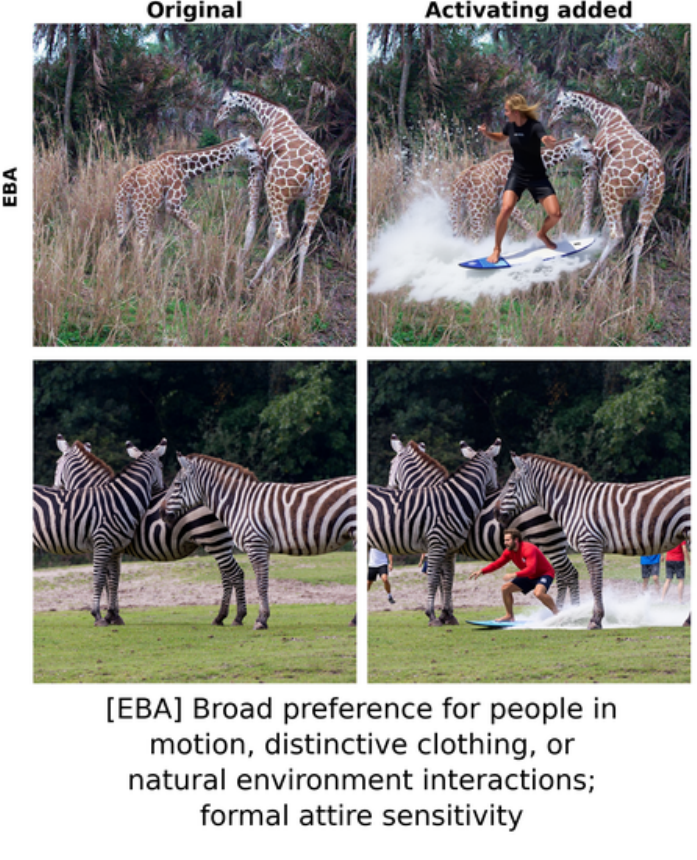}\caption*{\footnotesize voxel\_16563}\end{subfigure}\hfill
\begin{subfigure}[t]{0.18\textwidth}\centering\includegraphics[width=\textwidth]{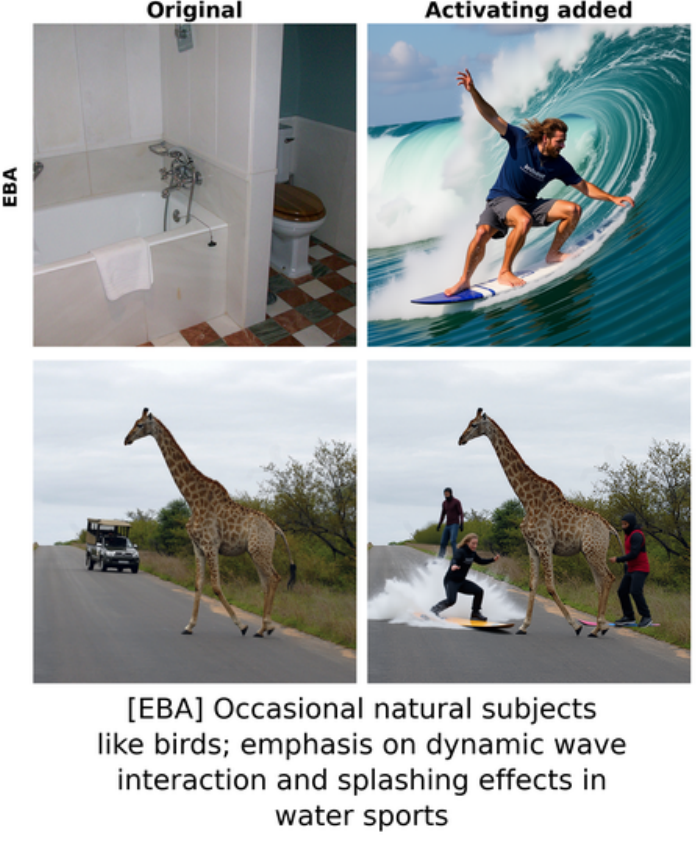}\caption*{\footnotesize voxel\_16852}\end{subfigure}
\caption{\textbf{EBA voxel profiles.} \textbf{Shared profile:} \textit{Dynamic human figures engaged in athletic activities and sports (particularly water sports like surfing and winter sports like skiing), wearing athletic or specialized gear, often in motion or action contexts. Secondary consistent preference for animals in natural or outdoor settings.}}
\label{app:fig:eba-profiles}
\end{figure}

\begin{figure}[!ht]
\centering
\begin{subfigure}[t]{0.18\textwidth}\centering\includegraphics[width=\textwidth]{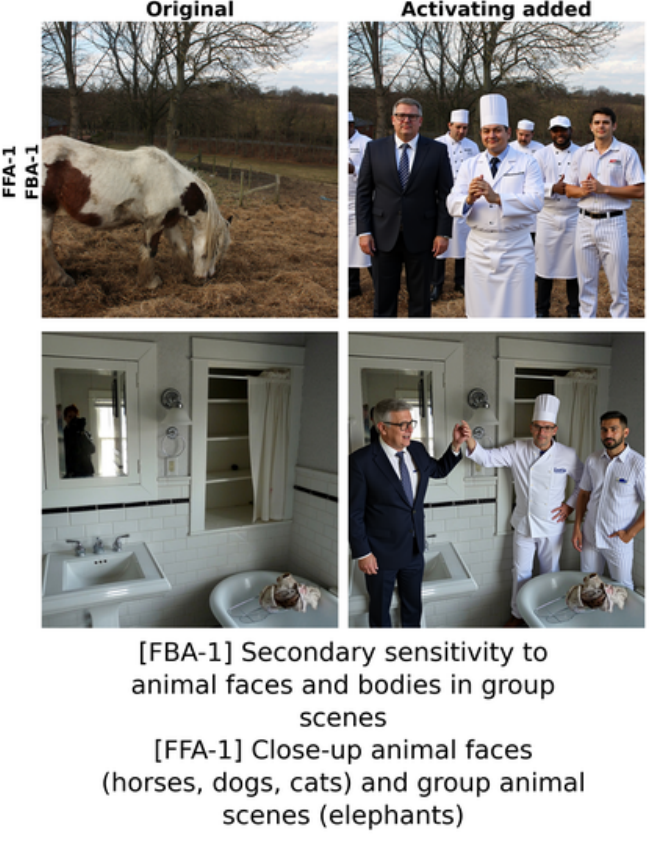}\caption*{\footnotesize voxel\_6063}\end{subfigure}\hfill
\begin{subfigure}[t]{0.18\textwidth}\centering\includegraphics[width=\textwidth]{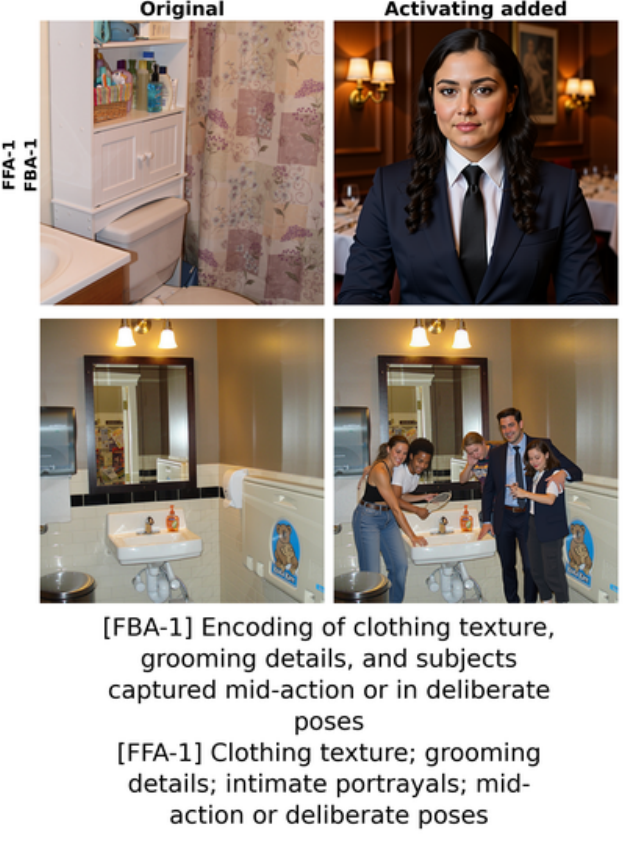}\caption*{\footnotesize voxel\_6064}\end{subfigure}\hfill
\begin{subfigure}[t]{0.18\textwidth}\centering\includegraphics[width=\textwidth]{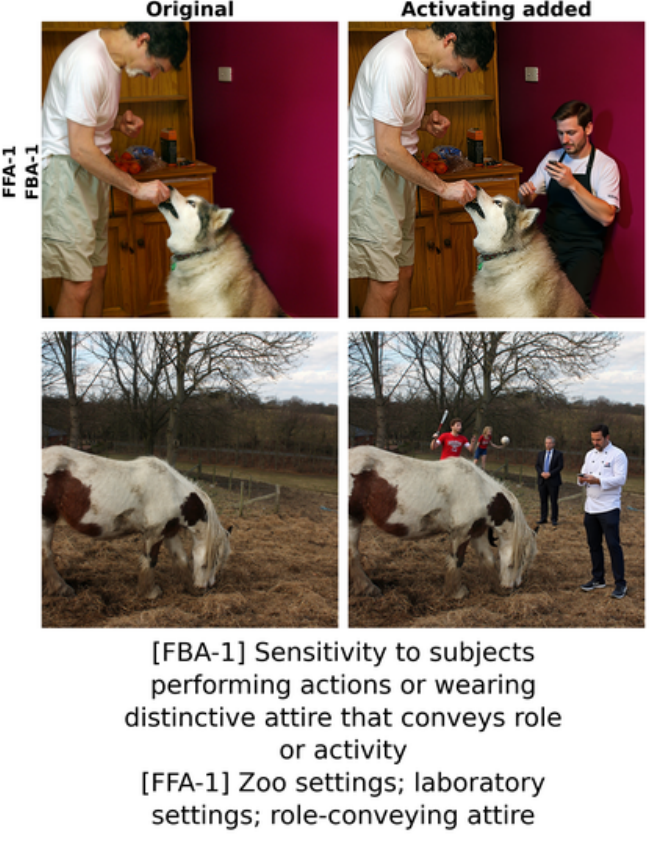}\caption*{\footnotesize voxel\_6287}\end{subfigure}\hfill
\begin{subfigure}[t]{0.18\textwidth}\centering\includegraphics[width=\textwidth]{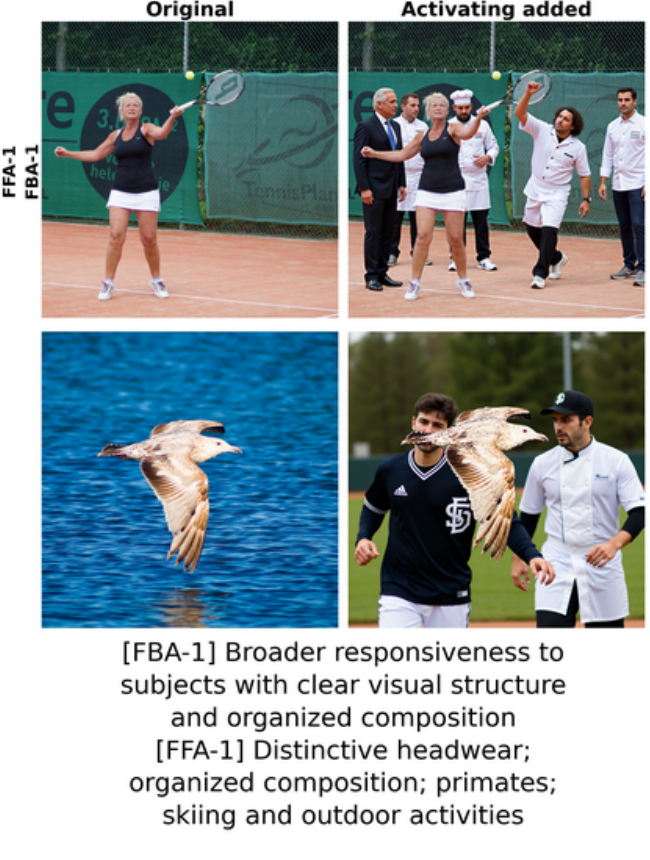}\caption*{\footnotesize voxel\_6483}\end{subfigure}\hfill
\begin{subfigure}[t]{0.18\textwidth}\centering\includegraphics[width=\textwidth]{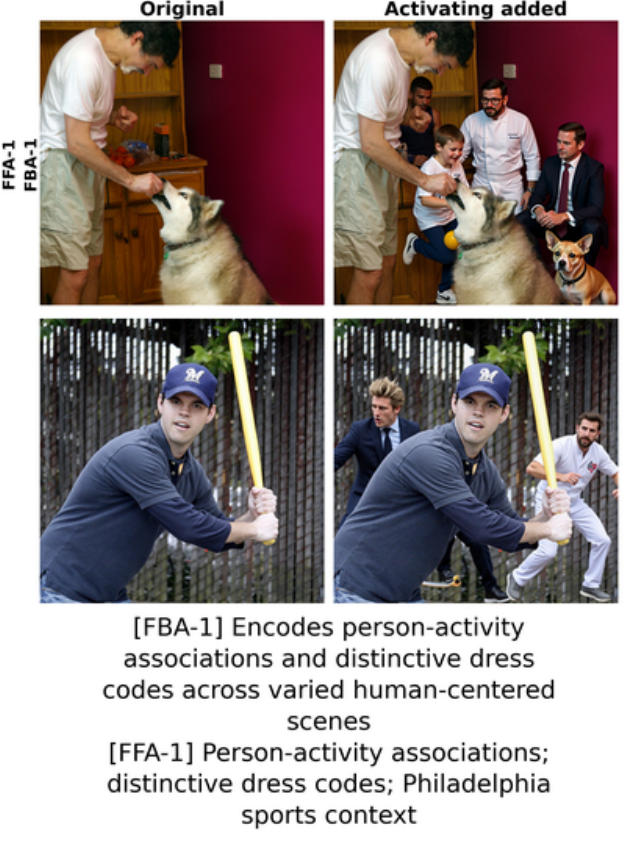}\caption*{\footnotesize voxel\_6681}\end{subfigure}\\[0.8em]
\begin{subfigure}[t]{0.18\textwidth}\centering\includegraphics[width=\textwidth]{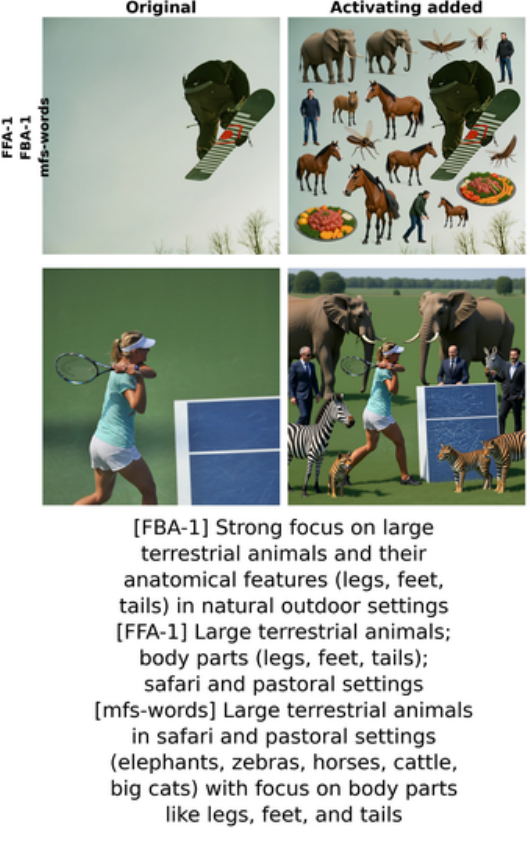}\caption*{\footnotesize voxel\_8646}\end{subfigure}\hfill
\begin{subfigure}[t]{0.18\textwidth}\centering\includegraphics[width=\textwidth]{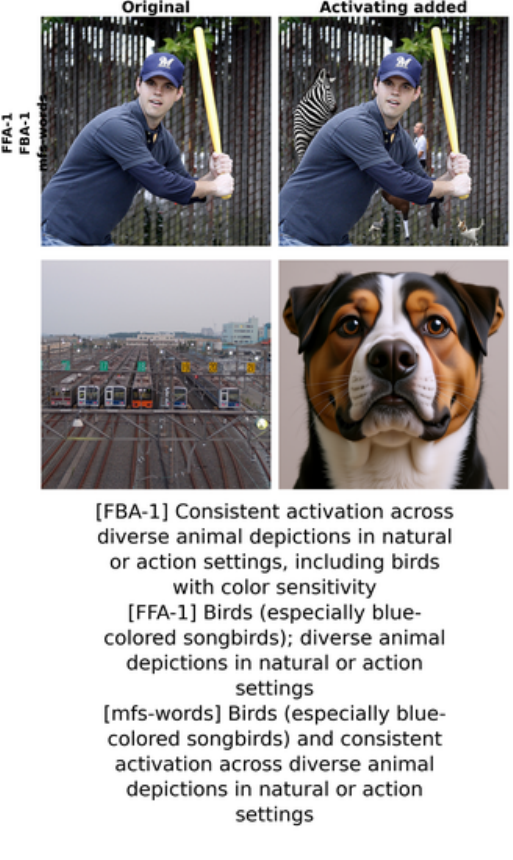}\caption*{\footnotesize voxel\_8754}\end{subfigure}\hfill
\begin{subfigure}[t]{0.18\textwidth}\centering\includegraphics[width=\textwidth]{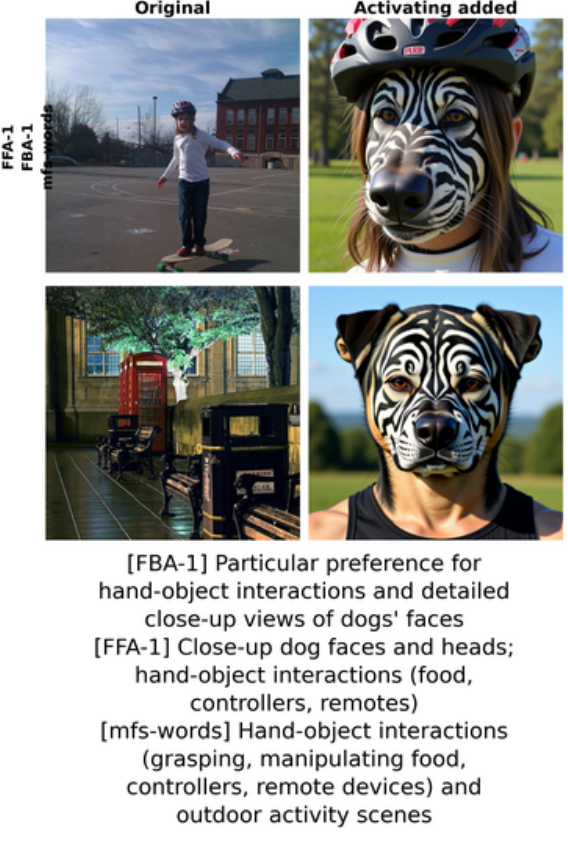}\caption*{\footnotesize voxel\_8755}\end{subfigure}\hfill
\begin{subfigure}[t]{0.18\textwidth}\centering\includegraphics[width=\textwidth]{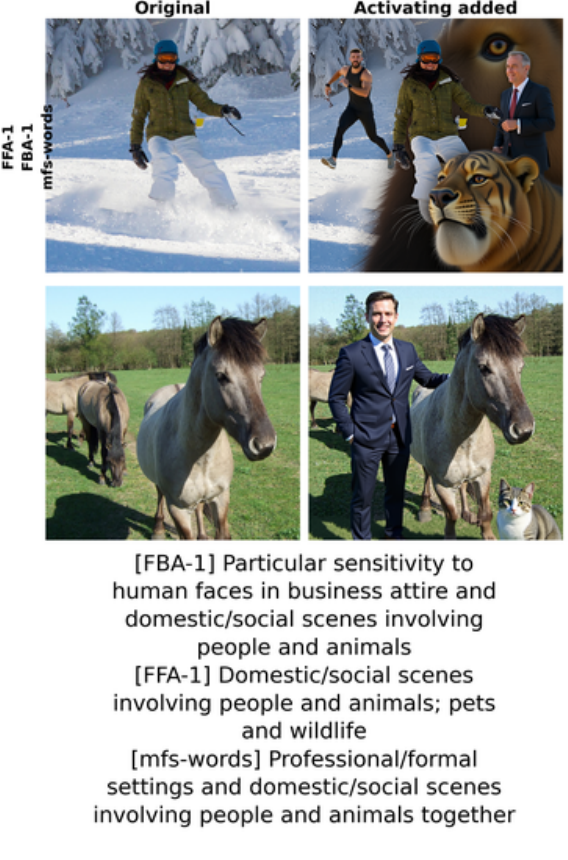}\caption*{\footnotesize voxel\_8758}\end{subfigure}\hfill
\begin{subfigure}[t]{0.18\textwidth}\centering\includegraphics[width=\textwidth]{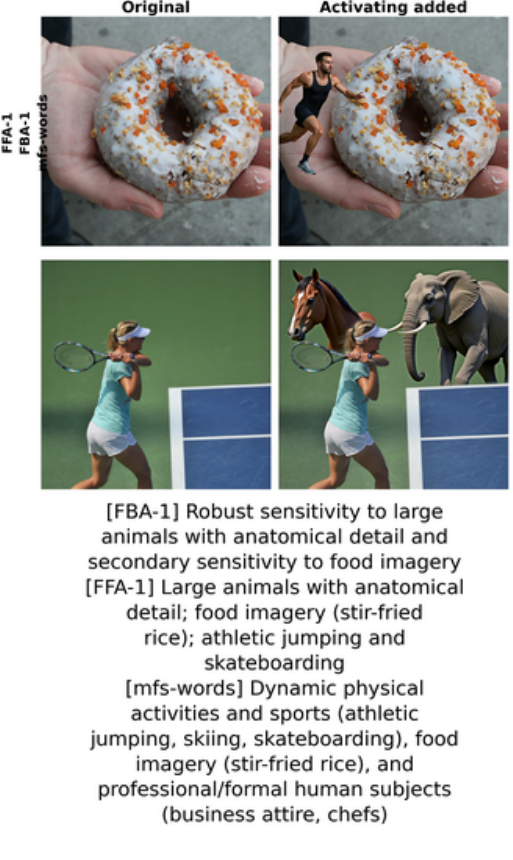}\caption*{\footnotesize voxel\_8759}\end{subfigure}\\[0.8em]
\begin{subfigure}[t]{0.18\textwidth}\centering\includegraphics[width=\textwidth]{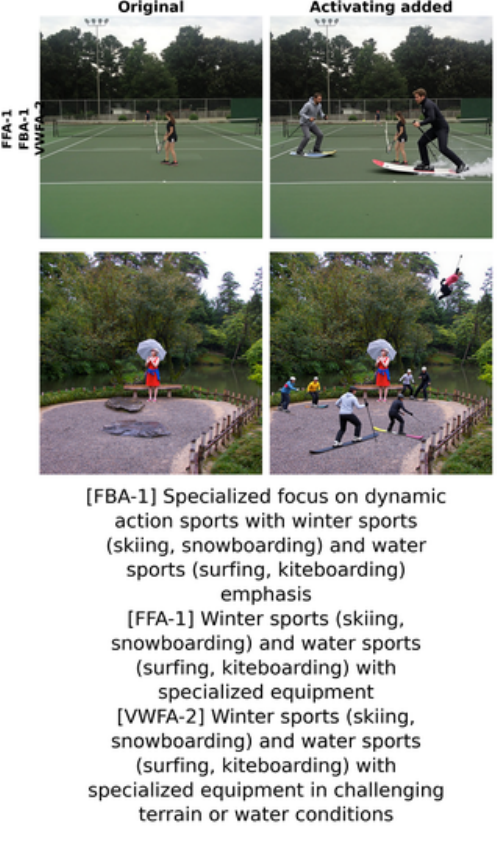}\caption*{\footnotesize voxel\_8912}\end{subfigure}\hfill
\begin{subfigure}[t]{0.18\textwidth}\centering\includegraphics[width=\textwidth]{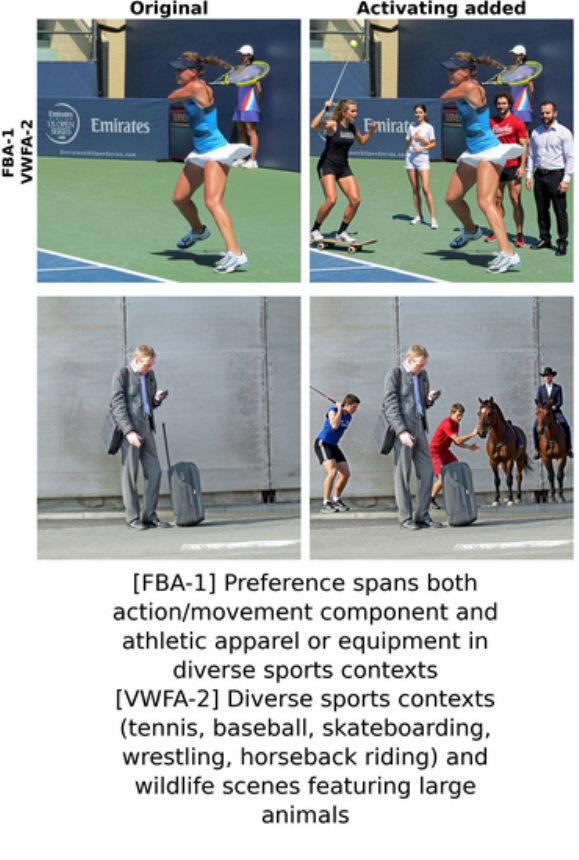}\caption*{\footnotesize voxel\_8915}\end{subfigure}\hfill
\begin{subfigure}[t]{0.18\textwidth}\centering\includegraphics[width=\textwidth]{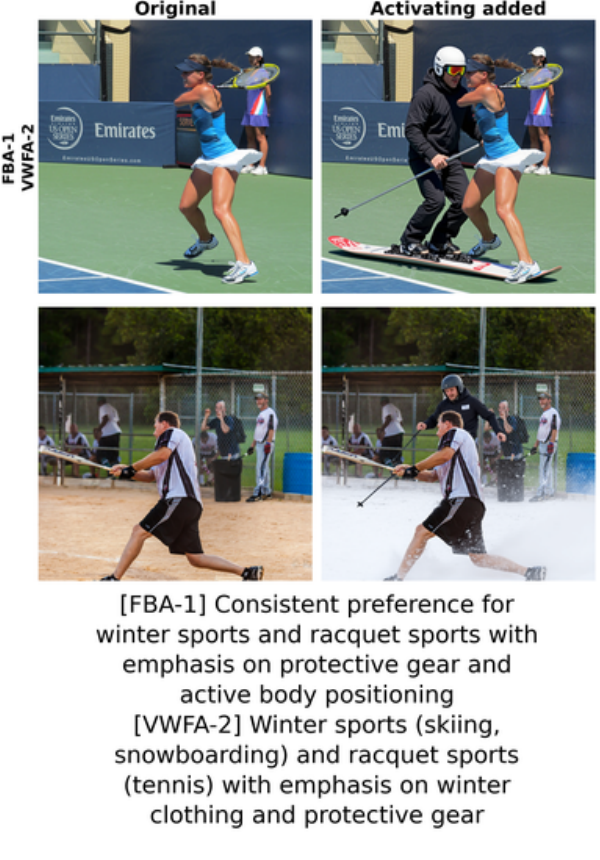}\caption*{\footnotesize voxel\_9023}\end{subfigure}
\caption{\textbf{FBA-1 voxel profiles.} \textbf{Shared profile:} \textit{Strong preference for human subjects in distinctive contexts (professional attire, sports activities, formal settings) and animals with visible anatomical or facial details. Consistent sensitivity to portraits, close-up facial features, and subjects engaged in identifiable roles or actions.}}
\label{app:fig:fba1-profiles}
\end{figure}

\begin{figure}[!ht]
\centering
\begin{subfigure}[t]{0.18\textwidth}\centering\includegraphics[width=\textwidth]{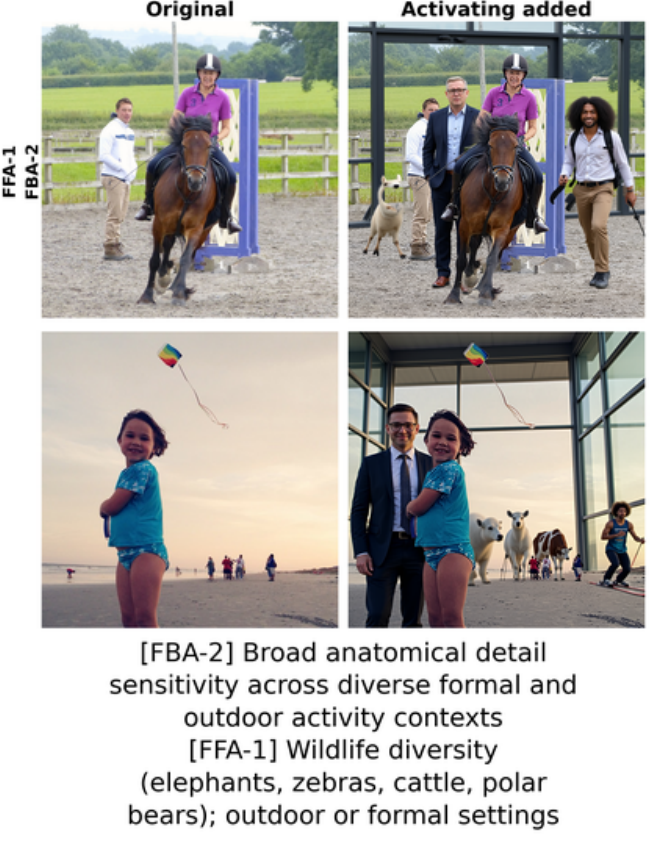}\caption*{\footnotesize voxel\_16449}\end{subfigure}\hfill
\begin{subfigure}[t]{0.18\textwidth}\centering\includegraphics[width=\textwidth]{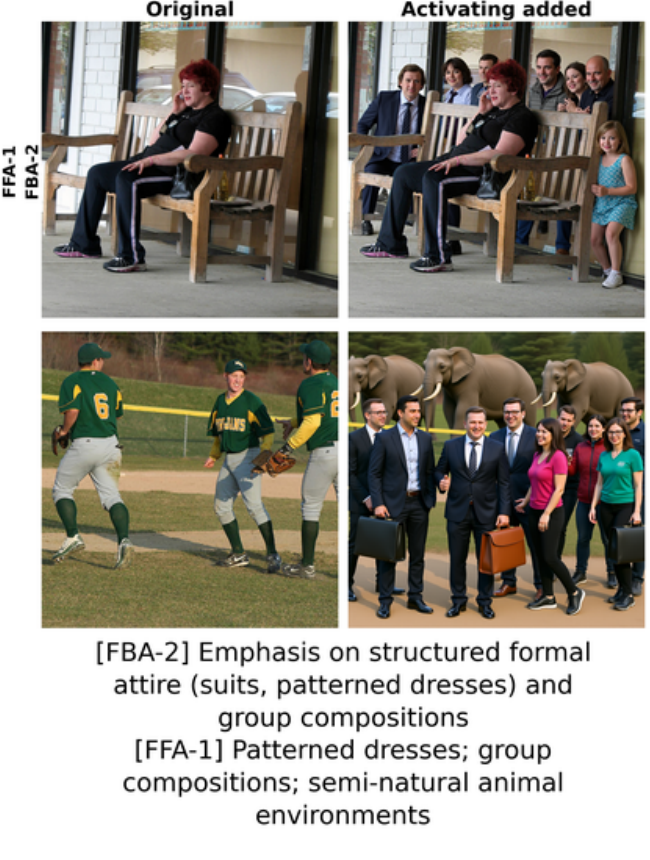}\caption*{\footnotesize voxel\_16741}\end{subfigure}\hfill
\begin{subfigure}[t]{0.18\textwidth}\centering\includegraphics[width=\textwidth]{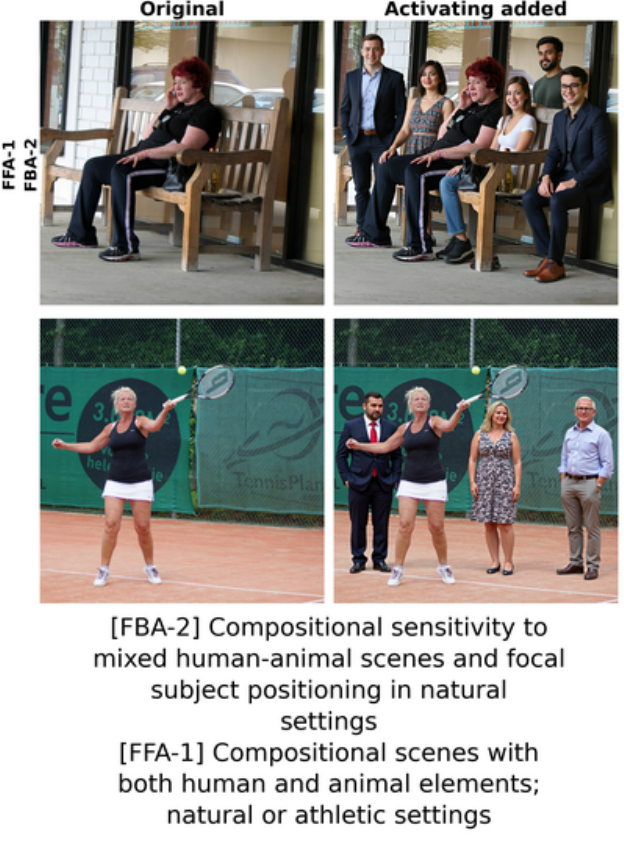}\caption*{\footnotesize voxel\_16744}\end{subfigure}\hfill
\begin{subfigure}[t]{0.18\textwidth}\centering\includegraphics[width=\textwidth]{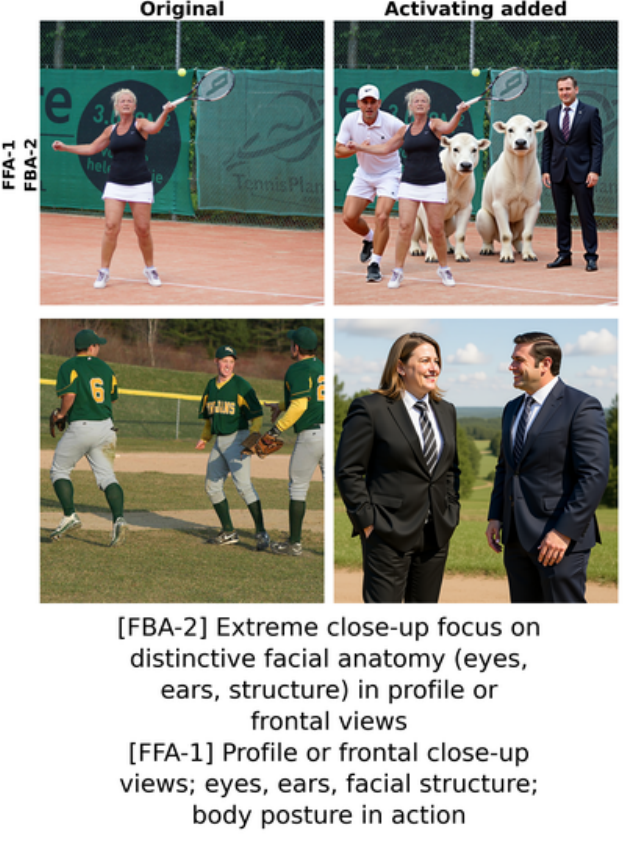}\caption*{\footnotesize voxel\_17042}\end{subfigure}\hfill
\begin{subfigure}[t]{0.18\textwidth}\centering\includegraphics[width=\textwidth]{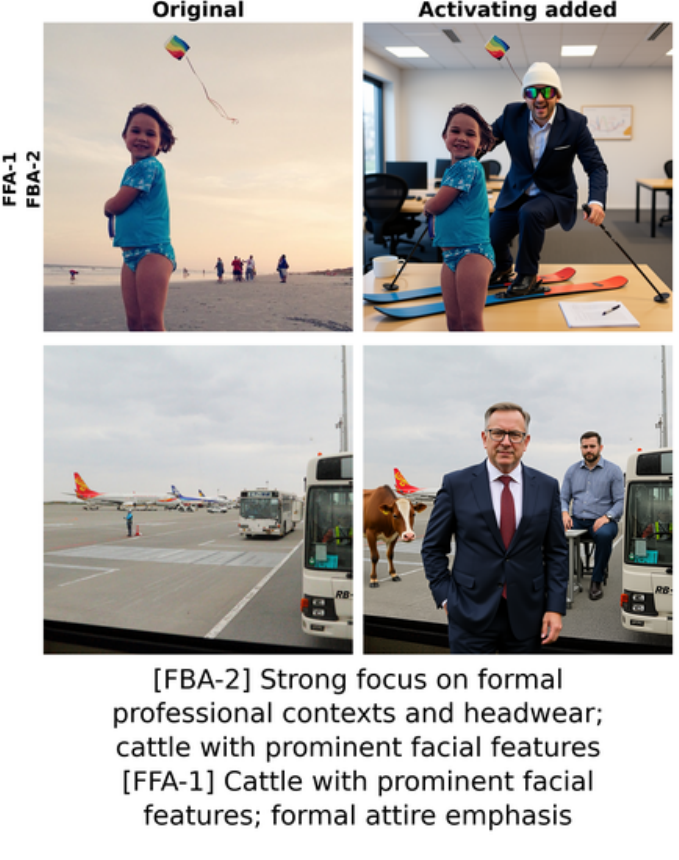}\caption*{\footnotesize voxel\_17046}\end{subfigure}\\[0.8em]
\begin{subfigure}[t]{0.18\textwidth}\centering\includegraphics[width=\textwidth]{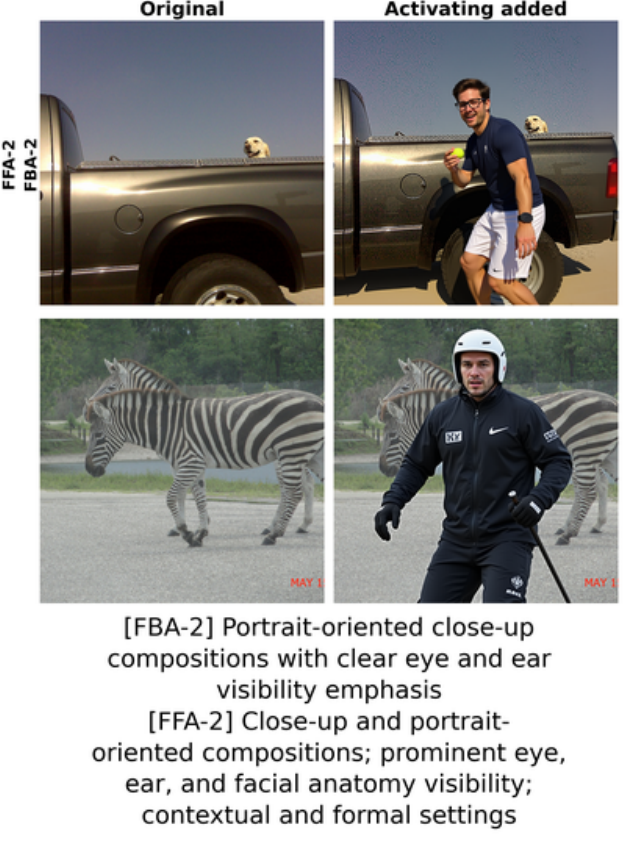}\caption*{\footnotesize voxel\_20662}\end{subfigure}\hfill
\begin{subfigure}[t]{0.18\textwidth}\centering\includegraphics[width=\textwidth]{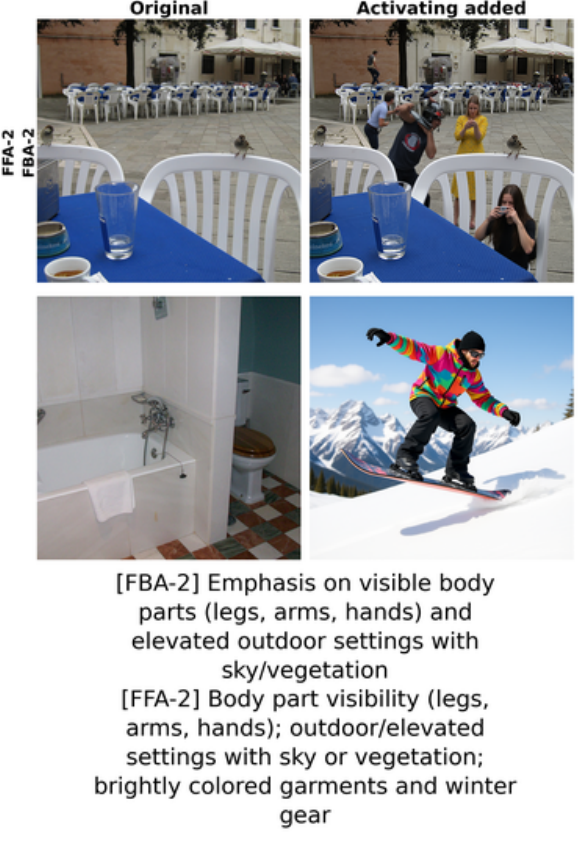}\caption*{\footnotesize voxel\_20732}\end{subfigure}\hfill
\begin{subfigure}[t]{0.18\textwidth}\centering\includegraphics[width=\textwidth]{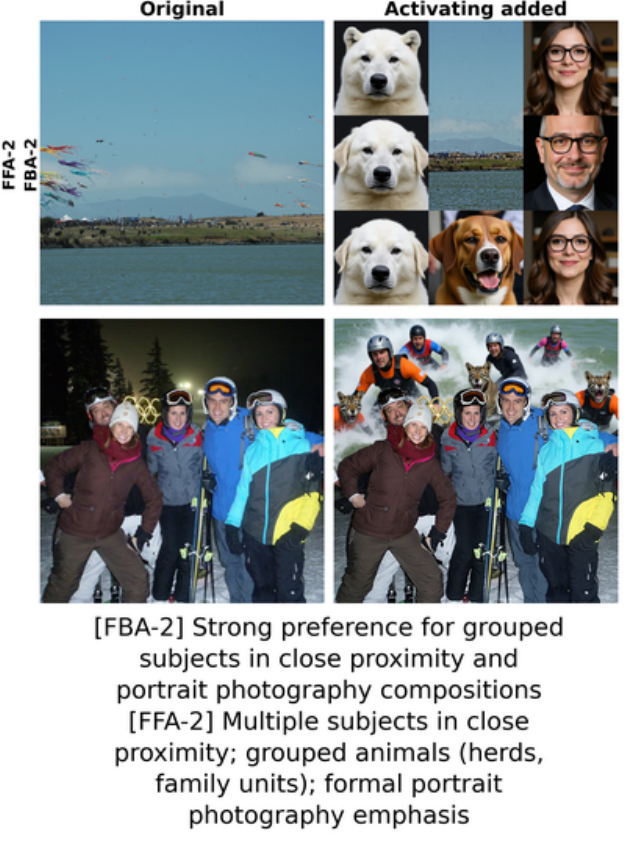}\caption*{\footnotesize voxel\_20737}\end{subfigure}\hfill
\begin{subfigure}[t]{0.18\textwidth}\centering\includegraphics[width=\textwidth]{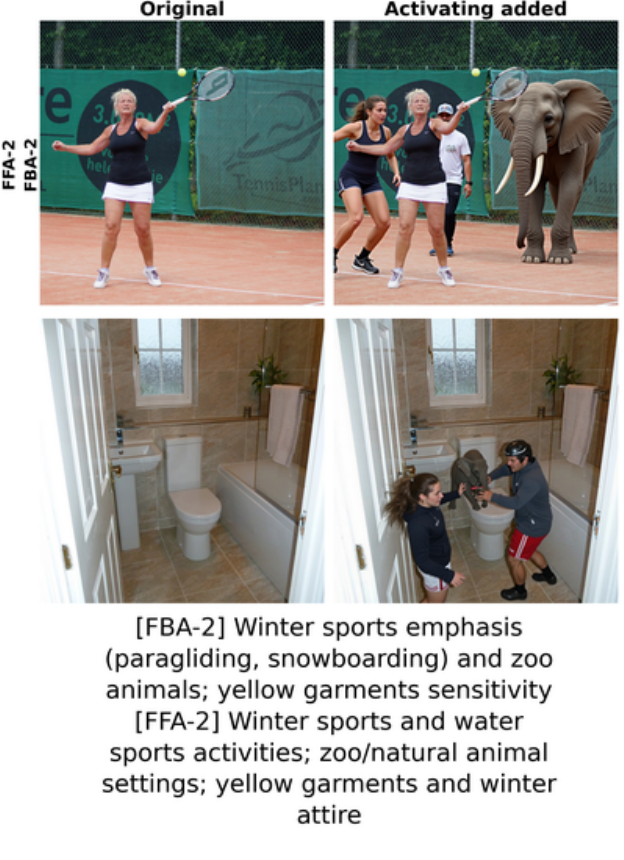}\caption*{\footnotesize voxel\_20794}\end{subfigure}\hfill
\begin{subfigure}[t]{0.18\textwidth}\centering\includegraphics[width=\textwidth]{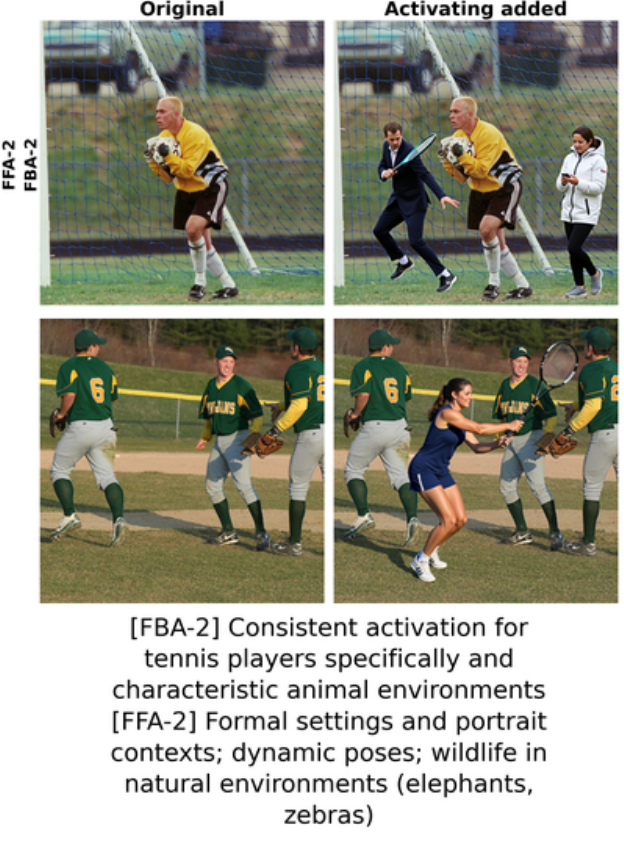}\caption*{\footnotesize voxel\_20795}\end{subfigure}
\caption{\textbf{FBA-2 voxel profiles.} \textbf{Shared profile:} \textit{Strong responsiveness to human and animal subjects in dynamic or formal contexts, with consistent sensitivity to facial features, distinctive clothing/attire, and outdoor or athletic settings. Activation across diverse subjects including people engaged in sports (skiing, surfing, tennis), wildlife (elephants, polar bears, cattle), and individuals in formal or winter wear.}}
\label{app:fig:fba2-profiles}
\end{figure}

\begin{figure}[!ht]
\centering
\begin{subfigure}[t]{0.18\textwidth}\centering\includegraphics[width=\textwidth]{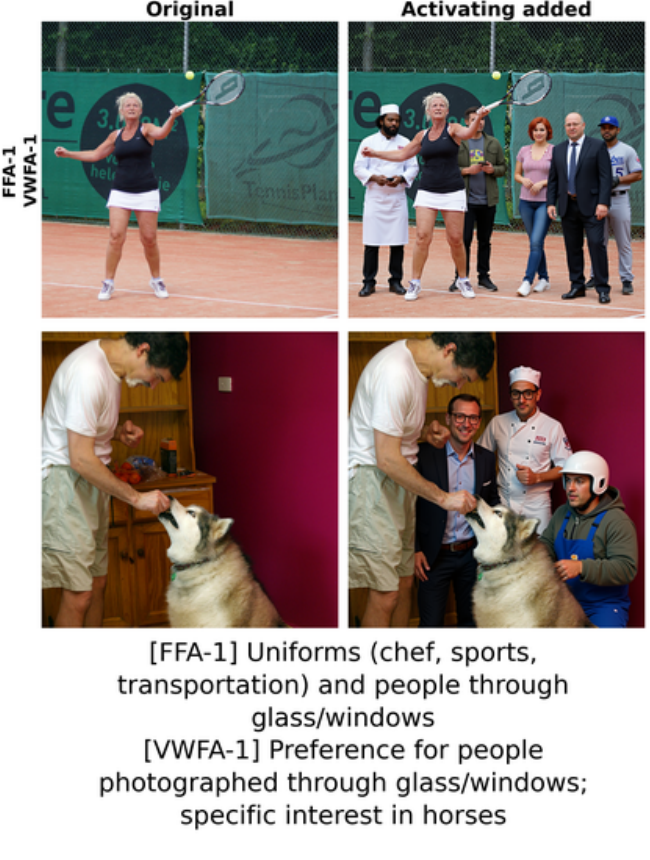}\caption*{\footnotesize voxel\_6016}\end{subfigure}\hfill
\begin{subfigure}[t]{0.18\textwidth}\centering\includegraphics[width=\textwidth]{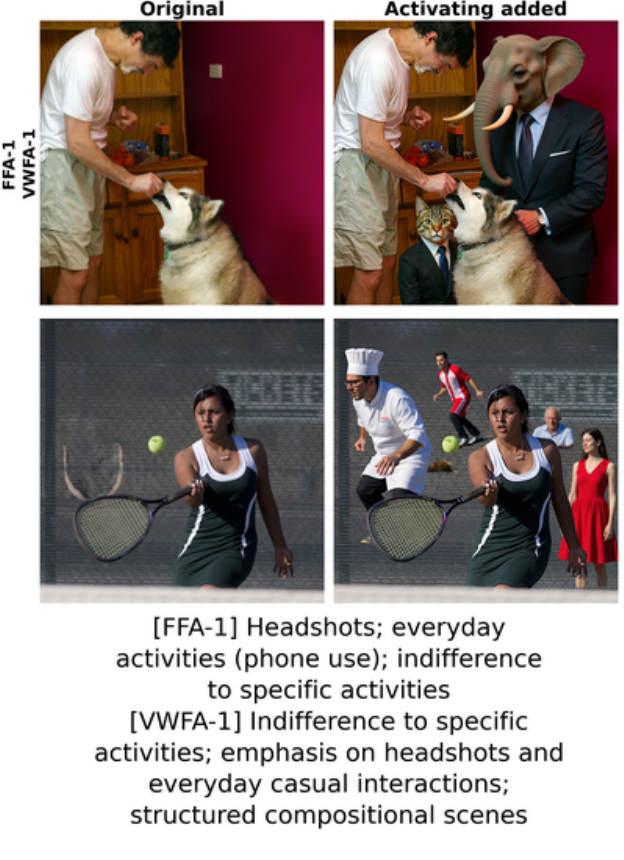}\caption*{\footnotesize voxel\_6023}\end{subfigure}\hfill
\begin{subfigure}[t]{0.18\textwidth}\centering\includegraphics[width=\textwidth]{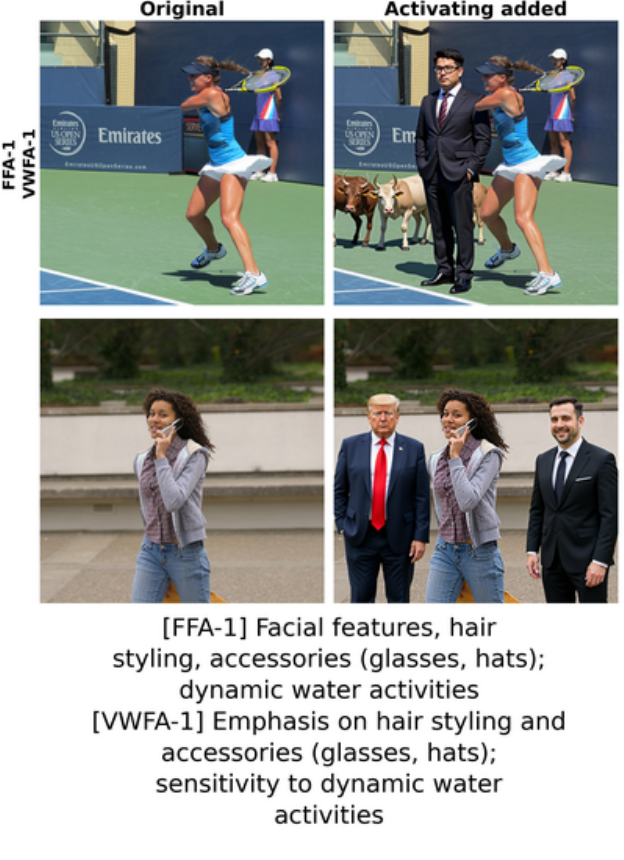}\caption*{\footnotesize voxel\_6035}\end{subfigure}\hfill
\begin{subfigure}[t]{0.18\textwidth}\centering\includegraphics[width=\textwidth]{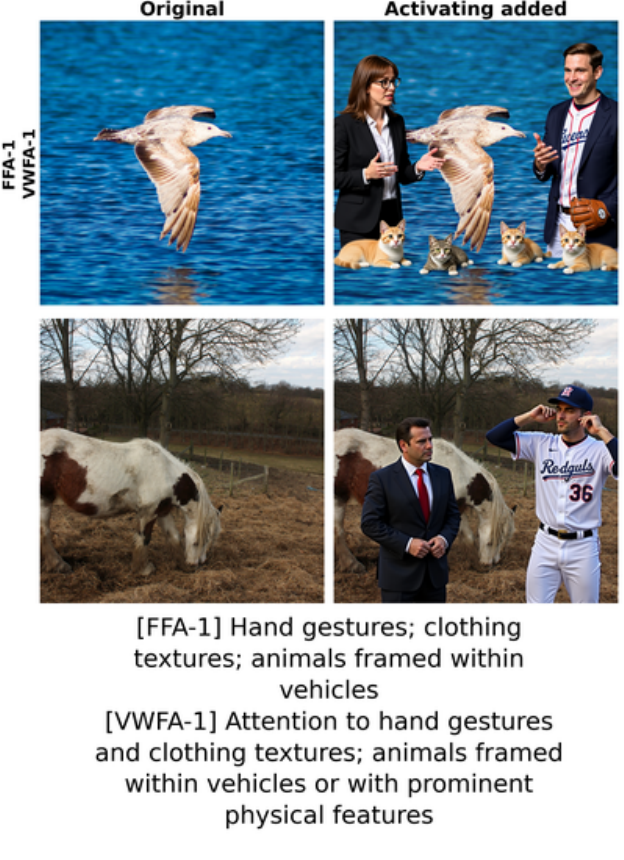}\caption*{\footnotesize voxel\_6036}\end{subfigure}\hfill
\begin{subfigure}[t]{0.18\textwidth}\centering\includegraphics[width=\textwidth]{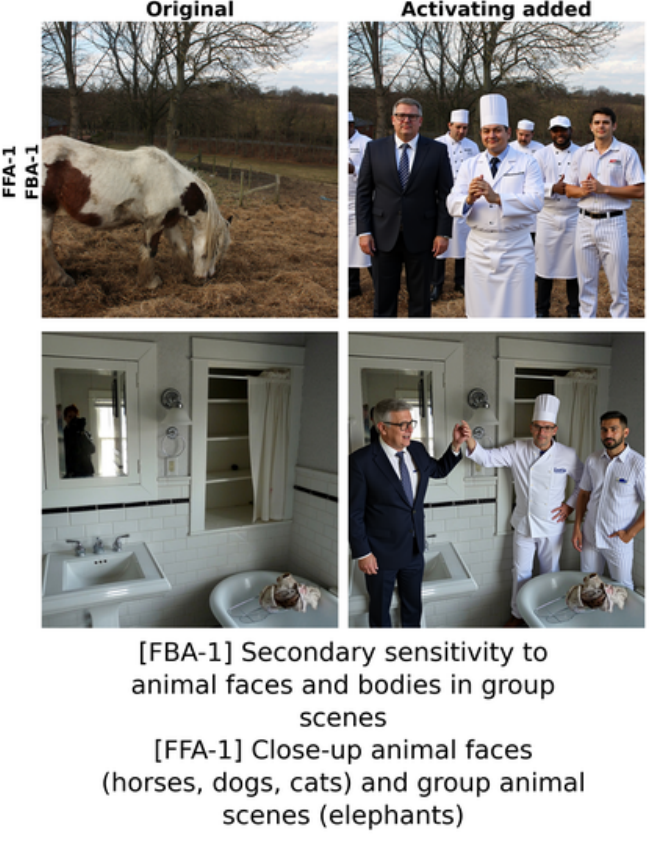}\caption*{\footnotesize voxel\_6063}\end{subfigure}\\[0.8em]
\begin{subfigure}[t]{0.18\textwidth}\centering\includegraphics[width=\textwidth]{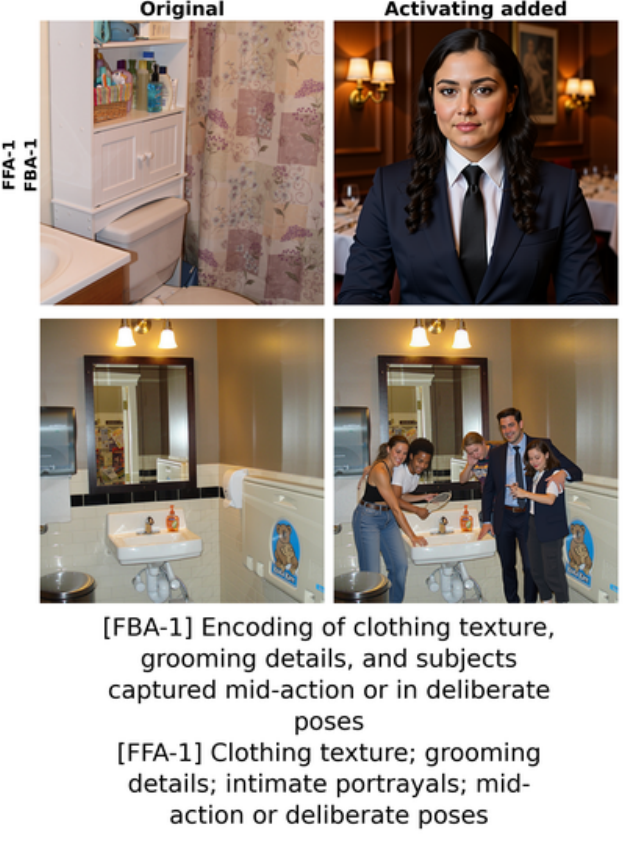}\caption*{\footnotesize voxel\_6064}\end{subfigure}\hfill
\begin{subfigure}[t]{0.18\textwidth}\centering\includegraphics[width=\textwidth]{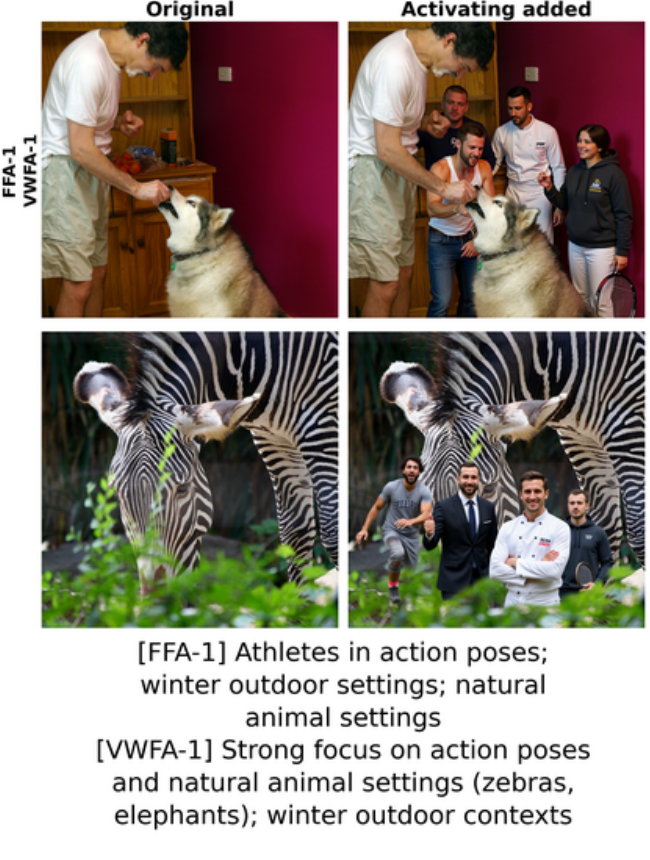}\caption*{\footnotesize voxel\_6258}\end{subfigure}\hfill
\begin{subfigure}[t]{0.18\textwidth}\centering\includegraphics[width=\textwidth]{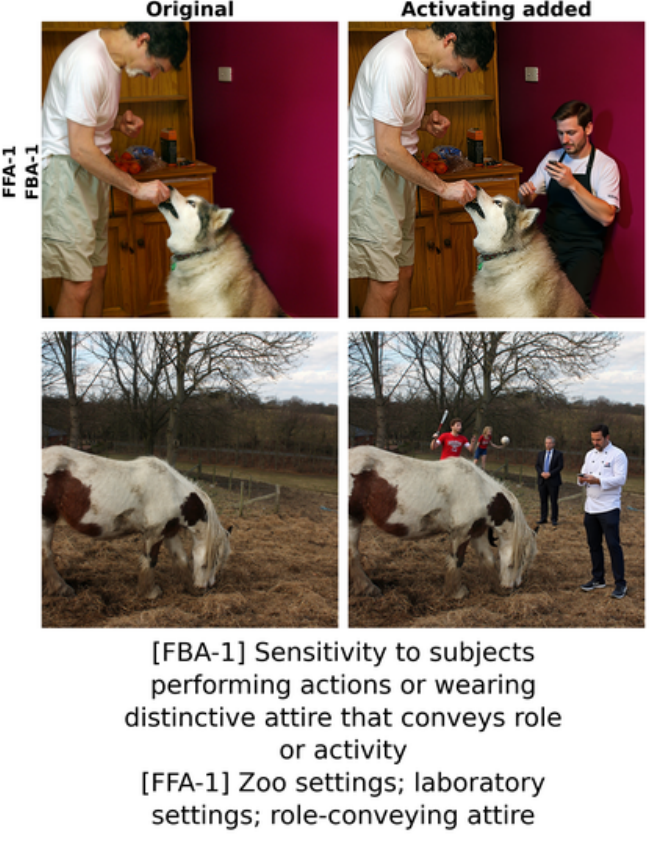}\caption*{\footnotesize voxel\_6287}\end{subfigure}\hfill
\begin{subfigure}[t]{0.18\textwidth}\centering\includegraphics[width=\textwidth]{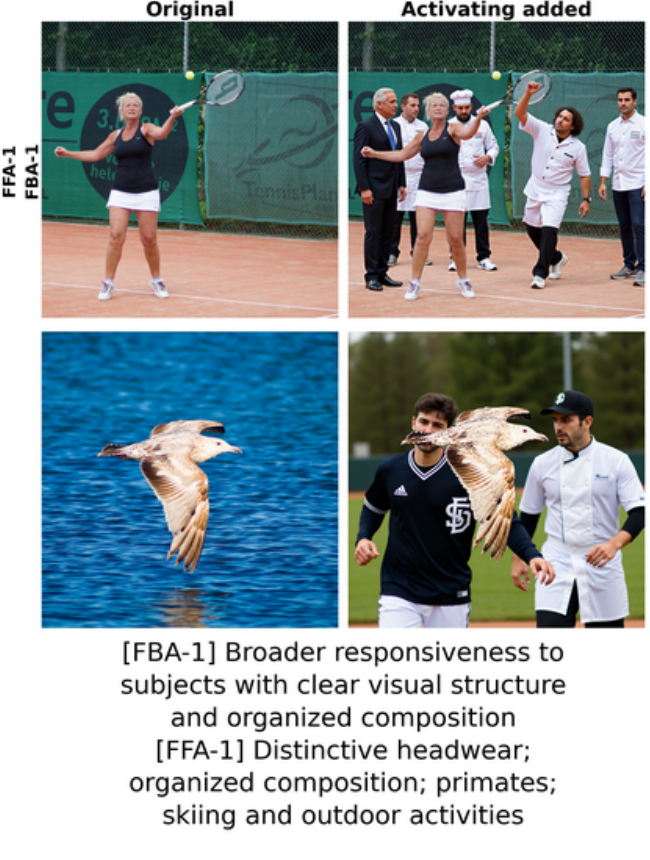}\caption*{\footnotesize voxel\_6483}\end{subfigure}\hfill
\begin{subfigure}[t]{0.18\textwidth}\centering\includegraphics[width=\textwidth]{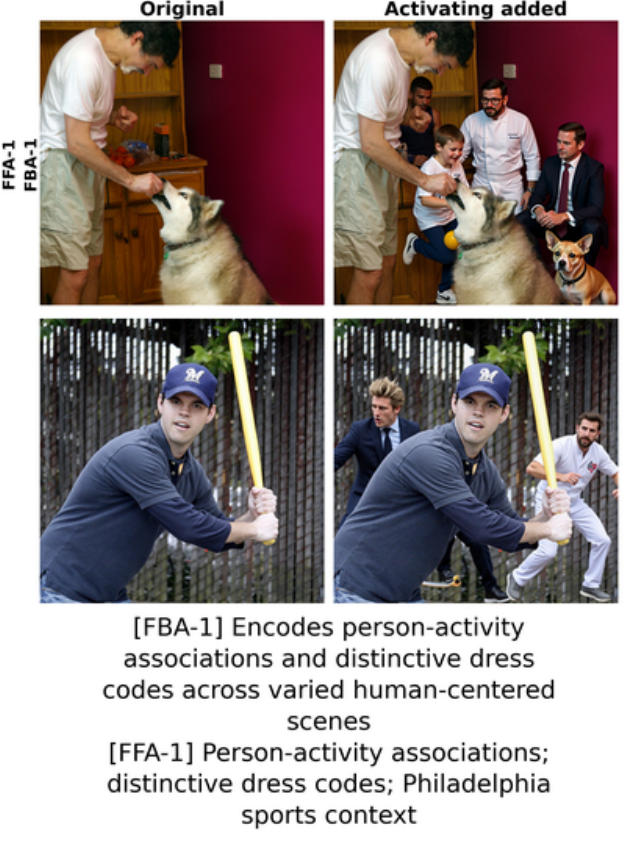}\caption*{\footnotesize voxel\_6681}\end{subfigure}\\[0.8em]
\begin{subfigure}[t]{0.18\textwidth}\centering\includegraphics[width=\textwidth]{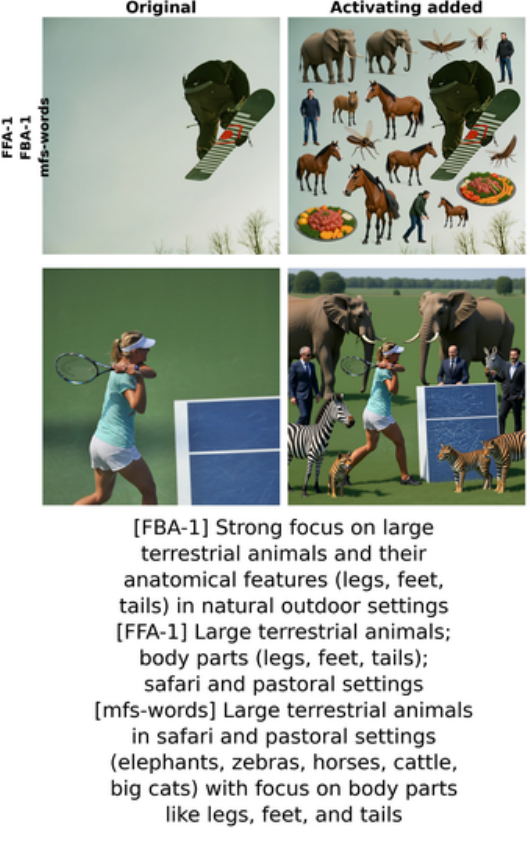}\caption*{\footnotesize voxel\_8646}\end{subfigure}\hfill
\begin{subfigure}[t]{0.18\textwidth}\centering\includegraphics[width=\textwidth]{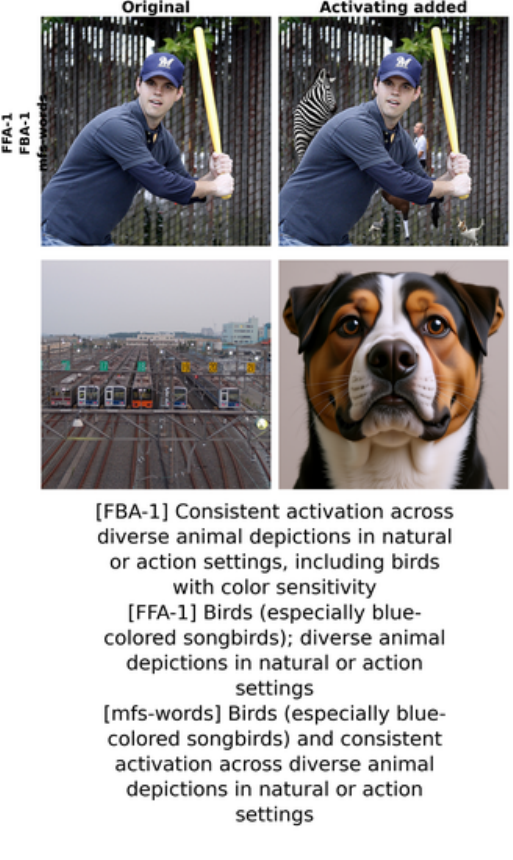}\caption*{\footnotesize voxel\_8754}\end{subfigure}\hfill
\begin{subfigure}[t]{0.18\textwidth}\centering\includegraphics[width=\textwidth]{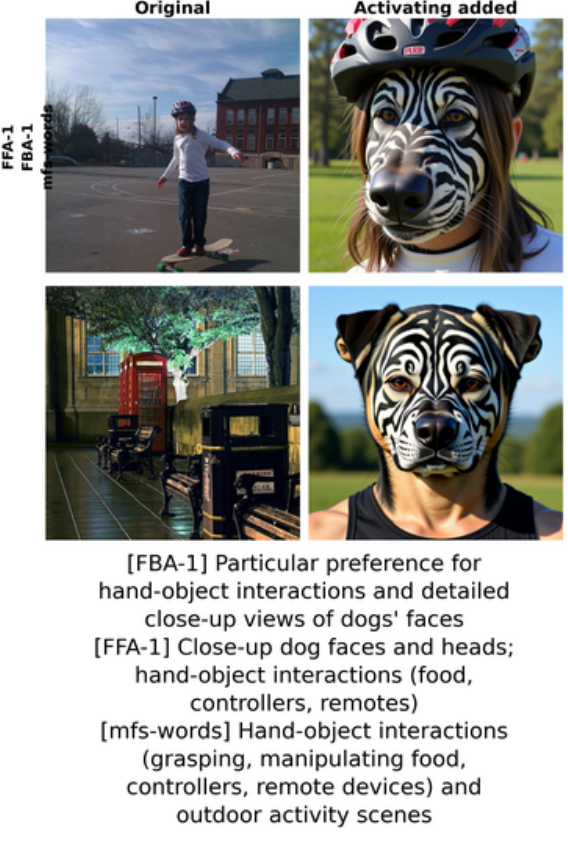}\caption*{\footnotesize voxel\_8755}\end{subfigure}\hfill
\begin{subfigure}[t]{0.18\textwidth}\centering\includegraphics[width=\textwidth]{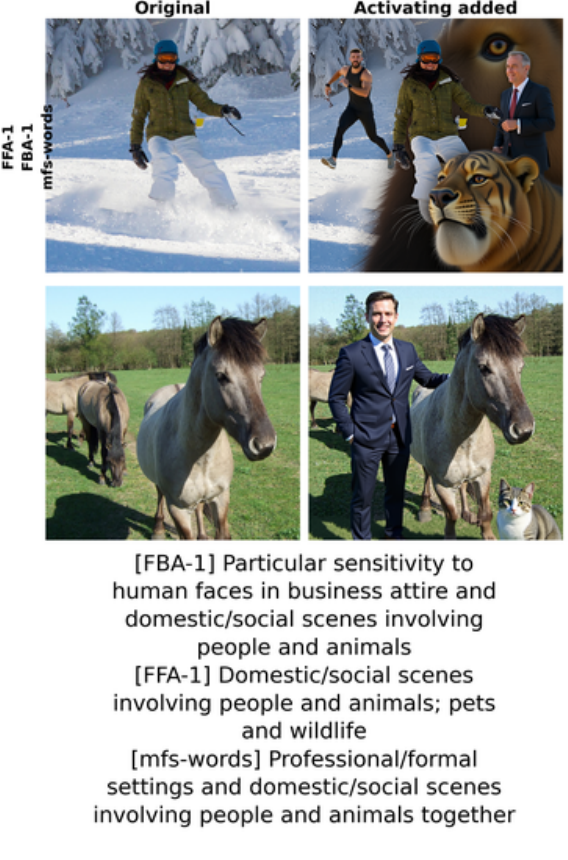}\caption*{\footnotesize voxel\_8758}\end{subfigure}\hfill
\begin{subfigure}[t]{0.18\textwidth}\centering\includegraphics[width=\textwidth]{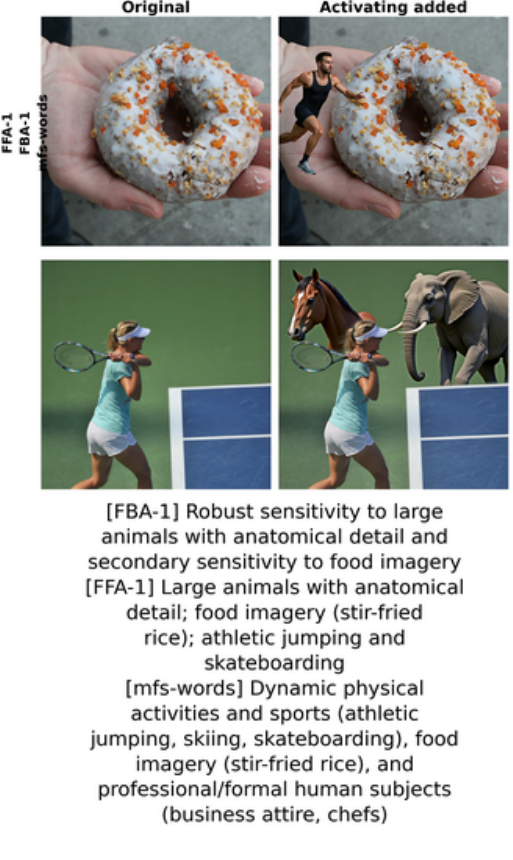}\caption*{\footnotesize voxel\_8759}\end{subfigure}
\caption{\textbf{FFA-1 voxel profiles. (1/2)} \textbf{Shared profile:} \textit{Human faces and upper bodies in professional, formal, or distinctive contexts (business attire, uniforms, sports wear); animals with visible anatomical and facial details; dynamic physical activities and action scenes; sensitivity to clothing, facial features, and compositional structure.}}
\label{app:fig:ffa1-profiles}
\end{figure}

\begin{figure}[!ht]
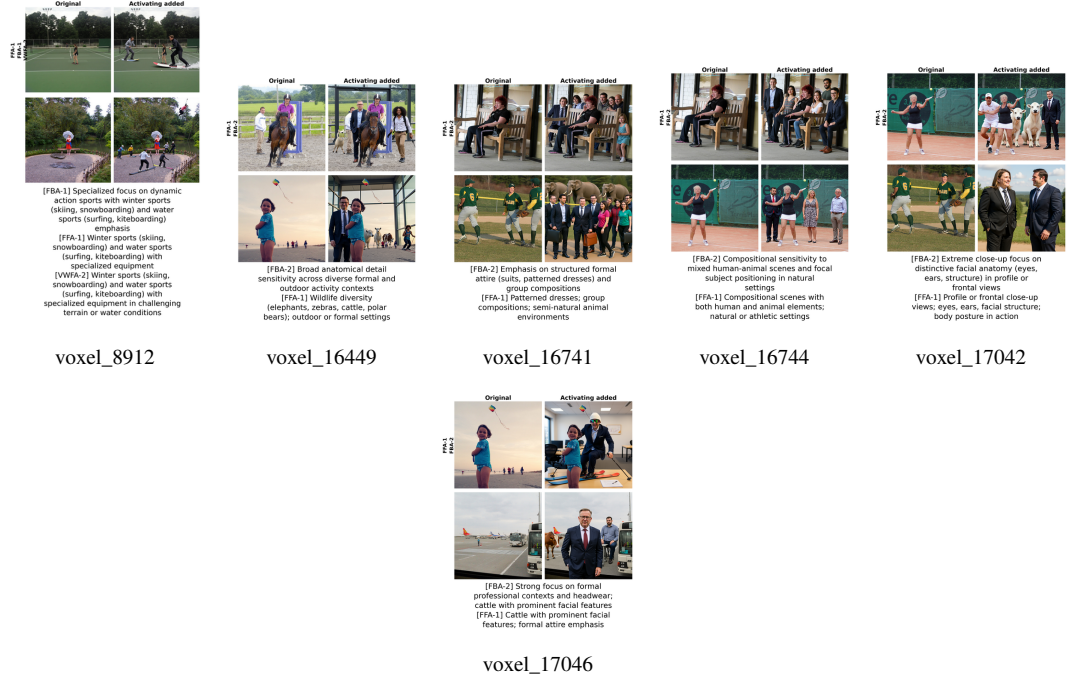
\ContinuedFloat
\centering
\begin{subfigure}[t]{0.18\textwidth}\centering\includegraphics[width=\textwidth]{figs/fig-app-voxel-profiles/FFA-1/voxel_008912_top2.pdf}\caption*{\footnotesize voxel\_8912}\end{subfigure}\hfill
\begin{subfigure}[t]{0.18\textwidth}\centering\includegraphics[width=\textwidth]{figs/fig-app-voxel-profiles/FFA-1/voxel_016449_top2.pdf}\caption*{\footnotesize voxel\_16449}\end{subfigure}\hfill
\begin{subfigure}[t]{0.18\textwidth}\centering\includegraphics[width=\textwidth]{figs/fig-app-voxel-profiles/FFA-1/voxel_016741_top2.pdf}\caption*{\footnotesize voxel\_16741}\end{subfigure}\hfill
\begin{subfigure}[t]{0.18\textwidth}\centering\includegraphics[width=\textwidth]{figs/fig-app-voxel-profiles/FFA-1/voxel_016744_top2.pdf}\caption*{\footnotesize voxel\_16744}\end{subfigure}\hfill
\begin{subfigure}[t]{0.18\textwidth}\centering\includegraphics[width=\textwidth]{figs/fig-app-voxel-profiles/FFA-1/voxel_017042_top2.pdf}\caption*{\footnotesize voxel\_17042}\end{subfigure}\\[0.8em]
\begin{subfigure}[t]{0.18\textwidth}\centering\includegraphics[width=\textwidth]{figs/fig-app-voxel-profiles/FFA-1/voxel_017046_top2.pdf}\caption*{\footnotesize voxel\_17046}\end{subfigure}
\caption{\textbf{FFA-1 voxel profiles (continued).}}
\end{figure}

\begin{figure}[!ht]
\centering
\begin{subfigure}[t]{0.18\textwidth}\centering\includegraphics[width=\textwidth]{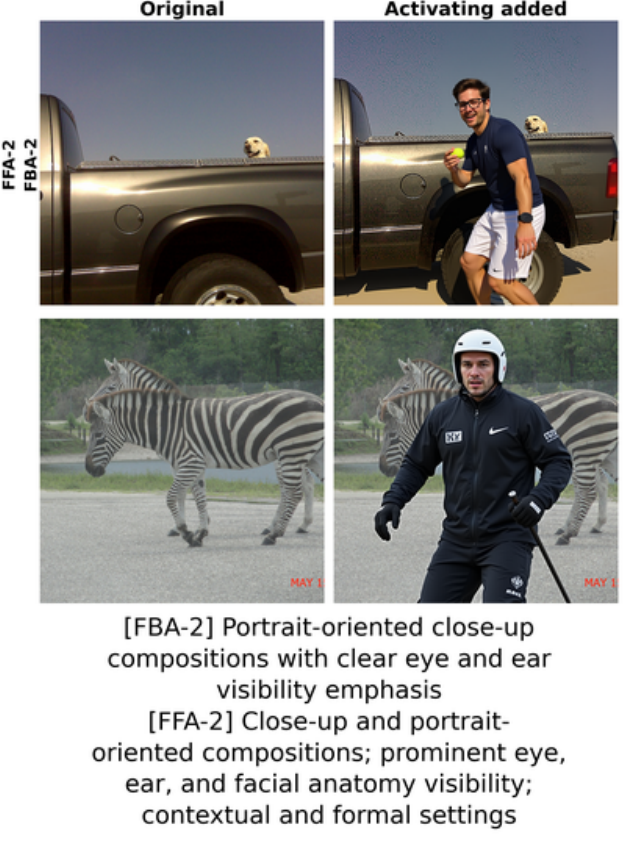}\caption*{\footnotesize voxel\_20662}\end{subfigure}\hfill
\begin{subfigure}[t]{0.18\textwidth}\centering\includegraphics[width=\textwidth]{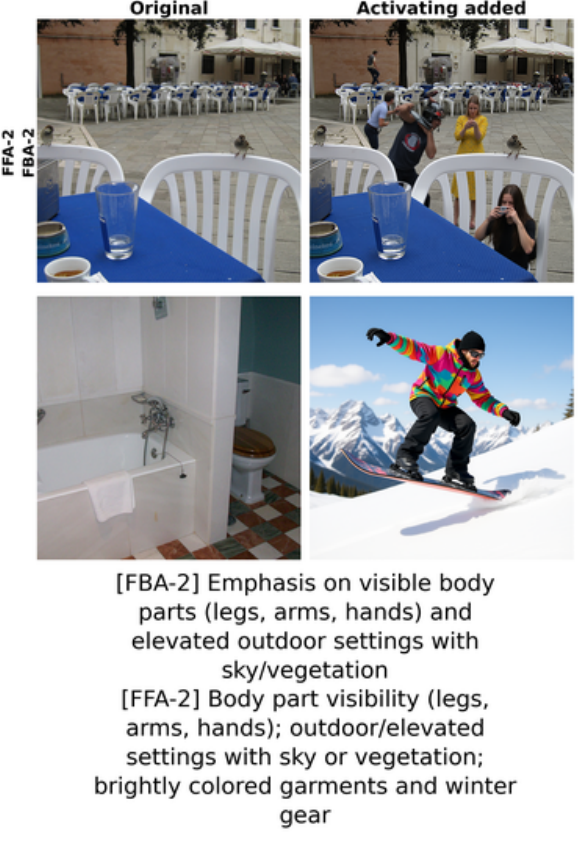}\caption*{\footnotesize voxel\_20732}\end{subfigure}\hfill
\begin{subfigure}[t]{0.18\textwidth}\centering\includegraphics[width=\textwidth]{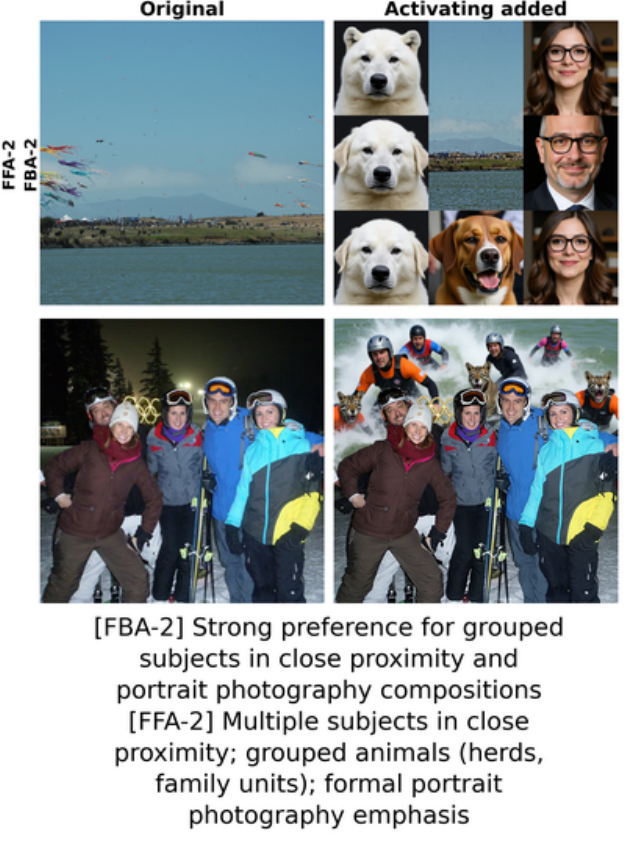}\caption*{\footnotesize voxel\_20737}\end{subfigure}\hfill
\begin{subfigure}[t]{0.18\textwidth}\centering\includegraphics[width=\textwidth]{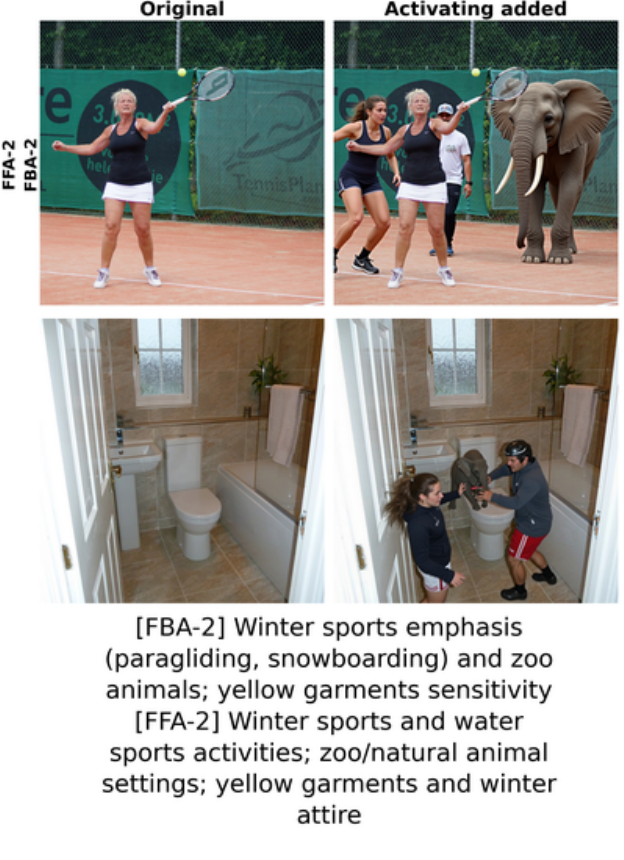}\caption*{\footnotesize voxel\_20794}\end{subfigure}\hfill
\begin{subfigure}[t]{0.18\textwidth}\centering\includegraphics[width=\textwidth]{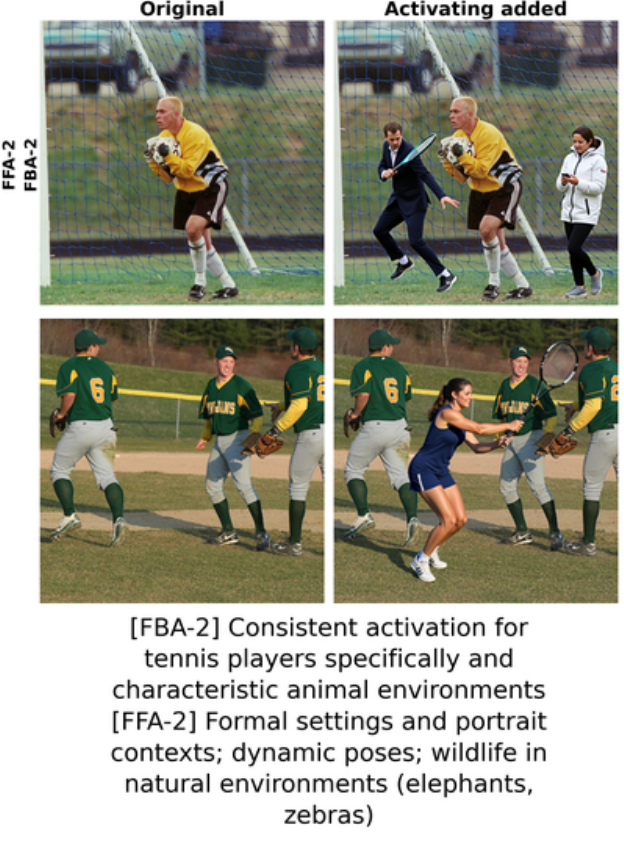}\caption*{\footnotesize voxel\_20795}\end{subfigure}
\caption{\textbf{FFA-2 voxel profiles.} \textbf{Shared profile:} \textit{Strong responsiveness to human figures and animals in dynamic or characteristic contexts, with sensitivity to distinctive visual features including clothing, body positioning, and facial/anatomical details. Consistent activation for active subjects (athletes, people in motion, wildlife) and outdoor or natural settings.}}
\label{app:fig:ffa2-profiles}
\end{figure}

\begin{figure}[!ht]
\centering
\begin{subfigure}[t]{0.18\textwidth}\centering\includegraphics[width=\textwidth]{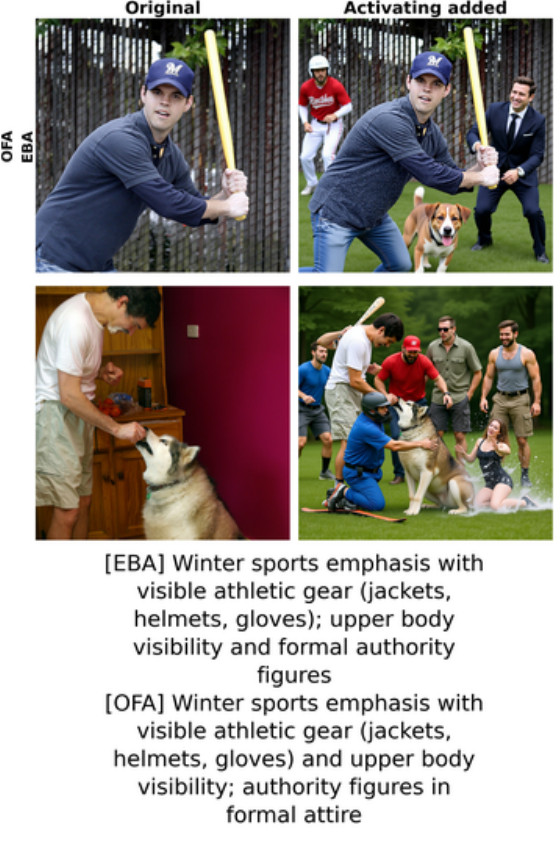}\caption*{\footnotesize voxel\_2271}\end{subfigure}\hfill
\begin{subfigure}[t]{0.18\textwidth}\centering\includegraphics[width=\textwidth]{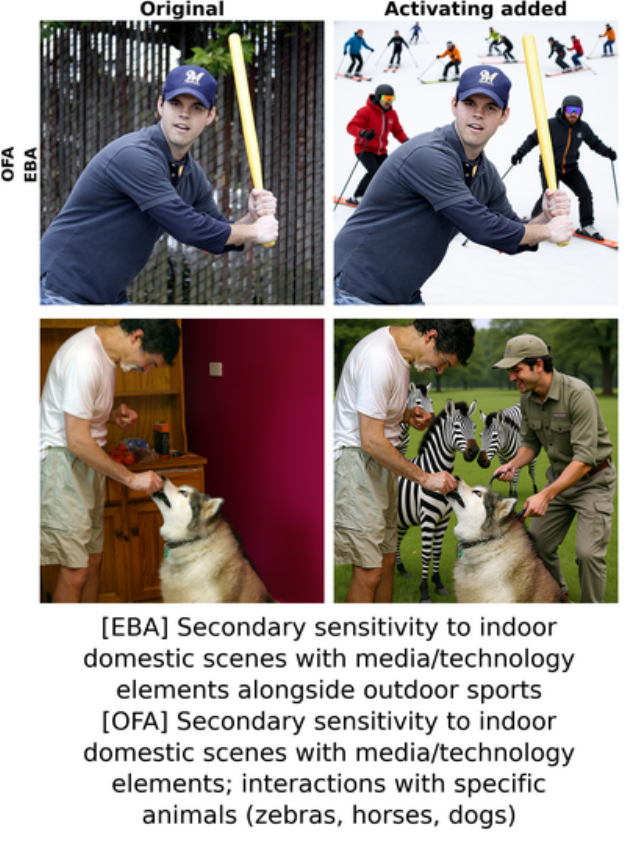}\caption*{\footnotesize voxel\_2528}\end{subfigure}\hfill
\begin{subfigure}[t]{0.18\textwidth}\centering\includegraphics[width=\textwidth]{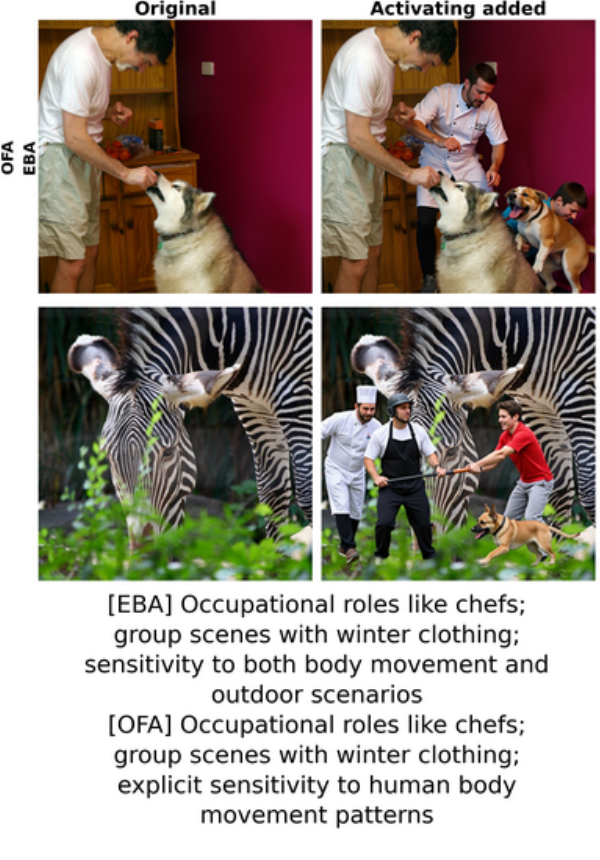}\caption*{\footnotesize voxel\_2529}\end{subfigure}\hfill
\begin{subfigure}[t]{0.18\textwidth}\centering\includegraphics[width=\textwidth]{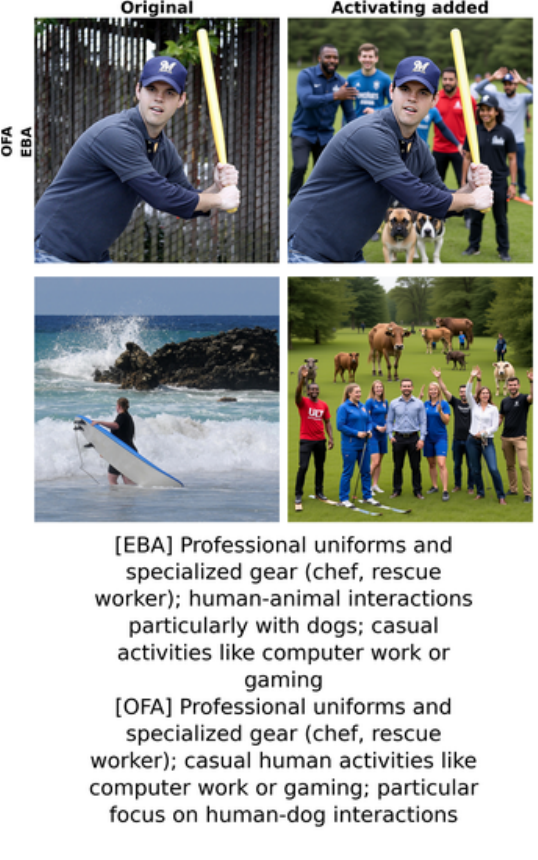}\caption*{\footnotesize voxel\_3094}\end{subfigure}\hfill
\begin{subfigure}[t]{0.18\textwidth}\centering\includegraphics[width=\textwidth]{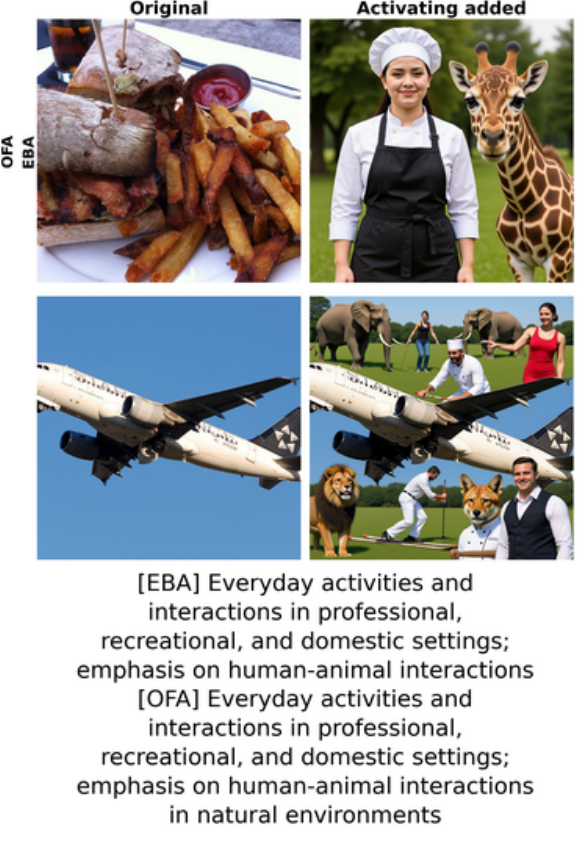}\caption*{\footnotesize voxel\_3095}\end{subfigure}
\caption{\textbf{OFA voxel profiles.} \textbf{Shared profile:} \textit{Dynamic human figures engaged in physical activities, sports, and movement across diverse contexts (skiing, baseball, skateboarding, tennis, surfing), combined with sensitivity to athletic/specialized gear and apparel, as well as animals in motion and outdoor/active scenarios.}}
\label{app:fig:ofa-profiles}
\end{figure}

\begin{figure}[!ht]
\centering
\begin{subfigure}[t]{0.18\textwidth}\centering\includegraphics[width=\textwidth]{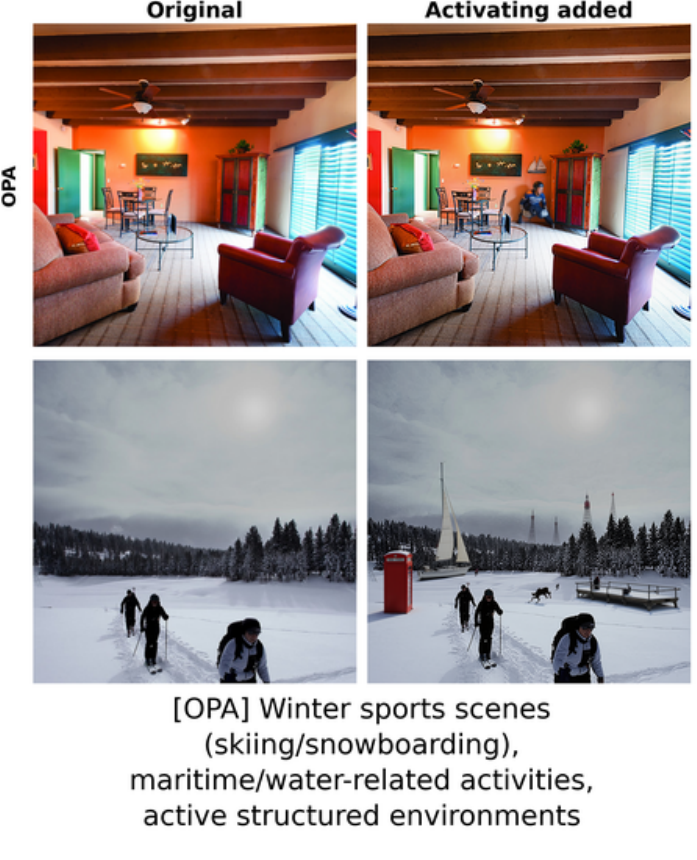}\caption*{\footnotesize voxel\_3866}\end{subfigure}\hfill
\begin{subfigure}[t]{0.18\textwidth}\centering\includegraphics[width=\textwidth]{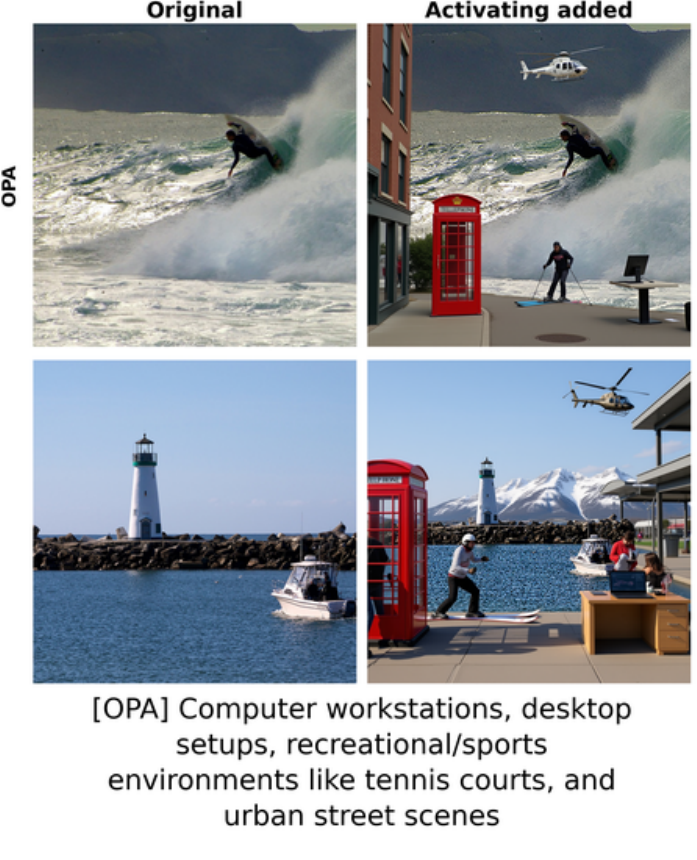}\caption*{\footnotesize voxel\_3867}\end{subfigure}\hfill
\begin{subfigure}[t]{0.18\textwidth}\centering\includegraphics[width=\textwidth]{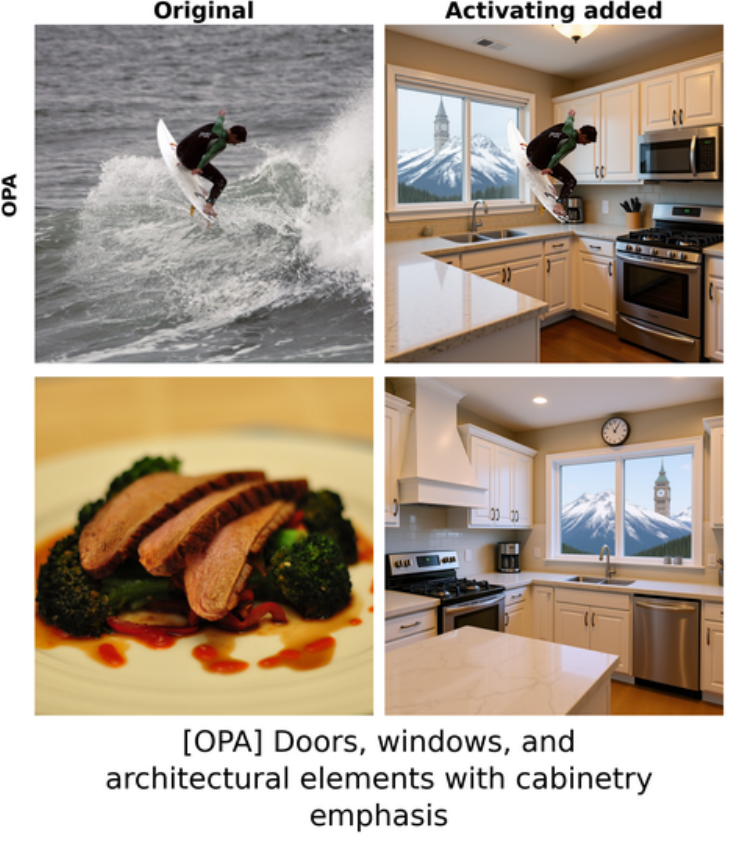}\caption*{\footnotesize voxel\_4190}\end{subfigure}\hfill
\begin{subfigure}[t]{0.18\textwidth}\centering\includegraphics[width=\textwidth]{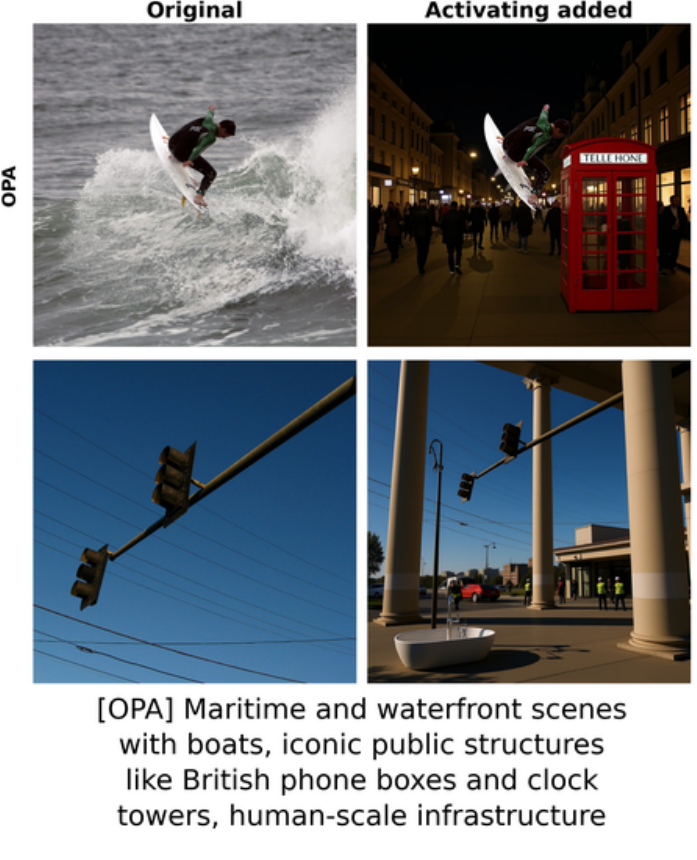}\caption*{\footnotesize voxel\_4191}\end{subfigure}\hfill
\begin{subfigure}[t]{0.18\textwidth}\centering\includegraphics[width=\textwidth]{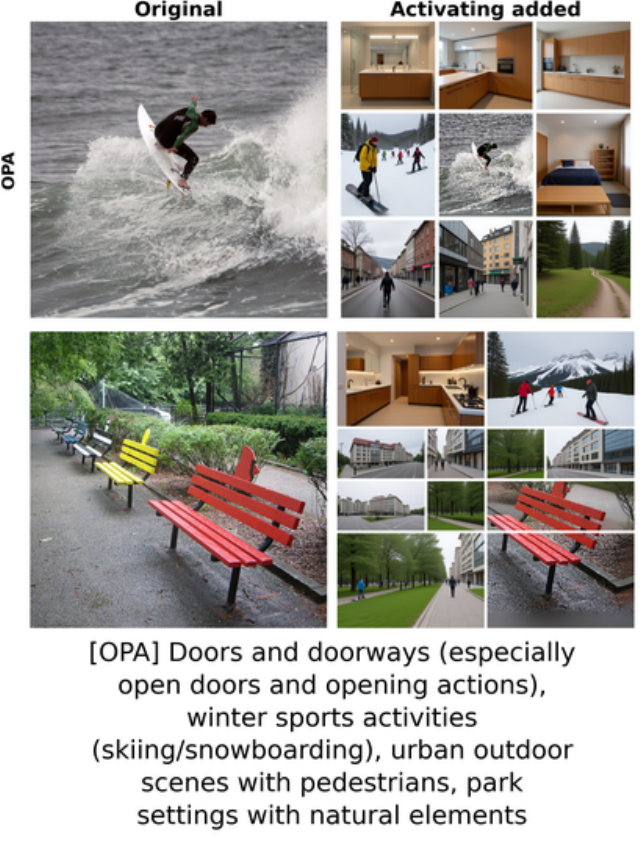}\caption*{\footnotesize voxel\_4194}\end{subfigure}
\caption{\textbf{OPA voxel profiles.} \textbf{Shared profile:} \textit{Strong preference for structured, functional indoor spaces (particularly kitchens and bathrooms with fixtures), organized domestic interiors with clear architectural elements, and human-made environments with purposeful design and infrastructure.}}
\label{app:fig:opa-profiles}
\end{figure}

\begin{figure}[!ht]
\centering
\begin{subfigure}[t]{0.18\textwidth}\centering\includegraphics[width=\textwidth]{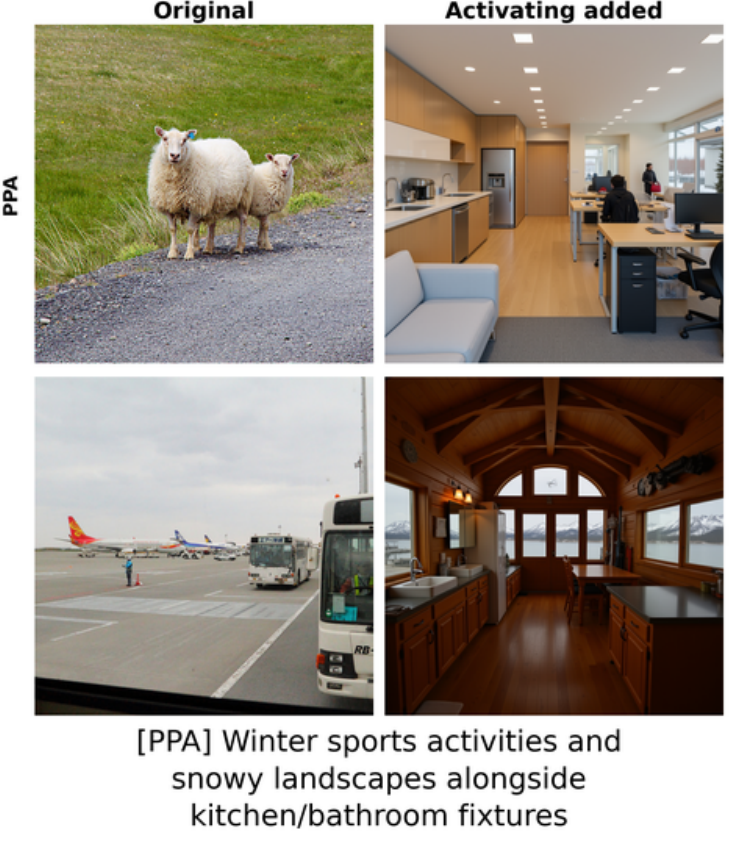}\caption*{\footnotesize voxel\_8629}\end{subfigure}\hfill
\begin{subfigure}[t]{0.18\textwidth}\centering\includegraphics[width=\textwidth]{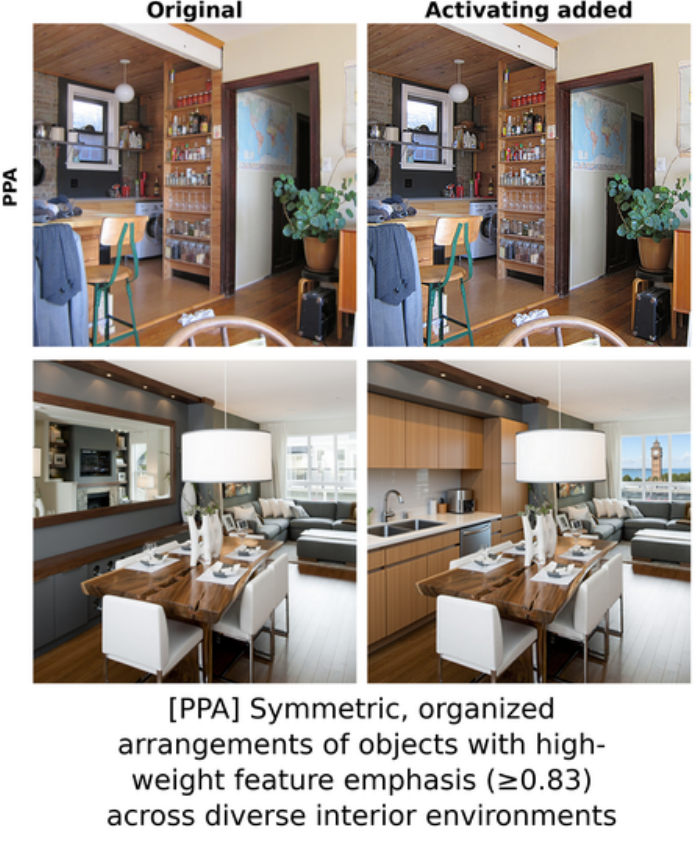}\caption*{\footnotesize voxel\_8734}\end{subfigure}\hfill
\begin{subfigure}[t]{0.18\textwidth}\centering\includegraphics[width=\textwidth]{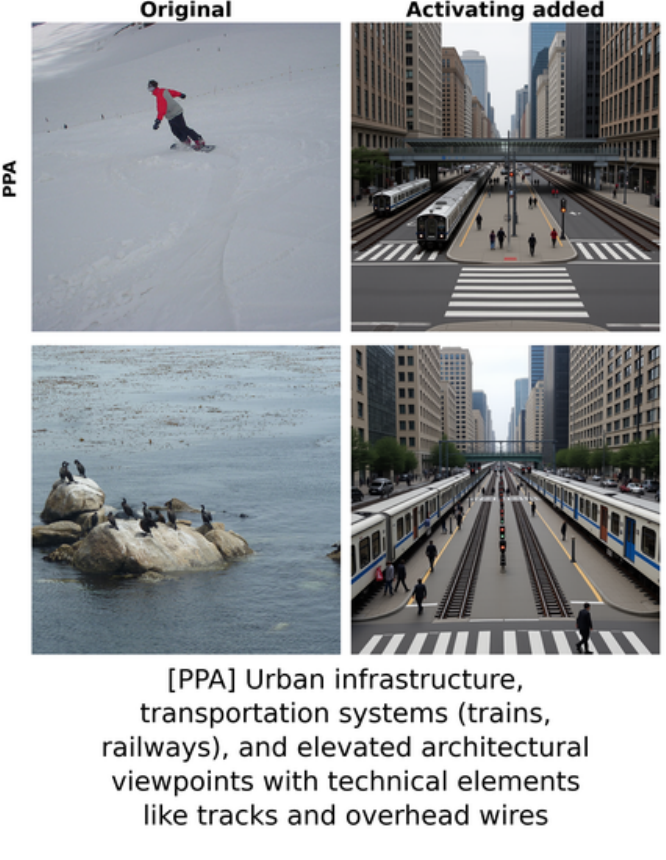}\caption*{\footnotesize voxel\_8851}\end{subfigure}\hfill
\begin{subfigure}[t]{0.18\textwidth}\centering\includegraphics[width=\textwidth]{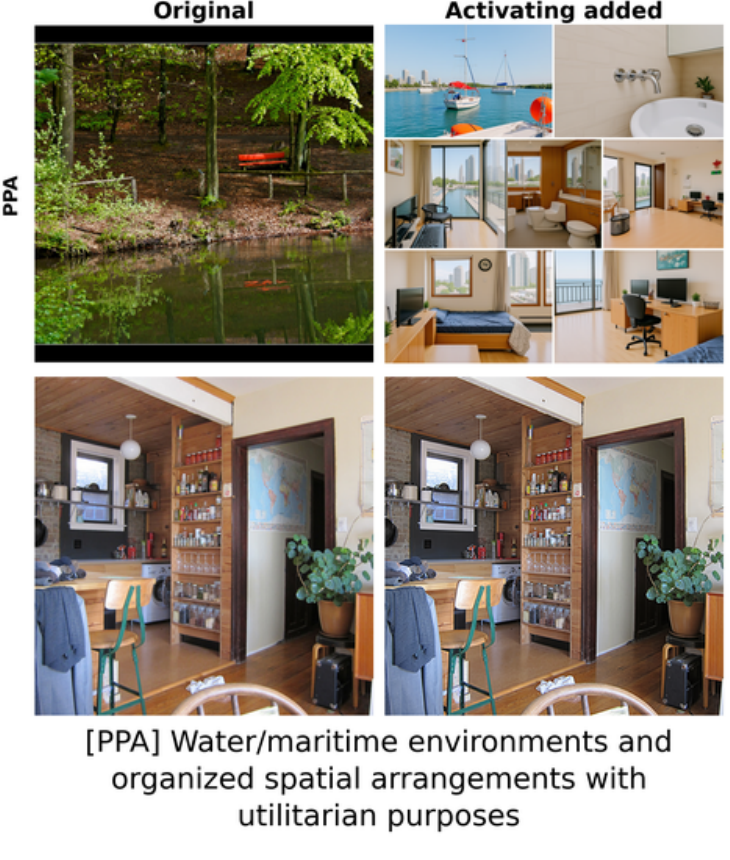}\caption*{\footnotesize voxel\_8862}\end{subfigure}\hfill
\begin{subfigure}[t]{0.18\textwidth}\centering\includegraphics[width=\textwidth]{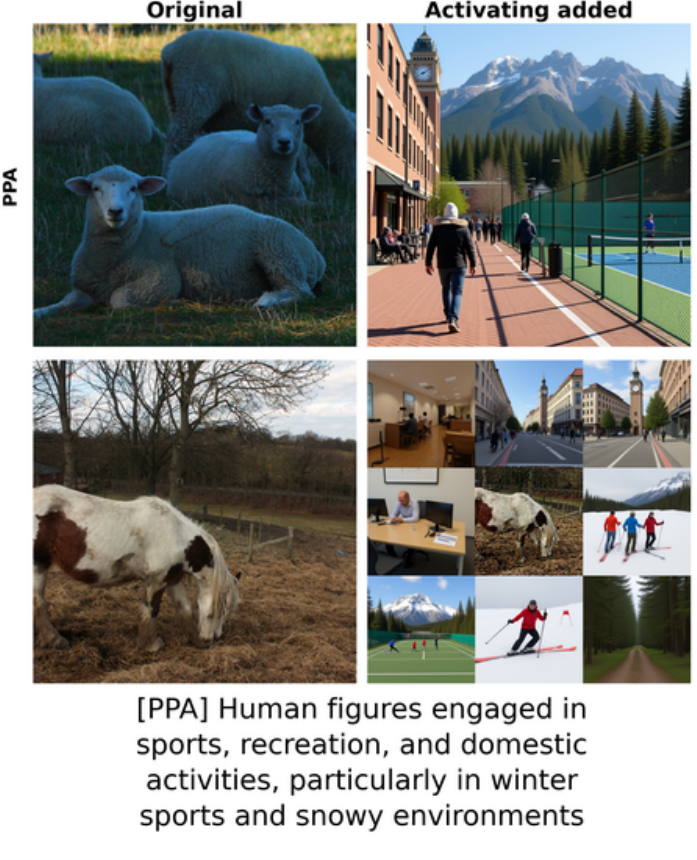}\caption*{\footnotesize voxel\_8869}\end{subfigure}
\caption{\textbf{PPA voxel profiles.} \textbf{Shared profile:} \textit{Preference for structured, organized environments with clear spatial hierarchies and functional purposes, including both indoor domestic/functional spaces (kitchens, bathrooms, offices) and outdoor scenes with defined architectural or infrastructural elements.}}
\label{app:fig:ppa-profiles}
\end{figure}

\begin{figure}[!ht]
\centering
\begin{subfigure}[t]{0.18\textwidth}\centering\includegraphics[width=\textwidth]{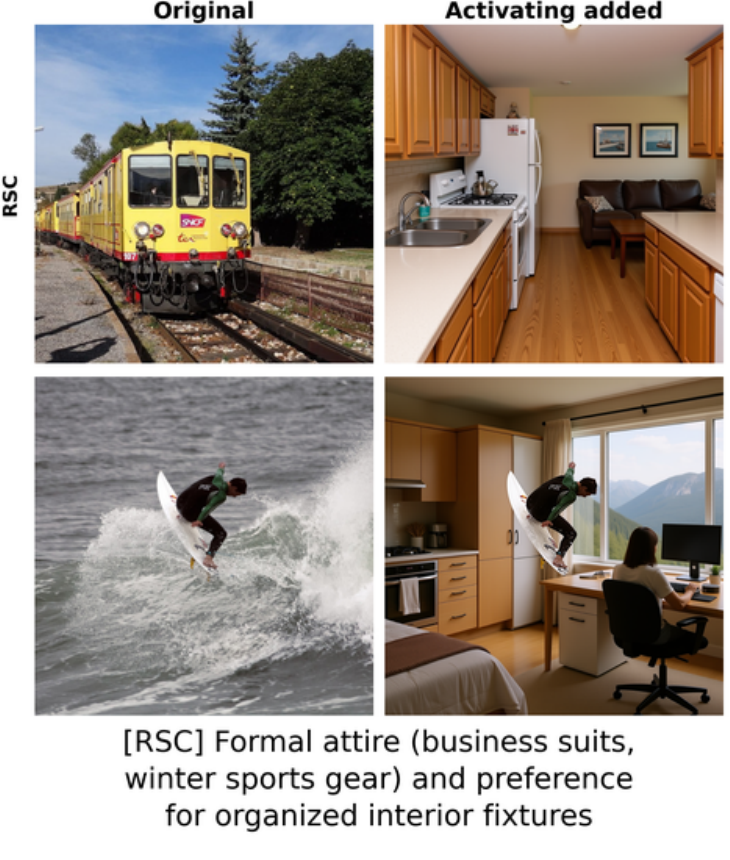}\caption*{\footnotesize voxel\_7650}\end{subfigure}\hfill
\begin{subfigure}[t]{0.18\textwidth}\centering\includegraphics[width=\textwidth]{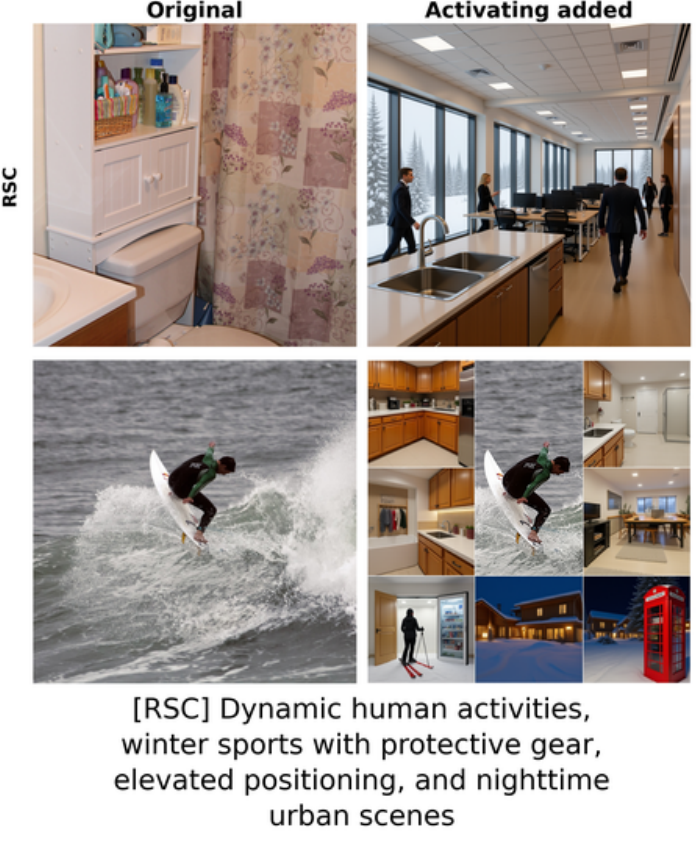}\caption*{\footnotesize voxel\_7651}\end{subfigure}\hfill
\begin{subfigure}[t]{0.18\textwidth}\centering\includegraphics[width=\textwidth]{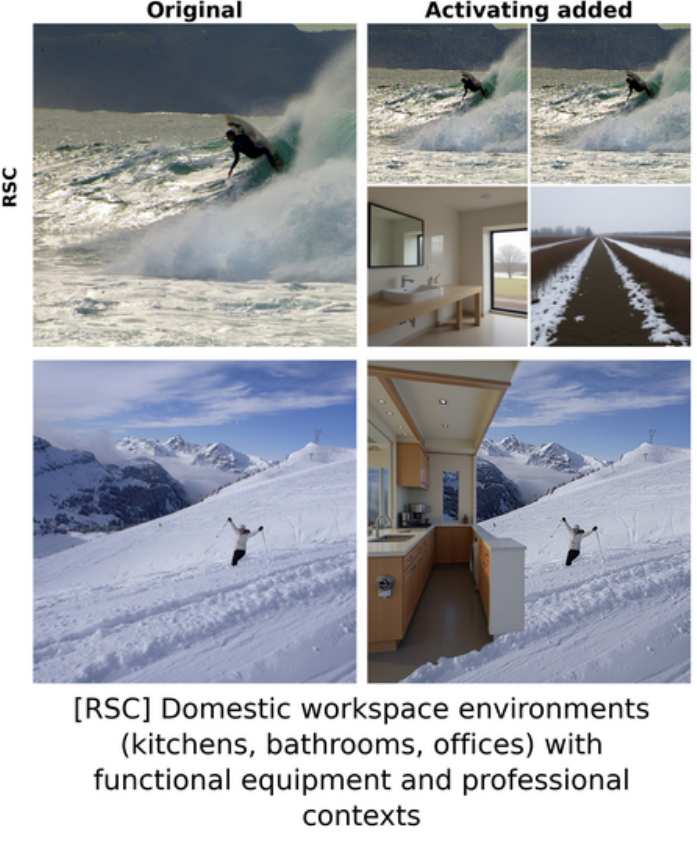}\caption*{\footnotesize voxel\_7835}\end{subfigure}\hfill
\begin{subfigure}[t]{0.18\textwidth}\centering\includegraphics[width=\textwidth]{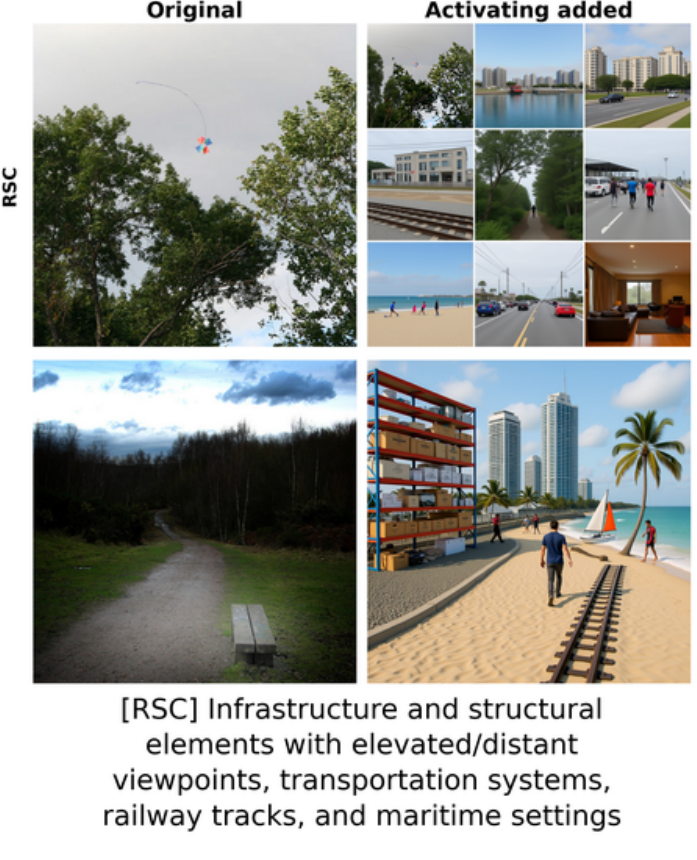}\caption*{\footnotesize voxel\_8015}\end{subfigure}\hfill
\begin{subfigure}[t]{0.18\textwidth}\centering\includegraphics[width=\textwidth]{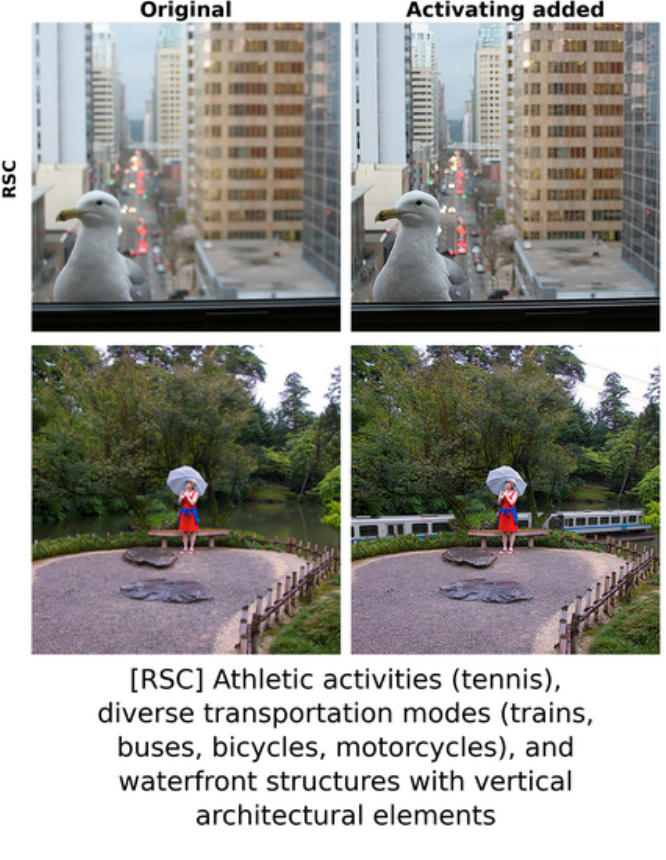}\caption*{\footnotesize voxel\_18720}\end{subfigure}
\caption{\textbf{RSC voxel profiles.} \textbf{Shared profile:} \textit{Structured environments with organized spatial elements, human figures in various contexts, and architectural or infrastructural frameworks. All voxels show sensitivity to both indoor functional spaces and outdoor scenes with clear organizational or structural components.}}
\label{app:fig:rsc-profiles}
\end{figure}

\begin{figure}[!ht]
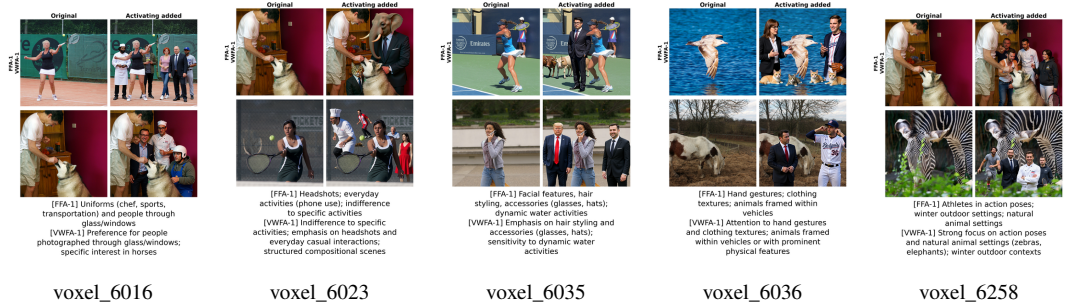

\centering
\begin{subfigure}[t]{0.18\textwidth}\centering\includegraphics[width=\textwidth]{figs/fig-app-voxel-profiles/VWFA-1/voxel_006016_top2.pdf}\caption*{\footnotesize voxel\_6016}\end{subfigure}\hfill
\begin{subfigure}[t]{0.18\textwidth}\centering\includegraphics[width=\textwidth]{figs/fig-app-voxel-profiles/VWFA-1/voxel_006023_top2.pdf}\caption*{\footnotesize voxel\_6023}\end{subfigure}\hfill
\begin{subfigure}[t]{0.18\textwidth}\centering\includegraphics[width=\textwidth]{figs/fig-app-voxel-profiles/VWFA-1/voxel_006035_top2.pdf}\caption*{\footnotesize voxel\_6035}\end{subfigure}\hfill
\begin{subfigure}[t]{0.18\textwidth}\centering\includegraphics[width=\textwidth]{figs/fig-app-voxel-profiles/VWFA-1/voxel_006036_top2.pdf}\caption*{\footnotesize voxel\_6036}\end{subfigure}\hfill
\begin{subfigure}[t]{0.18\textwidth}\centering\includegraphics[width=\textwidth]{figs/fig-app-voxel-profiles/VWFA-1/voxel_006258_top2.pdf}\caption*{\footnotesize voxel\_6258}\end{subfigure}
\caption{\textbf{VWFA-1 voxel profiles.} \textbf{Shared profile:} \textit{Strong preference for human subjects across professional, formal, and athletic contexts. Consistent sensitivity to facial features, clothing details (particularly formal wear like suits and uniforms), and human upper bodies/portraits. Secondary responsiveness to animals with prominent facial features.}}
\label{app:fig:vwfa1-profiles}
\end{figure}

\begin{figure}[!ht]
\centering
\begin{subfigure}[t]{0.18\textwidth}\centering\includegraphics[width=\textwidth]{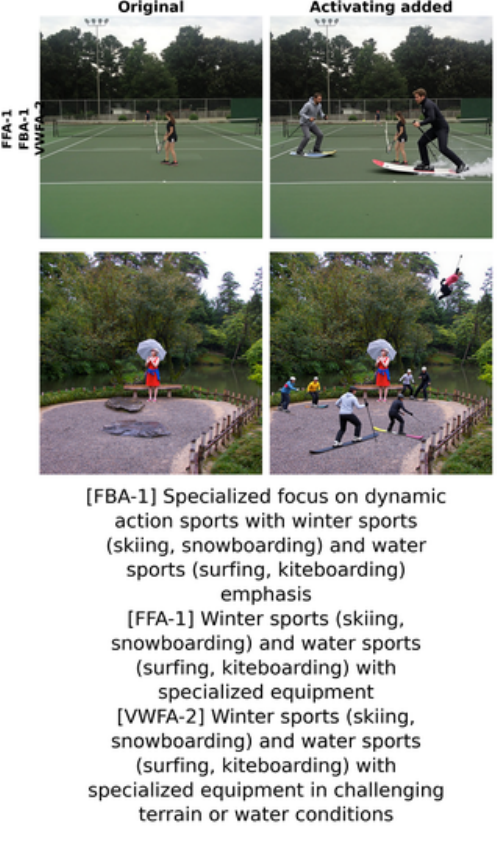}\caption*{\footnotesize voxel\_8912}\end{subfigure}\hfill
\begin{subfigure}[t]{0.18\textwidth}\centering\includegraphics[width=\textwidth]{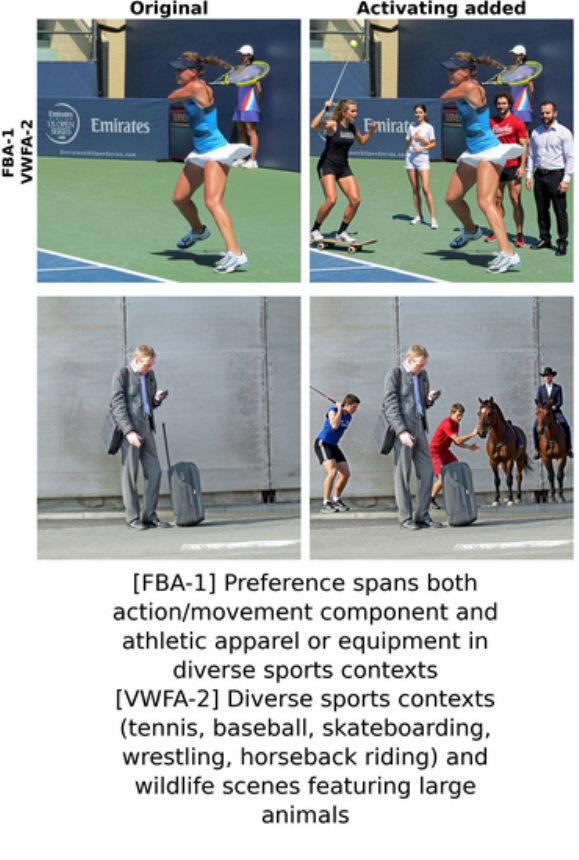}\caption*{\footnotesize voxel\_8915}\end{subfigure}\hfill
\begin{subfigure}[t]{0.18\textwidth}\centering\includegraphics[width=\textwidth]{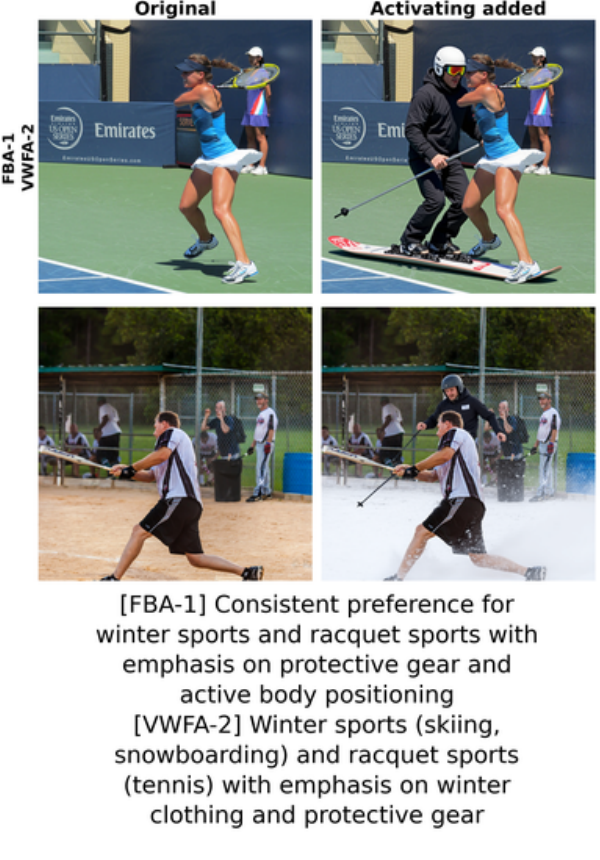}\caption*{\footnotesize voxel\_9023}\end{subfigure}\hfill
\begin{subfigure}[t]{0.18\textwidth}\centering\includegraphics[width=\textwidth]{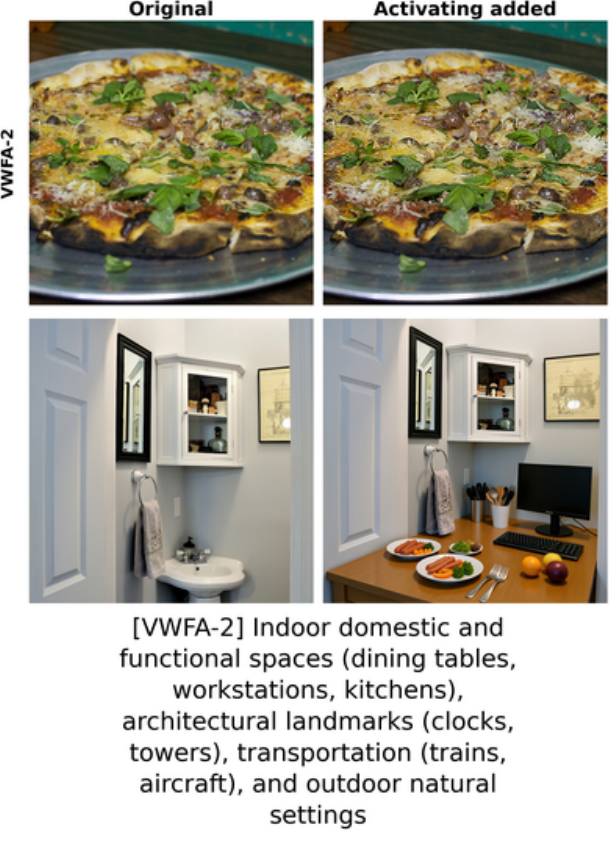}\caption*{\footnotesize voxel\_16964}\end{subfigure}\hfill
\begin{subfigure}[t]{0.18\textwidth}\centering\includegraphics[width=\textwidth]{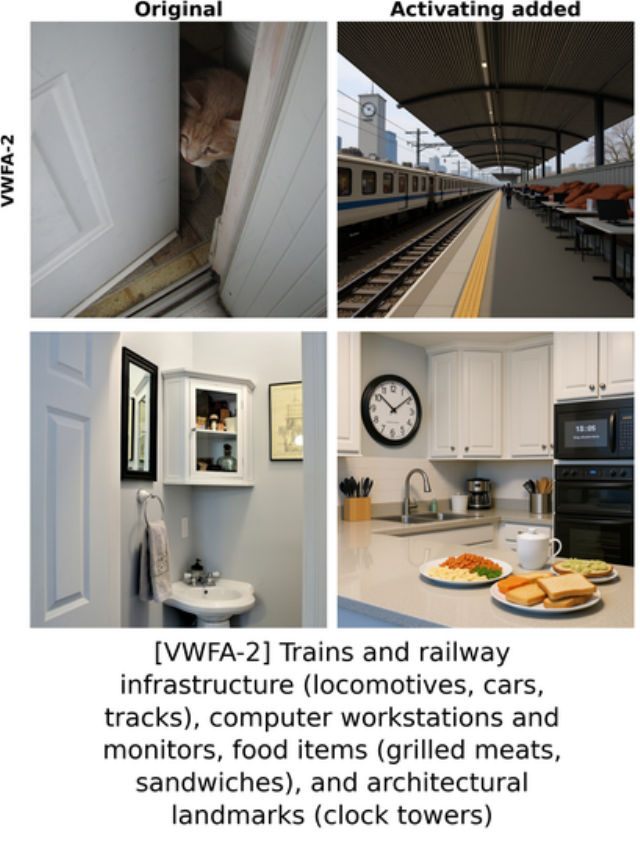}\caption*{\footnotesize voxel\_16974}\end{subfigure}
\caption{\textbf{VWFA-2 voxel profiles.} \textbf{Shared profile:} \textit{Dynamic action and athletic activities with people in motion, wearing sport-specific attire and protective gear, engaged in physical movement and sports contexts.}}
\label{app:fig:vwfa2-profiles}
\end{figure}

\begin{figure}[!ht]
\centering
\begin{subfigure}[t]{0.18\textwidth}\centering\includegraphics[width=\textwidth]{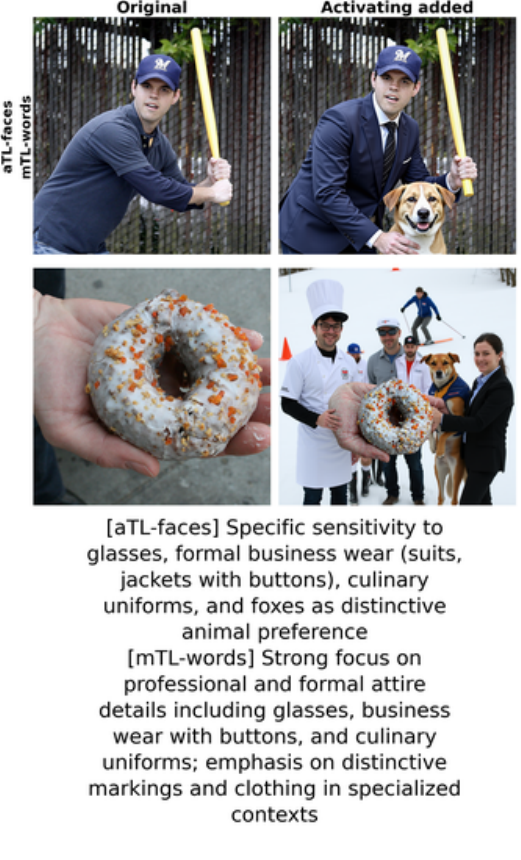}\caption*{\footnotesize voxel\_9865}\end{subfigure}\hfill
\begin{subfigure}[t]{0.18\textwidth}\centering\includegraphics[width=\textwidth]{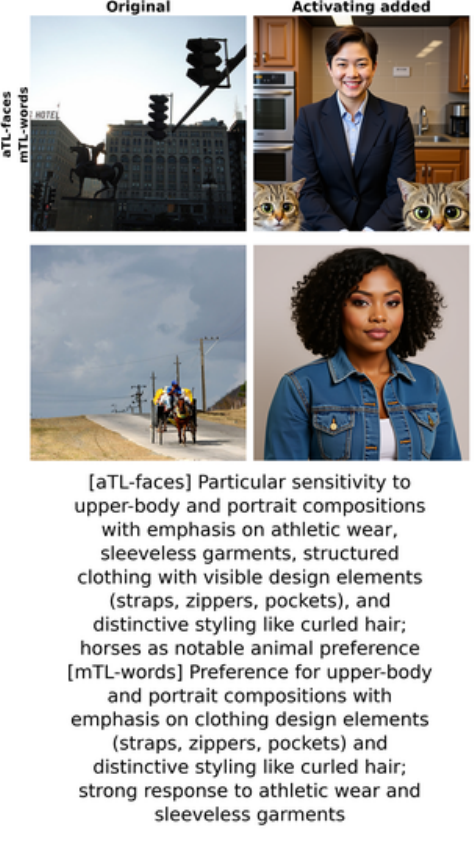}\caption*{\footnotesize voxel\_9866}\end{subfigure}\hfill
\begin{subfigure}[t]{0.18\textwidth}\centering\includegraphics[width=\textwidth]{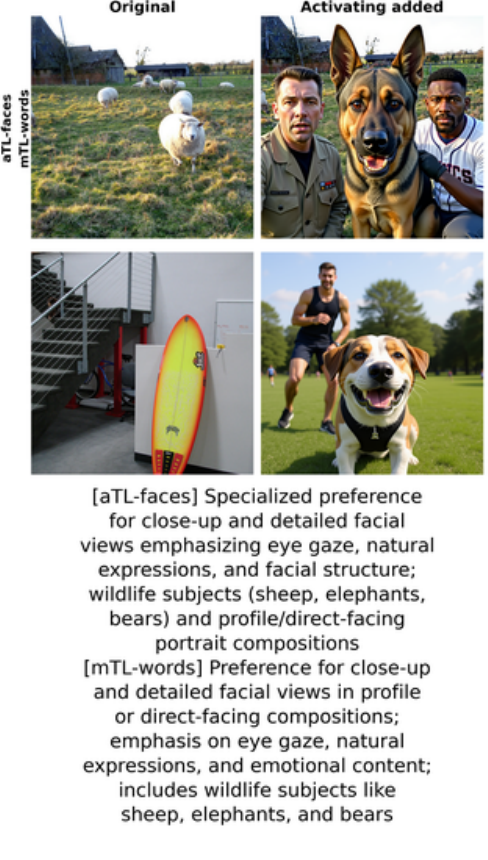}\caption*{\footnotesize voxel\_9867}\end{subfigure}\hfill
\begin{subfigure}[t]{0.18\textwidth}\centering\includegraphics[width=\textwidth]{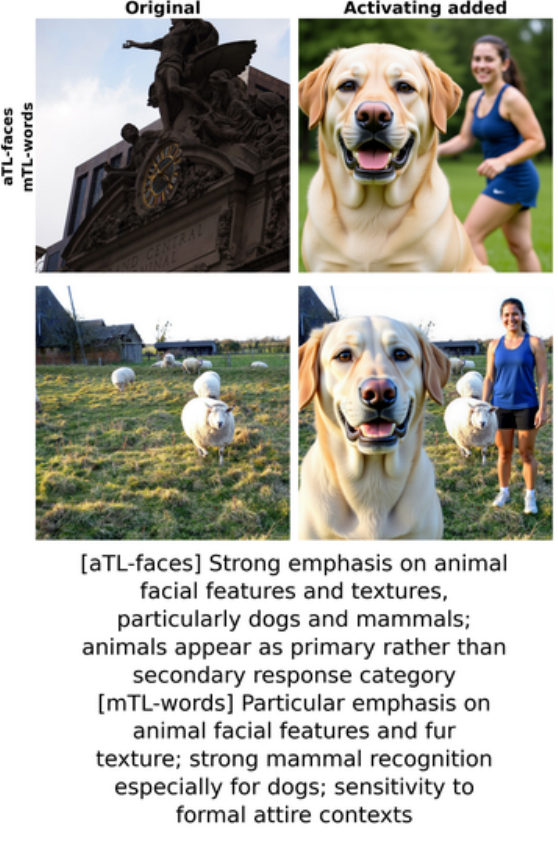}\caption*{\footnotesize voxel\_9868}\end{subfigure}\hfill
\begin{subfigure}[t]{0.18\textwidth}\centering\includegraphics[width=\textwidth]{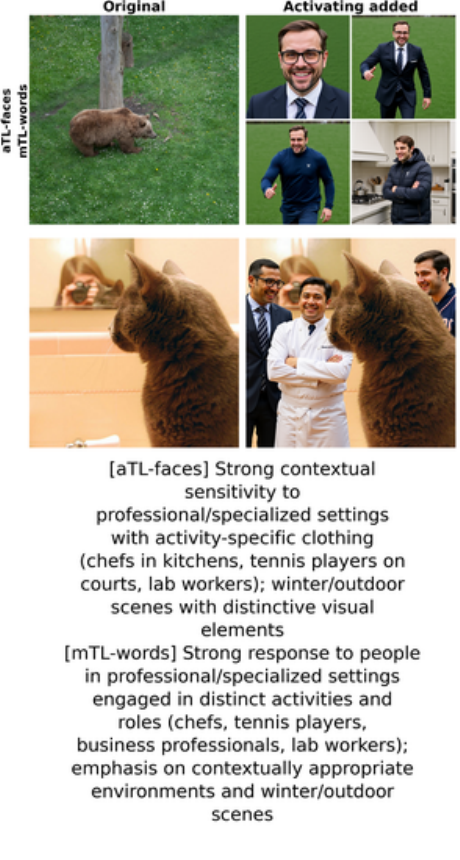}\caption*{\footnotesize voxel\_9870}\end{subfigure}\\[0.8em]
\begin{subfigure}[t]{0.18\textwidth}\centering\includegraphics[width=\textwidth]{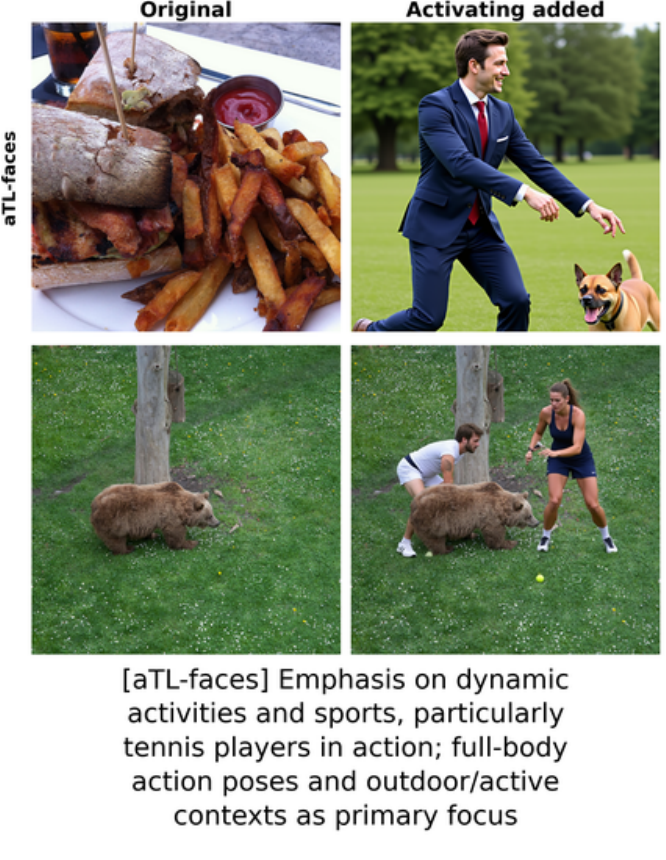}\caption*{\footnotesize voxel\_21321}\end{subfigure}
\caption{\textbf{aTL-faces voxel profiles.} \textbf{Shared profile:} \textit{All voxels show strong responsiveness to human subjects with attention to facial features, distinctive clothing details, and contextual settings. Secondary responsiveness to animals (particularly mammals like dogs) with emphasis on facial features and distinctive visual characteristics is present across voxels. Sensitivity to both portrait/close-up compositions and full-body depictions in various contexts (professional, athletic, formal) is shared.}}
\label{app:fig:atlfaces-profiles}
\end{figure}

\begin{figure}[!ht]
\centering
\begin{subfigure}[t]{0.18\textwidth}\centering\includegraphics[width=\textwidth]{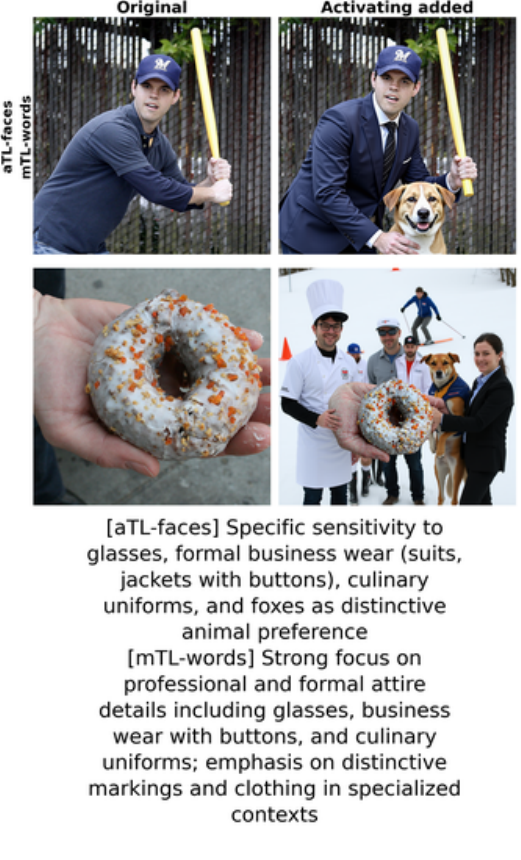}\caption*{\footnotesize voxel\_9865}\end{subfigure}\hfill
\begin{subfigure}[t]{0.18\textwidth}\centering\includegraphics[width=\textwidth]{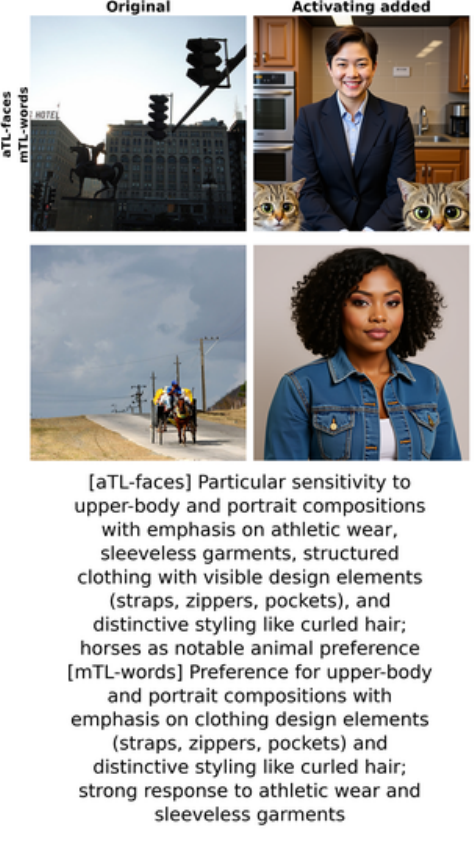}\caption*{\footnotesize voxel\_9866}\end{subfigure}\hfill
\begin{subfigure}[t]{0.18\textwidth}\centering\includegraphics[width=\textwidth]{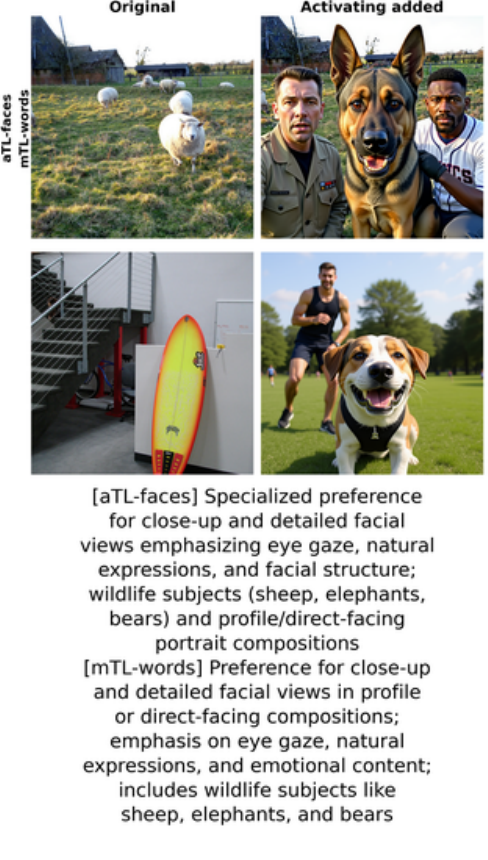}\caption*{\footnotesize voxel\_9867}\end{subfigure}\hfill
\begin{subfigure}[t]{0.18\textwidth}\centering\includegraphics[width=\textwidth]{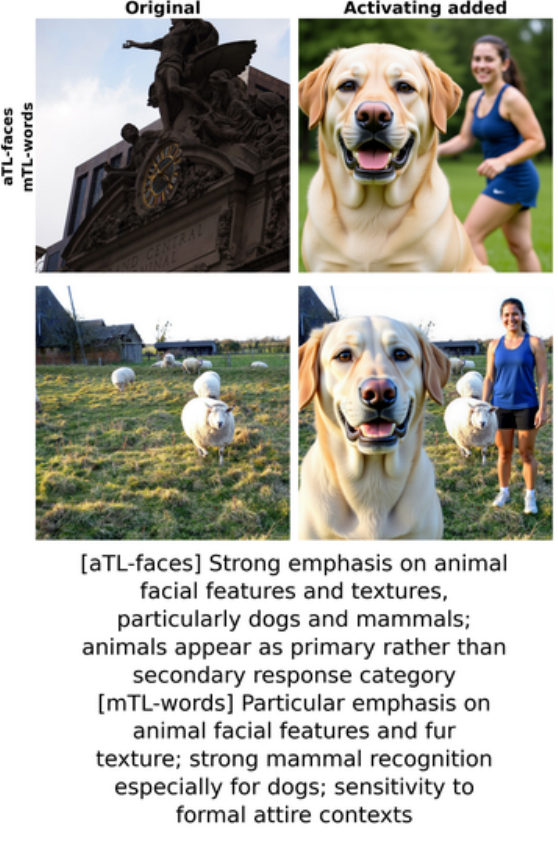}\caption*{\footnotesize voxel\_9868}\end{subfigure}\hfill
\begin{subfigure}[t]{0.18\textwidth}\centering\includegraphics[width=\textwidth]{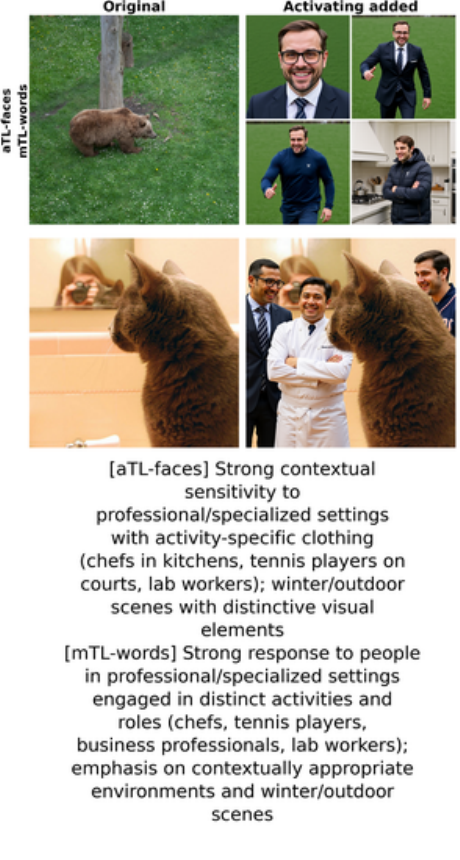}\caption*{\footnotesize voxel\_9870}\end{subfigure}
\caption{\textbf{mTL-words voxel profiles.} \textbf{Shared profile:} \textit{All voxels show strong responsiveness to human subjects and animals with emphasis on facial features, distinctive visual characteristics, and contextual details. There is consistent sensitivity to both people and animals across various compositions and attire types.}}
\label{app:fig:mtlwords-profiles}
\end{figure}

\begin{figure}[!ht]
\centering
\begin{subfigure}[t]{0.18\textwidth}\centering\includegraphics[width=\textwidth]{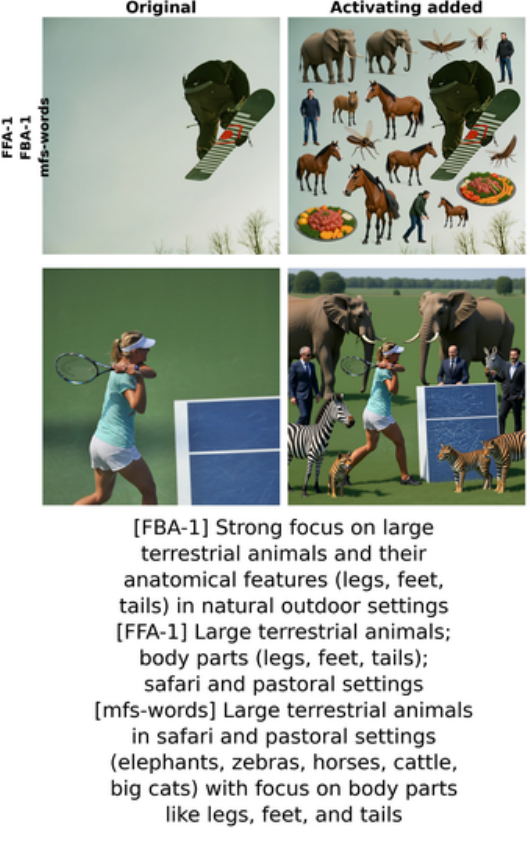}\caption*{\footnotesize voxel\_8646}\end{subfigure}\hfill
\begin{subfigure}[t]{0.18\textwidth}\centering\includegraphics[width=\textwidth]{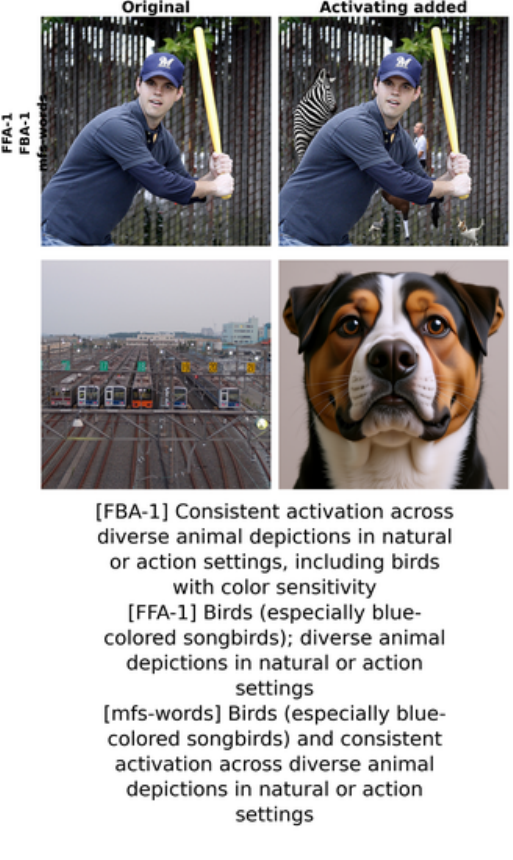}\caption*{\footnotesize voxel\_8754}\end{subfigure}\hfill
\begin{subfigure}[t]{0.18\textwidth}\centering\includegraphics[width=\textwidth]{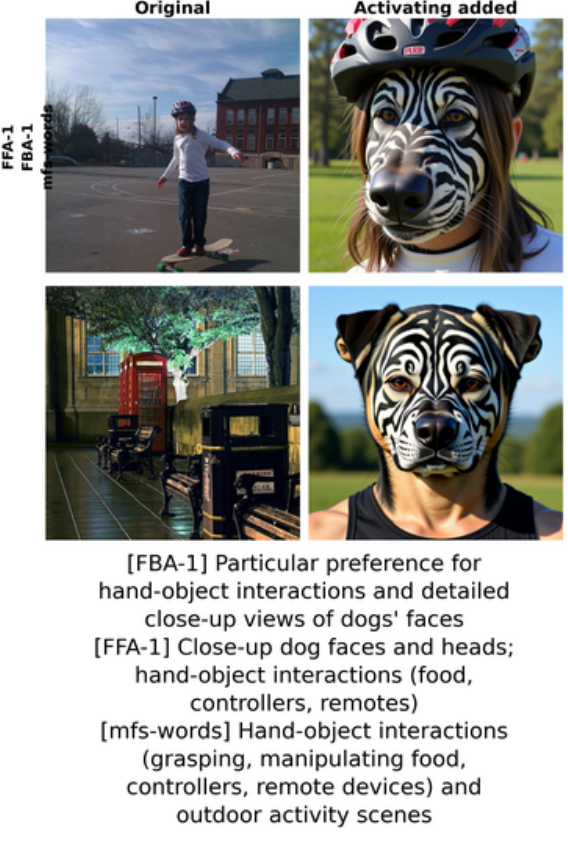}\caption*{\footnotesize voxel\_8755}\end{subfigure}\hfill
\begin{subfigure}[t]{0.18\textwidth}\centering\includegraphics[width=\textwidth]{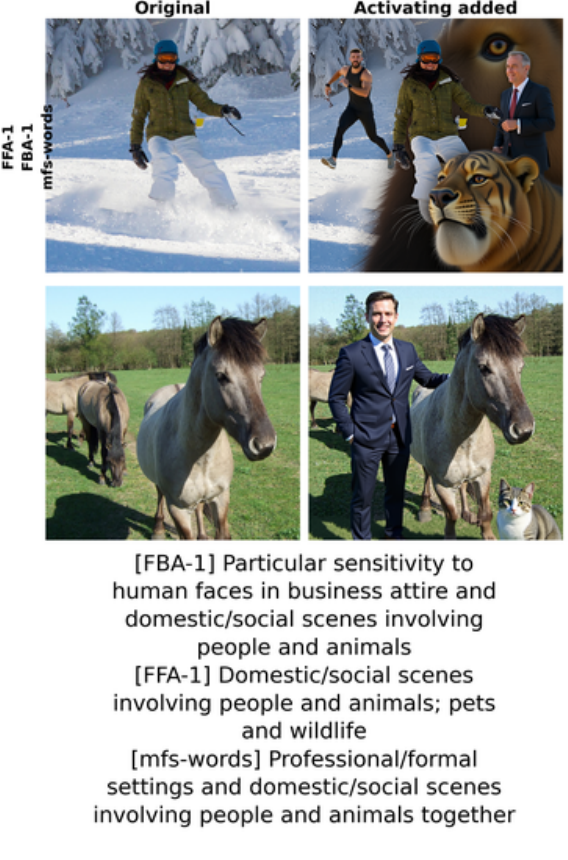}\caption*{\footnotesize voxel\_8758}\end{subfigure}\hfill
\begin{subfigure}[t]{0.18\textwidth}\centering\includegraphics[width=\textwidth]{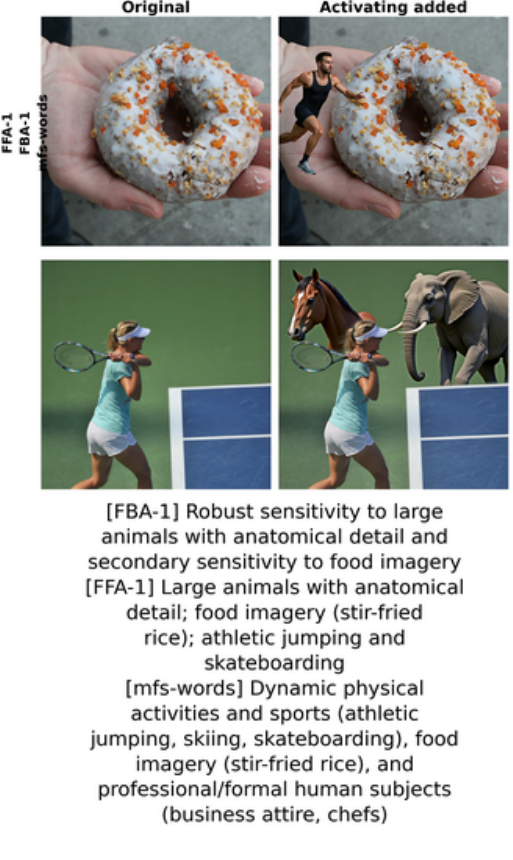}\caption*{\footnotesize voxel\_8759}\end{subfigure}
\caption{\textbf{mfs-words voxel profiles.} \textbf{Shared profile:} \textit{Close-up facial imagery and anatomical details of both humans and animals, with sensitivity to portraits, faces with distinct features, and animals depicted with visible body parts and physical characteristics.}}
\label{app:fig:mfswords-profiles}
\end{figure}

\section{Brief introduction to neuroimaging}\label{app:neuroimaging}
In this addendum we provide a brief, partial introduction to neuroimaging, with the goal of providing unfamiliar readers with the necessary background for a deeper understanding. Functional Magnetic Resonance Imaging (fMRI) is a neuroimaging technique that measures brain activity indirectly, through changes in blood oxygenation \citep{logothetisUnderpinningsBOLDFunctional2003,kimBiophysicalPhysiologicalOrigins2012}. When neurons in a region fire, local metabolic demand draws oxygenated blood to the area; the resulting shift in the ratio of oxygenated to deoxygenated hemoglobin alters local magnetic susceptibility, producing the \textit{blood-oxygen-level-dependent} (BOLD) signal that fMRI scanners record. The BOLD signal has been shown to correlate with the spiking activity of underlying neurons \citep{logothetisInterpretingBOLDSignal2004,nirBOLDSpikingActivity2008}, providing an indirect read-out of population firing. To localize these recordings, the brain volume is partitioned into a 3D grid of small cubes (typically $\sim$1\,mm$^3$) called \textit{voxels}, and one BOLD value is recorded per voxel per timepoint. Each voxel therefore aggregates the activity of many neurons (on the order of hundreds of thousands), but remains informative, as supported by noise-ceiling measurements. This pooling is consistent with the columnar organization of cortex \citep{mountcastleModalityTopographicProperties1957,mountcastleColumnarOrganizationNeocortex1997}, in which neurons within a small region share functional preferences; face-selective cortex provides the clearest demonstration, where an fMRI-localized patch was shown to consist almost entirely of face-tuned neurons via single-unit recordings \citep{CorticalRegionConsisting}. At the single-neuron level, decades of electrophysiology have established that visual neurons behave as approximate \textit{signal detectors}: a neuron responds strongly when its preferred features are present in the stimulus and remains near baseline otherwise \citep{hubelReceptiveFieldsBinocular1962}. Because a voxel pools neurons with similar tuning, the signal-detector framing carries up to the voxel level, and motivates the formal setting in Section~\ref{sec:methods:setting}, where we model each voxel's response as the probability that a stimulus contains the features the voxel is tuned to.

\clearpage

\end{document}